\documentclass[lettersize,journal]{IEEEtran}
\usepackage[utf8]{inputenc}
\usepackage[T1]{fontenc}
\usepackage{microtype}
\usepackage{lettrine}
\usepackage{xcolor}         
\usepackage{graphicx}       
\usepackage{float}          
\usepackage[caption=false,font=normalsize,labelfont=sf,textfont=sf]{subfig} 
\usepackage{graphicx}
\usepackage{subcaption}
\usepackage{enumitem}
\usepackage{amsmath,amssymb,amsfonts} 
\usepackage{nicefrac}                 

\usepackage{booktabs}   
\usepackage{multirow}   
\usepackage[table]{xcolor}
\usepackage{pdflscape}  
\usepackage{array}      
\usepackage{adjustbox}  
\usepackage[justification=raggedright,singlelinecheck=false]{caption}
\usepackage{algorithm}
\usepackage{algpseudocode} 
\usepackage{url}
\usepackage{stfloats}
\usepackage{hyperref}
\usepackage[numbers,sort&compress]{natbib}

\usepackage{hyperref}
\usepackage{url}
\hypersetup{
    colorlinks=true,
    linkcolor=red,
    citecolor=blue,
    urlcolor=cyan
}

\begin{document}

\title{SPORT: Structure-Aware Prototype Disentanglement for Incomplete Multi-View Clustering}

\author{Yaoyuan Guo,
Zhibin Gu,
Songhe Feng,
Yuhui Zheng,
Bing Li
\thanks{Y. Guo is with the College of Computer and Cyberspace Security, Hebei Normal University, Shijiazhuang 050024, China, and also with the Department of Statistics and Data Science, Southern University of Science and Technology, Shenzhen 518055, China. (e-mail: 12312902@mail.sustech.edu.cn).}

\thanks{Z. Gu is with the College of Computer and Cyberspace Security, Hebei Normal University, Shijiazhuang 050024, China, and also with the Key Laboratory of Tibetan Information Processing, Ministry of Education, Qinghai Normal University, Xining 810008, China. (e-mail: guzhibin@hebtu.edu.cn).}
\thanks{S. Feng is with the School of Computer Science and Technology, Beijing Jiaotong University, Beijing 100044, China. (e-mail: shfeng@bjtu.edu.cn).}
\thanks{Y. Zheng is with the Key Laboratory of Tibetan Information Processing, Ministry of Education, Qinghai Normal University, Xining 810008, China. (e-mail: zhengyh@vip.126.com).}

\thanks{B. Li is with the State Key Laboratory of Multimodal Artificial Intelligence Systems, Institute of Automation, Chinese Academy of Sciences, Beijing 100190, China. (e-mail: bli@nlpr.ia.ac.cn).}

\thanks{Corresponding author: Zhibin Gu.}}

\markboth{Journal of \LaTeX\ Class Files,~Vol.~14, No.~8, August~2021}%
{Shell \MakeLowercase{et al.}: A Sample Article Using IEEEtran.cls for IEEE Journals}

\maketitle

\begin{abstract}
Prototype-based Incomplete Multi-view Clustering has recently attracted increasing attention by exploiting prototypes as semantic anchors for missing-view imputation. However, existing approaches are still limited in three aspects. First, they typically focus on enforcing cross-view prototype consistency, while ignoring view-specific information embedded in prototypes, thus limiting multi-view expressiveness. Second, most methods rely on instance-level contrastive learning that only aligns paired samples across views, failing to preserve cluster-level relational structures. Third, missing-view imputation is usually performed using global prototypes alone, without considering local geometric neighborhood structures, leading to inaccurate recovery of missing representations. To address these limitations, we propose a novel framework termed \textbf{S}tructure-aware \textbf{P}r\textbf{O}totype disentanglement fo\textbf{R} incomplete multi-view clus\textbf{T}ering (SPORT), which explicitly disentangles shared and view-specific components of prototypes while preserving cluster-level relational structures. Specifically, we decouple prototypes into orthogonal shared and view-specific components, aligning only shared components to capture consensus semantics while decorrelating view-specific components to preserve complementary information. Meanwhile, a structure-aware contrastive learning mechanism is incorporated to explicitly model cluster-level relationships during cross-view representation learning. Furthermore, a hybrid imputation strategy integrates global prototype matching with local neighborhood matching, enabling joint exploitation of semantic prototypes and manifold structures for missing-view recovery. Extensive experiments on six benchmark datasets show that SPORT achieves superior performance over state-of-the-art methods under various missing rates.

\begin{IEEEkeywords}
Incomplete multi-view clustering, prototype disentanglement, structure-aware contrastive learning, missing-view imputation
\end{IEEEkeywords}

\end{abstract}

\section{Introduction}
\lettrine{With} the rapid development of knowledge discovery and data mining techniques, data are increasingly collected and stored in multiple modalities. Such multi-source data are referred to as multi-view data. Multi-View Learning (MVL), which aims to leverage the shared (common) and complementary information across different views to enhance the performance of various learning tasks, has been extensively studied \cite{xu2022multi2,white2012convex,xu2015multi3,zhang2020deep2}. As a representative unsupervised paradigm within MVL, Multi-View Clustering (MVC) seeks to partition unlabeled multi-view data by identifying shared clustering structures across multiple views, thereby offering a more comprehensive characterization of the underlying relationships among samples \cite{11458690,11475667,xu2016discriminatively,cui2023novel}.
Conventional MVC frameworks typically assume that all samples are fully observed across all views \cite{wang2026llm,xin2025multilevel,gu2026hypergraph,gu2026twin,trosten2023effects,zhang2016multi2}. However, in real-world scenarios, incomplete multi-view data are ubiquitous due to sensor limitations, hardware failures, or storage constraints, with some samples missing one or more views \citep{11277376,ding2025incomplete1,yuan2025prototype,liu2025reliable}, which significantly impair the performance of conventional multi-view clustering methods. To address this challenge, recent years have witnessed the emergence of numerous Incomplete Multi-View Clustering (IMVC) approaches \cite{wang2026adversarial,zhang2026dynamic,zhong2026gaussian,wu2025koala,zhang2019cpm,xue2021clustering}, which are designed to accommodate arbitrary missing-view patterns while leveraging complementary information from available views to uncover semantically consistent clustering structures.


According to the learning paradigm, existing IMVC methods can be broadly categorized into traditional methods \cite{zhao2017multiview,hu2018doubly,wen2024matrix,zhou2019consensus,wen2020incomplete,deng2024projective,xu2018partial,ye2017consensus,liu2019multiple,guo2019anchors} and 
deep learning-based methods \citep{ding2025incomplete1,chao2024incomplete,chao2021survey,wen2021structural,xiao2025easemvc,wen2020dimc,lin2023ccr,xu2023adaptive,zhang2024incomplete,xu2023untie}. 
Owing to their superior representation capacity and scalability, 
deep learning-based methods have garnered increasing attention in addressing IMVC tasks. For instance, \citet{lin2021completer} proposed an information-theoretic framework that jointly performs data recovery and consistency learning, achieving cross-view consistency and reconstructing missing data via dual prediction and contrastive losses. \citet{tang2022deep} designed a neighborhood-based IMVC framework that 
imputed missing views from dynamically mined $k$-nearest neighbors 
and introduced a weighting function to suppress low-quality imputations. Motivated by the strong generative capability of generative models, \citet{shang2017vigan} and \citet{wen2024diffusion} respectively adopt generative adversarial networks (GANs) and diffusion models for missing-view reconstruction. Moreover, a growing line of literature focuses on prototype-based methods \cite{yuan2025prototype,jin2023deep,wang2026learning,li2023incomplete,li2025attention,du2025pgformer,zhu2026prototype}, which exploit global semantics via representative prototypes to better handle missing information and improve clustering robustness. For example, \citet{wang2026learning} aligned the prototypes via contrastive learning and recovered the missing data through an explicit prototype graph matching strategy. \citet{li2023incomplete} developed a dual-stream model that mutually 
refined sample and prototype representations through attention-based 
interaction, with missing data reconstructed by the learned prototypes. Furthermore, \citet{li2025attention} designed a  bi-alignment guidance module to ensure discriminative representations and employed an attention layer to enhance the compactness of the view-specific sample-prototype relationship.

\begin{figure}[!t]
    \centering
    \includegraphics[width=1\linewidth]{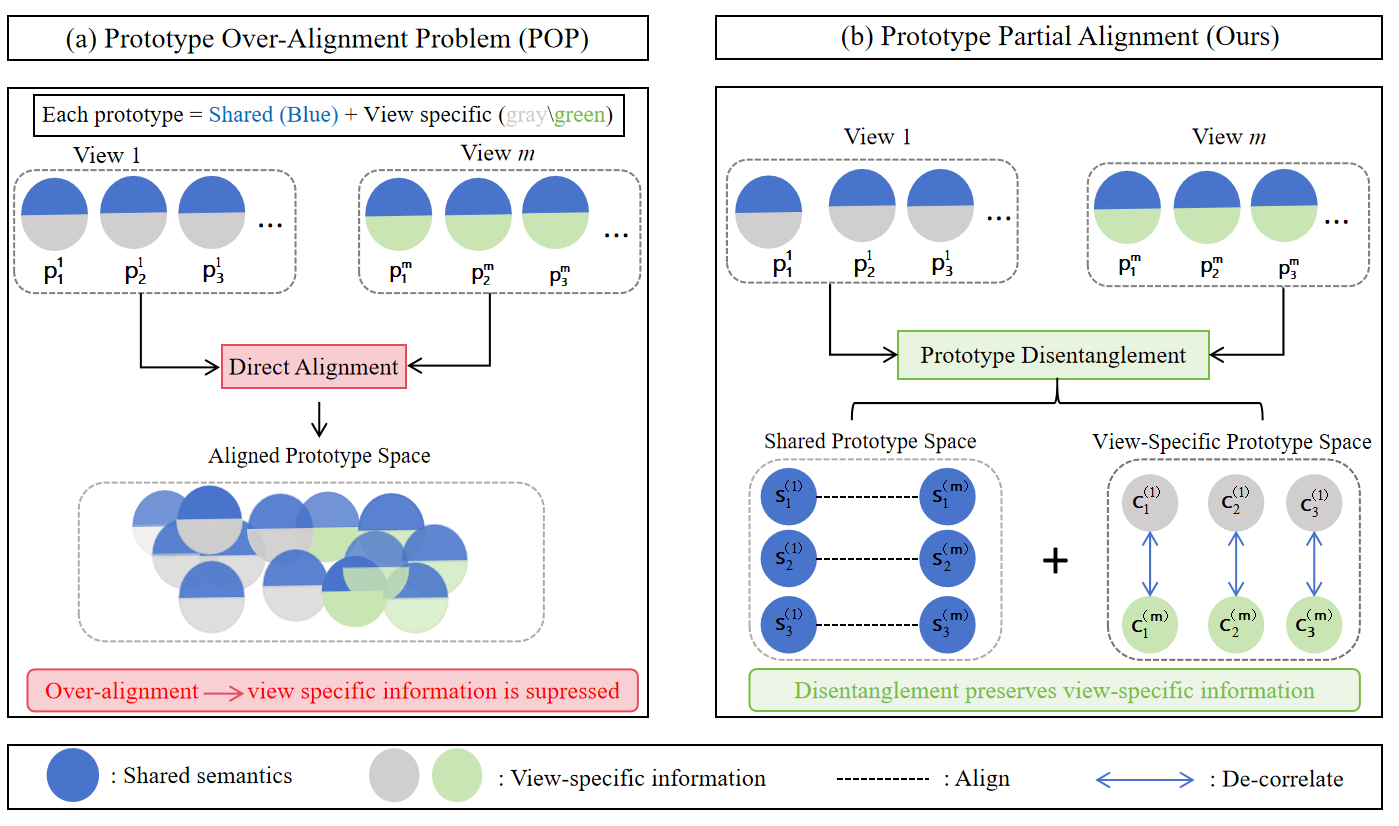}
\caption{Illustration of the Prototype Over-alignment Problem (POP) and the proposed prototype decomposition paradigm. (a) Full cross-view prototype alignment suppresses view-specific information and reduces representation diversity. (b) Our method decouples shared and view-specific components and align only shared semantics across views.}
    \label{fig:pop}
\end{figure}

Despite the notable advancement in prototype-based IMVC methods, 
three key limitations persist. First, existing methods typically enforce strict cross-view prototype alignment under the assumption that corresponding prototypes should share identical semantic representations across views. However, each view is generated from a distinct observation process, and its prototype inherently encodes both shared semantics and view-specific characteristics. Such overly strict alignment suppresses view-specific information, limiting the expressive capacity of multi-view representations, a phenomenon we term the Prototype Over-alignment Problem (POP), as illustrated in Fig.\ref{fig:pop}(a). Second, most methods employ hard contrastive learning  that treats only same-sample pairs as positive pairs, discarding cluster-level semantic similarity among neighboring instances (i.e., samples 
belonging to the same cluster generally exhibit closely related 
features which should also be harnessed), causing semantically 
similar samples to be pushed apart in the feature space. 
Furthermore, existing prototype-based imputation methods rely exclusively on prototypes as semantic anchors for missing-view imputation. Due to the coarse-grained nature of prototypes, they primarily capture global cluster-level semantics while overlooking local neighborhood relationships, resulting in structurally over-smoothed reconstructions that fail to preserve fine-grained instance-level consistency.

\begin{figure*}[!t]
    \centering
    \includegraphics[width=1\linewidth]{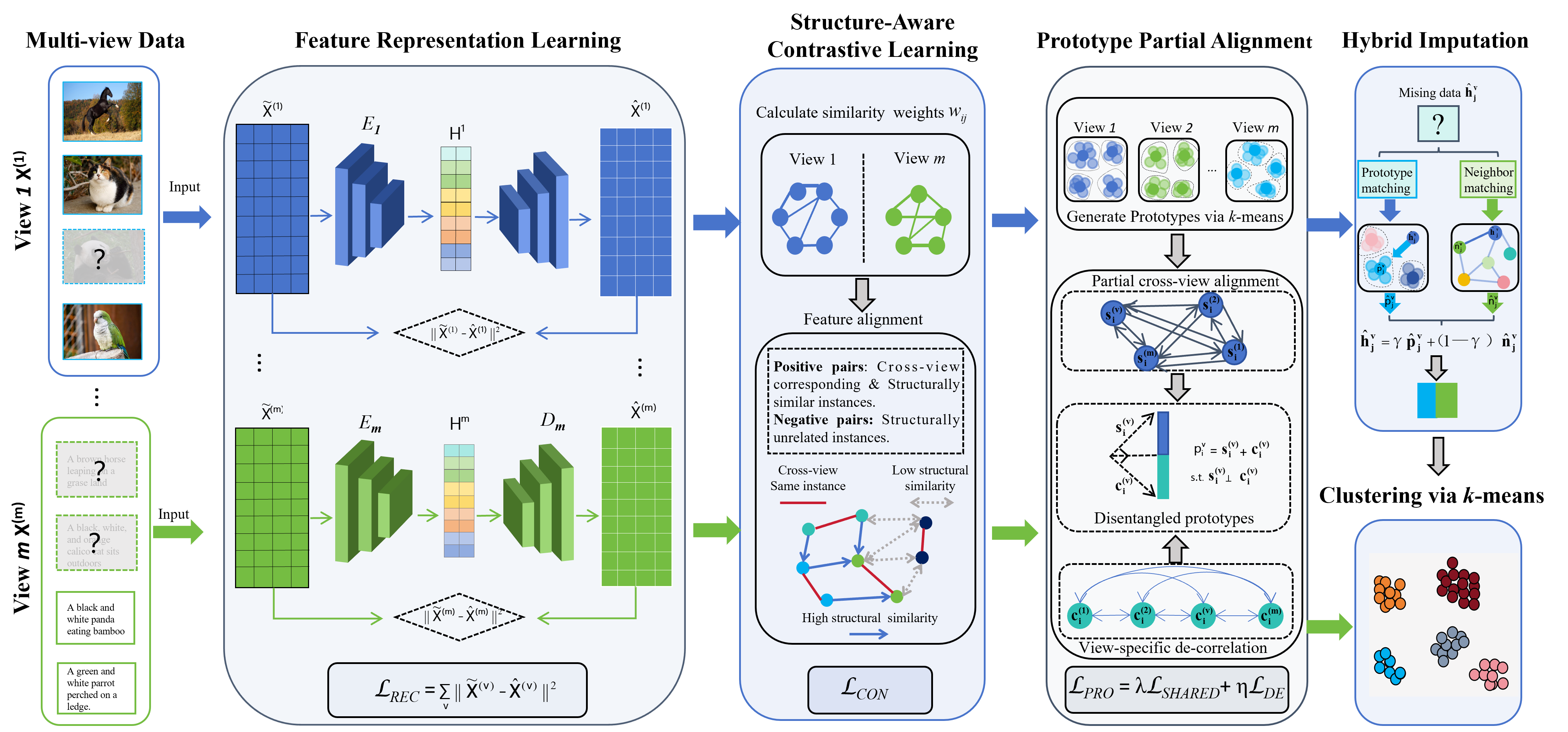}
    \caption{Framework of the proposed SPORT. We first learn feature representations by mapping complete inputs $\tilde{\mathrm{\textbf{X}}}^{(v)}$ into a shared latent space $\mathrm{\textbf{H}}^v \in \mathbb{R}^d$. Built upon these representations, a structure-aware relational contrastive learning module is introduced to jointly capture cross-view consistency and cluster-level structure by aligning semantically consistent samples. Next, we construct prototypes and decompose them into shared and view-specific components, where shared components are aligned across views while view-specific components are de-correlated to preserve complementary information. Finally, a hybrid prototype imputation strategy is developed, which leverages both matched prototypes and local neighborhood information to complete missing data effectively.}
    \label{fig:procedure}
\end{figure*}

To address the aforementioned issues, we propose a novel Structure-aware PrOtotype disentanglement foR
incompleTe multi-view clustering framework (SPORT), which jointly exploits cross-view consistency, view-specific semantics, and structural relationships to learn robust and discriminative clustering representations. Specifically, as illustrated in Fig. \ref{fig:procedure}, we explicitly decompose the multi-view prototypes obtained via $k$-means into complementary shared and view-specific components (Fig. \ref{fig:pop}(b)). The shared components are aligned across views to capture view-invariant semantics, while the view-specific components are regularized via a decorrelation constraint to preserve discriminative view-dependent information. This prototype decomposition strategy effectively disentangles cross-view consensus from view-specific semantics, leading to more discriminative clustering representations. In addition, a weighted contrastive learning mechanism is designed, where similarity-aware weights are introduced to characterize the structural relationships among samples, enabling cross-view feature alignment while preserving the underlying geometric structure of the data. Moreover, SPORT employs a hybrid imputation scheme that reconstructs missing samples by jointly exploiting information from matched prototypes and neighboring instances. By integrating cluster-level semantics with local structural information, the proposed strategy improves the quality of missing-data recovery and further enhances clustering performance. In summary, the contributions of our work can be summarized as follows:

\begin{itemize}

\item We propose a unified prototype alignment framework by decomposing prototypes into shared and view-specific components. The shared components are aligned across views, while the view-specific components are regularized via decorrelation, enabling effective integration of shared semantics and view-specific information and alleviating prototype misalignment.

\item We develop a structure-aware feature alignment mechanism by introducing instance-wise similarity weights to softly encode inter-sample relationships. This enables the contrastive objective to enforce cross-view consistency while preserving cluster-level relational structures, yielding more discriminative representations.

\item We design a hybrid prototype imputation strategy that imputes missing data by integrating information from matched prototypes and neighboring samples, thereby jointly exploiting global cluster semantics and local structural information to improve reconstruction quality. 

\item Extensive experiments on six benchmark datasets against 14 state-of-the-art incomplete multi-view clustering methods demonstrate that SPORT achieves superior performance.

\end{itemize}
\vspace{-\baselineskip}

\section{Related Work}

Benefiting from the representation learning capability of deep neural networks, various deep learning-based incomplete multi-view clustering (DIMVC) methods have been developed \cite{zhang2026dynamic,wang2026adversarial,wu2025imputation,pu2024adaptive,bai2024graph,lin2023consistent,chen2025deep,dong2025selective,feng2024partial}. These methods can be broadly categorized into four classes: generator-based, predictor-based, neighborhood-based, and prototype-based methods.

\subsection{Generator-based methods} 
Generator-based methods leverage the strong generative capability of generative models to reconstruct missing information from observed representations \cite{shang2017vigan,wang2023self,xu2019adversarial,wang2018partial,wen2024diffusion,fang2024incomplete,zhang2025incomplete,goodfellow2020generative,wang2021generative}. Two predominant generative paradigms applied to IMVC tasks are generative adversarial networks (GANs) \cite{goodfellow2020generative} and diffusion models \cite{ho2020denoising}. Early works predominantly adopt GANs due to their ability to learn high-fidelity representations via adversarial training. For example, \citet{xu2019adversarial} proposed an adversarial feature recovery framework based on GANs, while \citet{wang2021generative} incorporated cycle-consistency and adaptive fusion to handle incomplete views. More recently, diffusion models have attracted increasing attention due to their training stability and strong mode coverage. Representative works include \citet{fang2024incomplete}, which initialize missing features with Gaussian noise and progressively recover them via a diffusion-based completion process, and \citet{zhang2025incomplete}, which further integrates diffusion modeling with contrastive learning, yielding more discriminative and compact clustering representations.

\subsection{Predictor-based methods}

{Predictor-based methods} estimate missing features from observed ones from an information-theoretic perspective \cite{lin2021completer, lin2022dual, xu2023adaptive, 
lu2024decoupled, ding2025incomplete1}.  For example, \citet{lin2021completer} first established 
a unified information-theoretic framework that jointly learned consistent representations and recovers 
missing views by maximizing cross-view mutual information. Building upon this work, \citet{lin2022dual}  introduced a dual contrastive loss 
and a dual prediction loss to simultaneously enforce cross-view 
consistency and data recoverability. To further mitigate the distribution 
discrepancy between complete and incomplete data, \citet{xu2023adaptive} 
designed an adaptive feature projection mechanism that aligned the distributions between complete and incomplete data by
optimizing mutual information, yielding more faithful feature recoverability. Beyond single-level prediction, \citet{yin2025incomplete} formulated a multi-level dual prediction strategy in which both high-level and semantic features are extracted via an attention layer and optimized for missing data prediction.

\subsection{Neighborhood-based methods} 
Neighborhood-based methods exploit structural relationships to address data incompleteness by leveraging local topology within or across views \cite{yang2022robust,tang2022deep,wen2021structural,teng2024urrl,yan2025neighbor}. K-nearest neighbor (kNN)  \cite{peterson2009k} is a classic machine learning method that is used to integrate features from the top-K spatial neighbors. For example, \citet{teng2024urrl} integrated multi-view and neighborhood information using complementary encoders with kNN-based augmentation and reconstruction. \citet{yan2025neighbor} explored the latent common clustering information embedded in multi-view data and performed neighbor-based completion via kNN. Beyond instance-level aggregation, graph neural networks (GNNs) \cite{corso2024graph} provide an alternative strategy for recovering missing views from adjacent samples. Typical works include \citet{wen2021structural}  used a GCN to aggregate adjacent information and fused a consensus graph to repair missing view features and \citet{chao2024incomplete} which transferred multi-view consistency relations 
via GCN, while applying high-confidence guiding with 
instance-level contrastive learning for missing view reconstruction.

 \subsection{Prototype-based methods} Prototype-based methods capture global semantics by learning view-wise prototypes as representative anchors and leveraging them for missing-data imputation \cite{yuan2025prototype,jin2023deep,li2025attention,wang2026learning,li2023incomplete,du2025pgformer,zhu2026prototype}. These prototypes are often misaligned across views, which has motivated extensive research on cross-view prototype alignment. For example, \citet{yuan2025prototype} addressed the prototype unaligned problem (PUP) by aligning the generated prototypes across views via contrastive learning and proposed a novel prototype matching framework to impute the missing data. \citet{jin2023deep} considered the IMVC problem from a partially-aligned setting and generated view-specific prototypes via $l_2$ loss minimization while reconstructing missing views through contrastive learning-free aligned prototypes. Other works utilized graphs to capture semantic dependencies among prototypes to enhance prototype alignment. For instance, \citet{du2025pgformer} formulated a prototype-guided graph transformer framework that integrated dual attention mechanisms with cross-view alignment which mitigated the influence of noise and outliers.  \citet{zhu2026prototype} introduced a similarity graph-based framework that captures semantic-invariant structural information across prototypes for prototype alignment, further improving quality of the aligned prototypes. 

\section{Method}

\subsection{Notation and Problem Statement}
In this paper, $\{\mathrm{\textbf{X}}^{(v)} \in \mathbb{R}^{n \times d_v}\}_{v=1}^{m}$ denotes the multi-view incomplete dataset consisting of $n$ samples across $m$ views with $k$ clusters, where $d_v$ is the feature dimension of the $v$-th view. Due to incomplete observations, we define the observed view set of sample $i$ as :
\begin{equation}
    \mathcal{V}_i = \{v \mid \mathrm{\textbf{x}}_i^{(v)} \text{ is observed}\} \subseteq \{1, \dots, m\}
    \label{eq:v}
\end{equation}

Let $\tilde{\mathrm{\textbf{X}}}^{(v)} \in \mathbb{R}^{n_v \times d_v}$ denote the observed instances in view $v$, where $n_v = |\{i \mid v \in \mathcal{V}_i\}|$ is the number of observed instances in view $v$. $E_v$ and $D_v$ are the respective encoder and decoder for view $v$, $
\mathrm{\textbf{H}}^{v} \in \mathbb{R}^{n_{v} \times d}$  and $\hat{\mathrm{\textbf{X}}}^{(v)}\in \mathbb{R}^{n_{v} \times d_v}$ are the unified and reconstructed representations of $\tilde{\mathrm{\textbf{X}}}^{(v)}$. We use $\mathrm{\textbf{h}}_i^v$ to denote the $i$th row of $\mathrm{\textbf{H}}^v$. $\mathrm{\textbf{P}}_v \in \mathbb{R}^{k \times d}$ refers to the representation of $k$ prototypes in view $v$.

Incomplete multi-view clustering aims to partition $n$ unlabeled samples into $k$ disjoint 
clusters through unsupervised learning, where each sample 
is only observed in certain views. 
To achieve this objective, we propose a structure-aware prototype disentanglement framework (SPORT), whose overall architecture is illustrated in Fig.~\ref{fig:procedure}. Specifically, given incomplete multi-view data, we first use autoencoders to project all complete features into a common subspace. Subsequently, similarity weights are calculated to measure inter-instance affinities, which guide the relational contrastive objective to jointly capture cross-view semantic consistency and cluster-level structural consistency among samples. In addition, prototypes are generated via $k$-means and decomposed into shared and view-specific components, where the shared parts are aligned across views while the view-specific parts are de-correlated to preserve both consensus and complementary information. Finally, missing data are imputed hybridly via related neighbors and prototypes and cluster assignments are obtained via $k$-means.

\subsection{Feature Representation Learning}

 Given the observed dataset $\tilde{\mathrm{\textbf{X}}}^{(v)} \in \mathbb{R}^{n_v \times d_v}$ of view $v$, to capture the relationships among samples, we utilize view-specific autoencoders \cite{li2023comprehensive} to project all features into a common latent space. Specifically, using an encoder $E_v: \mathbb{R}^{n_v \times d_v} \rightarrow \mathbb{R}^{n_v \times d}$ and a decoder $D_v: \mathbb{R}^{n_v \times d} \rightarrow \mathbb{R}^{n_v \times d_v}$, the latent representation and the subsequent reconstruction process are formulated as:
\begin{align}
    \mathrm{\textbf{H}}^v &= E_v(\tilde{\mathrm{\textbf{X}}}^{(v)}), \label{eq:encoder} \\[5pt]
    \hat{\mathrm{\textbf{X}}}^{(v)} &= D_v(\mathrm{\textbf{H}}^v), \label{eq:decoder}
\end{align}
where $\mathrm{\textbf{H}}^v \in \mathbb{R}^{n_v \times d}$ serves as the latent representation, and $\hat{\mathrm{\textbf{X}}}^{(v)} \in \mathbb{R}^{n_v \times d_v}$ is the reconstructed feature corresponding to view $v$.

To ensure the mapping fidelity of the autoencoders and preserve inherent structural characteristics, the overall reconstruction loss is defined as the Frobenius norm between the input and the reconstructed output across all views:
\begin{equation}
    \mathcal{L}_{\text{REC}} = \sum_{v=1}^{m} \left\| \tilde{\mathrm{\textbf{X}}}^{(v)} - \hat{\mathrm{\textbf{X}}}^{(v)} \right\|_F^2,
    \label{eq:rec_loss}
\end{equation}
where $\| \cdot \|_F$ denotes the Frobenius norm.

\subsection{Structure-Aware Relational Contrastive Learning}

\begin{figure}[t]
    \centering
    \includegraphics[width=1\linewidth]{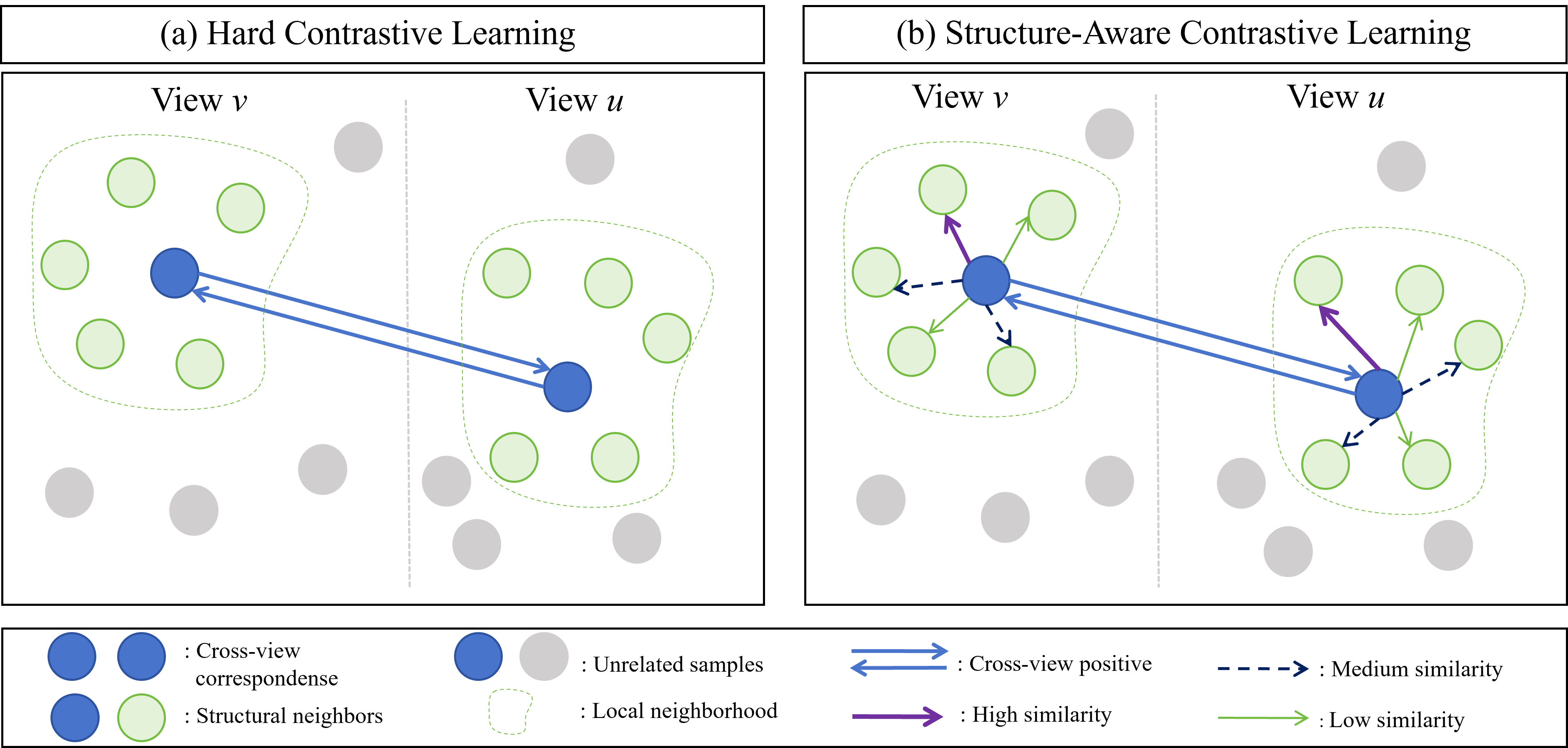}
\caption{Illustration of conventional hard contrastive learning and our structure-aware relational contrastive learning. (a) Hard contrastive learning ignores inter-sample structure by treating all non-positive samples as negatives. (b) Our method introduces similarity-aware weighting to preserve cluster-level consistency.}
    \label{fig:contrastive}
\end{figure}

Although the feature representation learning module maps all features into a unified latent subspace with high reconstruction fidelity, it neglects the dual consistency nature of multi-view data which also needs to be harnessed. Specifically, such consistency arises from two complementary aspects: (1) cross-view consistency, where different views of the same sample share common semantic information, and (2) structural consistency, where samples belonging to the same cluster tend to exhibit similar feature characteristics. Neglecting either aspect may hinder the learning of discriminative and semantically coherent representations. However, existing contrastive learning-based IMVC methods primarily exploit cross-view consistency by aligning representations of the same sample across different views \cite{li2023incomplete,jin2023deep,yuan2025prototype}, while largely neglecting structural consistency among semantically related instances. As illustrated in Fig.~\ref{fig:contrastive}(a), these methods typically treat only cross-view pairs of the same instance as positives, while all other instances are uniformly treated as negatives. Consequently, semantically similar samples may be undesirably separated in the latent space, failing to preserve cluster-level relational structures and ultimately leading to suboptimal clustering performance. 

To address this limitation, we propose a structure-aware relational contrastive learning module to jointly model cross-view and structural consistency, as shown in Fig.~\ref{fig:contrastive}(b). Specifically, the contrastive objective takes cross-view features of the same sample as positive pairs, and further uses pairwise similarity weights to characterize inter-instance affinities. As a result, similar samples are softly involved as positives, whereas unrelated samples serve as negatives. The pairwise similarity weight $w_{ij}$ is defined as:
\begin{equation}
    w_{ij} = 
    \begin{cases} 
        1, & i = j, \\[8pt]
        \dfrac{
            \sum\limits_{v \in \mathcal{V}_i \cap \mathcal{V}_j} 
            \dfrac{\exp(S(\mathrm{\textbf{h}}_i^v, \mathrm{\textbf{h}}_j^v)/\tau_w)}{\exp(1/\tau_w)}
        }{
            |\mathcal{V}_i \cap \mathcal{V}_j|
        }, 
        & |\mathcal{V}_i \cap \mathcal{V}_j| \geq 1, \\[12pt]
        0, & |\mathcal{V}_i \cap \mathcal{V}_j| = 0.
    \end{cases}
\end{equation}

\noindent where $w_{ij}$ denotes the inter-instance similarity weight between sample $i$ and $j$ with $w_{ij} = w_{ji}$ and $w_{ii} = 1$. $\mathcal{V}_i$ is the set of views where sample $i$ has complete data defined in Eq.(\ref{eq:v}). ~$|\mathcal{V}_i \cap \mathcal{V}_j|$ is the cardinality of $\mathcal{V}_i \cap \mathcal{V}_j$ which represents the set of views where sample $i$ and $j$ both have complete data. $\tau_w$ is a temperature hyperparameter, and $S(\mathrm{\textbf{h}}_i^v, \mathrm{\textbf{h}}_j^v)$ is the cosine similarity between $\mathrm{\textbf{h}}_i^v$ and $\mathrm{\textbf{h}}_j^v$ calculated as: 
\begin{equation}
    S(\mathrm{\textbf{h}}_i^v, \mathrm{\textbf{h}}_j^v) = \frac{(\mathrm{\textbf{h}}_i^v)^\top \mathrm{\textbf{h}}_j^v}{\|\mathrm{\textbf{h}}_i^v\|\|\mathrm{\textbf{h}}_j^v\|}
\end{equation}

These weights characterize the sample-wise similarity between samples, which captures the relationship between samples. Furthermore, for a view pair $(v, u)$, we define the set of instances simultaneously observed in both views as:
\begin{equation}
    \mathcal{N}_{v,u} = \{i \mid v \in \mathcal{V}_i \ \text{and} \ u \in \mathcal{V}_i\}
\end{equation}

To achieve the objective of aligning the closely related features of samples while deviating from the uncorrelated ones, the objective loss for feature alignment is defined as~Eq. (\ref{eq:2_main}):
\begin{equation}
    \mathcal{L}_{\text{CON}} = - \sum_{\substack{v,u=1 \\ u \neq v}}^{m}  \frac{1}{n_v} \sum_{i=1}^{n_v} \mathbf{1}[i \in \mathcal{N}_{v,u}] \log P_i^{v \to u},
\label{eq:2_main}
\end{equation}

\noindent where $P_i^{v \to u}$ is defined as:
\begin{equation}
    P_i^{v \to u} = \frac{\sum_{j=1}^{n_u} w_{ij} E_{i,j}^{v,u}}
    {\sum_{j=1}^{n_v} E_{i,j}^{v,v} + \sum_{j=1}^{n_u} E_{i,j}^{v,u}}.
\label{eq:2_prob}
\end{equation}

\noindent Here, the notation $E_{i,j}^{v,u}$ represents the exponential similarity between representations from view $v$ and view $u$, computed as:
\begin{equation}
    E_{i,j}^{v,u} = \exp(S(\mathrm{\textbf{h}}_i^v, \mathrm{\textbf{h}}_j^u)/\tau),
\label{eq:2_exp}
\end{equation}

\noindent where $\tau$ is the temperature hyperparameter, and $\mathbf{1}[\cdot]$ is the indicator function taking the value of $1$ if the condition holds, and $0$ otherwise.

\subsection{Prototype Partial Alignment}
Prototype-based IMVC methods typically learn prototypes $\mathrm{\textbf{P}}_v$ from $\mathrm{\textbf{H}}^v$ via $k$-means for missing feature reconstruction \cite{yuan2025prototype,jin2023deep,li2023incomplete}. Since these prototypes are independently generated across views, they are inherently unaligned. Existing methods address this issue by enforcing cross-view prototype alignment, implicitly assuming that prototypes from different views should share consistent semantic representations.
However, this assumption overlooks a key fact: prototypes in different views simultaneously encode both shared semantics and view-specific information. Consequently, enforcing strict alignment effectively constrains the representation space toward view-invariant semantics, suppressing view-specific components. This leads to reduced representational diversity and weakens the ability of prototypes to capture complementary cross-view information. We refer to this issue as the Prototype Over-alignment Problem (POP), which constitutes a fundamental bottleneck in prototype-based IMVC methods.



To tackle POP, we design a prototype partial alignment module, which jointly utilizes cross-view consistency and view-specific semantics. Specifically, the $i$th prototype $\mathrm{\textbf{p}}_i^v\in \mathbb{R}^d$ of view $v$ is decomposed each into two orthogonal complementary parts, namely the shared part $\mathrm{\textbf{s}}_i^{v}$ and view-specific $\mathrm{\textbf{c}}_i^{v}$ part. We separate
these two components through a learnable weight matrix $\mathrm{\textbf{U}} \in \mathbb{R}^{d \times d_c}$ satisfying 
$\mathrm{\textbf{U}}^\top \mathrm{\textbf{U}} = \mathrm{\textbf{I}}_{d_c}$, where $\mathrm{\textbf{I}}_{d_c}$ is the identity matrix 
$\in \mathbb{R}^{d_c \times d_c}$, $d_c$ is a dimension smaller than $d$. Each prototype is decomposed into:
\begin{equation}
     \mathrm{\textbf{p}}_i^v =\mathrm{\textbf{s}}_i^{v} +  \mathrm{\textbf{c}}_i^{v}
\end{equation}
where $\mathrm{\textbf{s}}_i^{v}$ and $\mathrm{\textbf{c}}_i^{v}$ are derived following:
\begin{equation}
    \mathrm{\textbf{s}}_i^{v} = \mathrm{\textbf{U}}\mathrm{\textbf{U}}^\top \mathrm{\textbf{p}}_i^{v} \in \mathbb{R}^{d}, 
    \quad
    \mathrm{\textbf{c}}_i^{v} = (\mathrm{\textbf{I}}_d - \mathrm{\textbf{U}}
    \mathrm{\textbf{U}}^\top) \mathrm{\textbf{p}}_i^{v} \in \mathbb{R}^d
    \label{eq:4}
\end{equation}

Here $\mathrm{\textbf{U}}^\top \mathrm{\textbf{U}} = \mathrm{\textbf{I}}_{d_c}$ preserves the orthogonality between $\mathrm{\textbf{s}}_i^{v}$ and $\mathrm{\textbf{c}}_i^{v}$ by:
\begin{align}
    (\mathrm{\textbf{s}}_i^{v})^\top \, \mathrm{\textbf{c}}_i^{v} &= \left(\mathrm{\textbf{U}}\mathrm{\textbf{U}}^\top \, \mathrm{\textbf{p}}_i^v\right)^\top \left((\mathrm{\textbf{I}}_d - \mathrm{\textbf{U}}\mathrm{\textbf{U}}^\top) \, \mathrm{\textbf{p}}_i^v\right) \nonumber\\
    &= (\mathrm{\textbf{p}}_i^v)^\top (\mathrm{\textbf{U}}\mathrm{\textbf{U}}^\top)^\top (\mathrm{\textbf{I}}_d - \mathrm{\textbf{U}}\mathrm{\textbf{U}}^\top) \, \mathrm{\textbf{p}}_i^v \, \nonumber\\
    &= (\mathrm{\textbf{p}}_i^v)^\top \mathrm{\textbf{U}}\mathrm{\textbf{U}}^\top (\mathrm{\textbf{I}}_d - \mathrm{\textbf{U}}\mathrm{\textbf{U}}^\top) \, \mathrm{\textbf{p}}_i^v \, \nonumber\\
    &= (\mathrm{\textbf{p}}_i^v)^\top \left(\mathrm{\textbf{U}}\mathrm{\textbf{U}}^\top - \mathrm{\textbf{U}}\mathrm{\textbf{U}}^\top \mathrm{\textbf{U}}\mathrm{\textbf{U}}^\top\right) \mathrm{\textbf{p}}_i^v \, \nonumber\\
    &= (\mathrm{\textbf{p}}_i^v)^\top \left(\mathrm{\textbf{U}}\mathrm{\textbf{U}}^\top - \mathrm{\textbf{U}}\mathrm{\textbf{I}}_{d_c}\mathrm{\textbf{U}}^\top\right) \mathrm{\textbf{p}}_i^v \, \nonumber\\
    &= (\mathrm{\textbf{p}}_i^v)^\top \left(\mathrm{\textbf{U}}\mathrm{\textbf{U}}^\top - \mathrm{\textbf{U}}\mathrm{\textbf{U}}^\top\right) \mathrm{\textbf{p}}_i^v \, \nonumber\\
    &= 0
\end{align}

To ensure that the above decomposition is well-defined, the matrix
${\textbf{U}}$ should remain column-orthonormal during optimization, such that
$\textbf{U}{\textbf{U}}^\top$ forms a valid orthogonal projection onto the shared subspace.
Instead of imposing an regularization term,
we adopt a re-parameterization strategy based on the Cayley transform\cite{jauch2020random},
which maps a skew-symmetric matrix to an orthogonal matrix. Specifically,
we introduce a free matrix $\mathrm{\textbf{W}} \in \mathbb{R}^{d \times d}$ and construct
a skew-symmetric matrix $\mathrm{\textbf{A}} = \mathrm{\textbf{W}} - \mathrm{\textbf{W}}^\top$. Then, an auxiliary orthogonal
matrix $ \mathrm{\textbf{U}}_{\text{full}} \in \mathbb{R}^{d \times d}$ is obtained at each
forward pass as
\begin{equation}
    \mathrm{\textbf{U}}_{\text{full}} = (\mathrm{\textbf{I}}_{d} - \mathrm{\textbf{A}})(\mathrm{\textbf{I}}_{d} + \mathrm{\textbf{A}})^{-1}
\end{equation}

 Since $\mathbf{A}$ is skew-symmetric, i.e., $\mathbf{A}^\top = -\mathbf{A}$, the orthogonality of
$\mathrm{\textbf{U}}_{\text{full}}$ holds by:


\begin{equation}
\begin{aligned}
    \mathrm{\textbf{U}}_{\text{full}}^\top &= \left( (\mathrm{\textbf{I}}_{d} - \mathrm{\textbf{A}})(\mathrm{\textbf{I}}_{d} + \mathrm{\textbf{A}})^{-1} \right)^\top \\
    &= (\mathrm{\textbf{I}}_{d} - \mathrm{\textbf{A}})^{-1} (\mathrm{\textbf{I}}_{d} + \mathrm{\textbf{A}}) \\[10pt]
\end{aligned}
\end{equation}    

and:
\begin{align}    
    \mathrm{\textbf{U}}_{\text{full}}^\top \mathrm{\textbf{U}}_{\text{full}} &= \left( (\mathrm{\textbf{I}}_{d} - \mathrm{\textbf{A}})^{-1} (\mathrm{\textbf{I}}_{d} + \mathrm{\textbf{A}}) \right) \left( (\mathrm{\textbf{I}}_{d} - \mathrm{\textbf{A}})(\mathrm{\textbf{I}}_{d} + \mathrm{\textbf{A}})^{-1} \right) \nonumber\\
    &= (\mathrm{\textbf{I}}_{d} - \mathrm{\textbf{A}})^{-1} \left( (\mathrm{\textbf{I}}_{d} + \mathrm{\textbf{A}})(\mathrm{\textbf{I}}_{d} - \mathrm{\textbf{A}}) \right) (\mathrm{\textbf{I}}_{d} + \mathrm{\textbf{A}})^{-1} \nonumber\\
    &= (\mathrm{\textbf{I}}_{d} - \mathrm{\textbf{A}})^{-1} \left( (\mathrm{\textbf{I}}_{d} - \mathrm{\textbf{A}})(\mathrm{\textbf{I}}_{d} + \mathrm{\textbf{A}}) \right) (\mathrm{\textbf{I}}_{d} + \mathrm{\textbf{A}})^{-1} \nonumber\\
    &= \left( (\mathrm{\textbf{I}}_{d} - \mathrm{\textbf{A}})^{-1}(\mathrm{\textbf{I}}_{d} - \mathrm{\textbf{A}}) \right) \left( (\mathrm{\textbf{I}}_{d} + \mathrm{\textbf{A}})(\mathrm{\textbf{I}}_{d} + \mathrm{\textbf{A}})^{-1} \right) \nonumber\\
    &= \mathrm{\textbf{I}}_{d} \cdot \mathrm{\textbf{I}}_{d} \nonumber\\
    &= \mathrm{\textbf{I}}_{d}
\end{align}

Finally $\mathrm{\textbf{U}}$ is derived by taking the first $d_c$ columns of $\mathrm{\textbf{U}}_{\text{full}}$, and $\mathrm{\textbf{U}}^\top\mathrm{\textbf{U}} = \mathrm{\textbf{I}}_{d_c}$ holds immediately.

After the prototypes are decomposed, to simultaneously capture cross-view consistency and view-specific semantics, we align the shared parts of the prototypes across different views while de-correlating the view-specific parts. This objective is achieved through a combination of the shared alignment loss and view-specific de-correlation loss.
\paragraph{Shared Alignment Loss}
To enforce the shared parts $\mathrm{\textbf{s}}_k^{v}$ is consistent across 
views for the same prototype-index $g$, we adopt a prototype contrastive 
objective. Define the cross-view alignment probability as:
\begin{equation}
    Q_g^{vu} = \frac{\exp(S(\mathrm{\textbf{s}}_g^{v},\, \mathrm{\textbf{s}}_g^{(u)})/\tau)}
    {\sum_{j=1}^{m} \exp(S(\mathrm{\textbf{s}}_g^{v},\, \mathrm{\textbf{s}}_j^{(u)})/\tau)}
\end{equation}
where $S(\cdot,\cdot)$ denotes cosine similarity, $\tau$ is a hyperparameter. The shared alignment 
loss is then defined as:
\begin{equation}
    \mathcal{L}_{\text{SHARED}} =  \sum_{\substack{v,u=1 \\ u \neq v}}^{m}
    \frac{1}{k}\sum_{r=1}^{t}\sum_{g=1}^{k} \frac{(1 - Q_g^{vu})^r}{r}
\end{equation}
where $t$ is a hyperparameter, $k$ is the number of prototypes for each view. This loss simultaneously 
pulls together shared components of the same class across views and pushes 
apart those of different classes.

\begin{algorithm}[t]
\caption{The training procedure of SPORT}
\label{alg:multi_view}

\textbf{Input:} Multi-view raw data $\{\mathrm{\textbf{X}}^v\}_{v=1}^m$, number of clusters $k$, 
pretraining epoch $E_{PRE}$, fine-tuning epoch $E_{FIN}$, \\
Multi-view raw complete data $\{\tilde{\mathrm{\textbf{X}}}^{v}\}_{v=1}^m$ with size $s$, Batch size $B$ \\
\textbf{Output:} Clustering results via $k$-means
\vspace{0.05in}

\begin{algorithmic}[1]
\State \textbf{Pretraining:}
\For{$t = 1,\dots,E_{PRE}$}
    \For{$i = 1,\dots,s/B$}
    \State Pretrain the autoencoder by optimizing Eq.~(\ref{eq:rec_loss})
    \EndFor
\EndFor
\State \textbf{Feature construction:}
\State Obtain features $\{\mathrm{\textbf{H}}^v\}_{v=1}^V$ from the encoders
\State \textbf{Fine-Tuning:}
\For{$t = 1,\dots,E_{FIN}$}
    \For{$i = 1,\dots,n/B$}
    \State Features are aligned by Eq.~(\ref{eq:2_main})
    \State Initial prototypes $\{\mathrm{\textbf{P}}_v\}_{v=1}^m$ via $k$-means
    \State Prototypes are decomposed by Eq.~(\ref{eq:4})
    \State Prototypes are optimized by Eq.~(\ref{eq:5})
    \State Model optimized by overall loss in Eq.~(\ref{eq:3})
    \EndFor
\EndFor
\State \textbf{Prototype imputation:}
\State Impute missing data via hybrid prototype imputation
\State \textbf{Clustering:}
\State Apply $k$-means to obtain final clustering results
\end{algorithmic}
\end{algorithm}

\paragraph{View-specific De-correlation Loss}
To allow the view-specific parts $\mathrm{\textbf{c}}_k^{v}$ capture genuinely 
distinct information across views rather than redundant features, we 
minimize the cosine alignment between view-specific parts of different views:
\begin{equation}
    \mathcal{L}_{\text{DE}} = \frac{1}{k} \sum_{g=1}^{k} 
    \sum_{\substack{v,u=1 \\ u \neq v}}^{m}
    \left| \left\langle 
    \frac{\mathrm{\textbf{c}}_g^{v}}{\|\mathrm{\textbf{c}}_g^{v}\|},\, 
    \frac{\mathrm{\textbf{c}}_g^{(v')}}{\|\mathrm{\textbf{c}}_g^{(v')}\|} 
    \right\rangle \right|
\end{equation}

Minimizing this term drives the view-specific parts of different views 
toward orthogonal directions, ensuring each view retains its own 
distinctive representation.

The overall combinatorial loss is defined as:
\begin{equation}
    \mathcal{L}_{\text{PRO}} = \lambda\mathcal{L}_{\text{SHARED}}+\eta\mathcal{L}_{\text{DE}}
    \label{eq:5}
\end{equation}
where $\lambda$ and $\eta$ are hyperparameters used to balance the shared and view-specific loss.

\subsection{Overall objective function}
To achieve our IMVC objective, we combine the losses defined above, and the overall loss function for our SPORT model is formulated as:
\begin{equation}
    \mathcal{L} = \mathcal{L}_{\text{REC}}+\alpha\mathcal{L}_{\text{CON}}+\beta\mathcal{L}_{\text{PRO}}
    \label{eq:3}
\end{equation}
where $\alpha$ and $\beta$ are hyperparameters to reconcile the contrastive loss and the prototype partial alignment loss.

The training of our SPORT model consists of two phases as shown in Algorithm.{(\ref{alg:multi_view})}. In the pretraining phase, only the reconstruction loss defined in Eq.{(\ref{eq:rec_loss})} is used to train the autoencoder. After the pretraining phase, we enter a fine-tuning phase where the model uses the overall loss defined in Eq.{(\ref{eq:3})} to better align the features and prototypes across different views. The missing data is then padded via a hybrid prototype imputation strategy discussed in the next subsection. Finally, $k$-means is performed on the imputed complete data to obtain clustering results.

\begin{figure}
    \centering
    \includegraphics[width=1\linewidth]{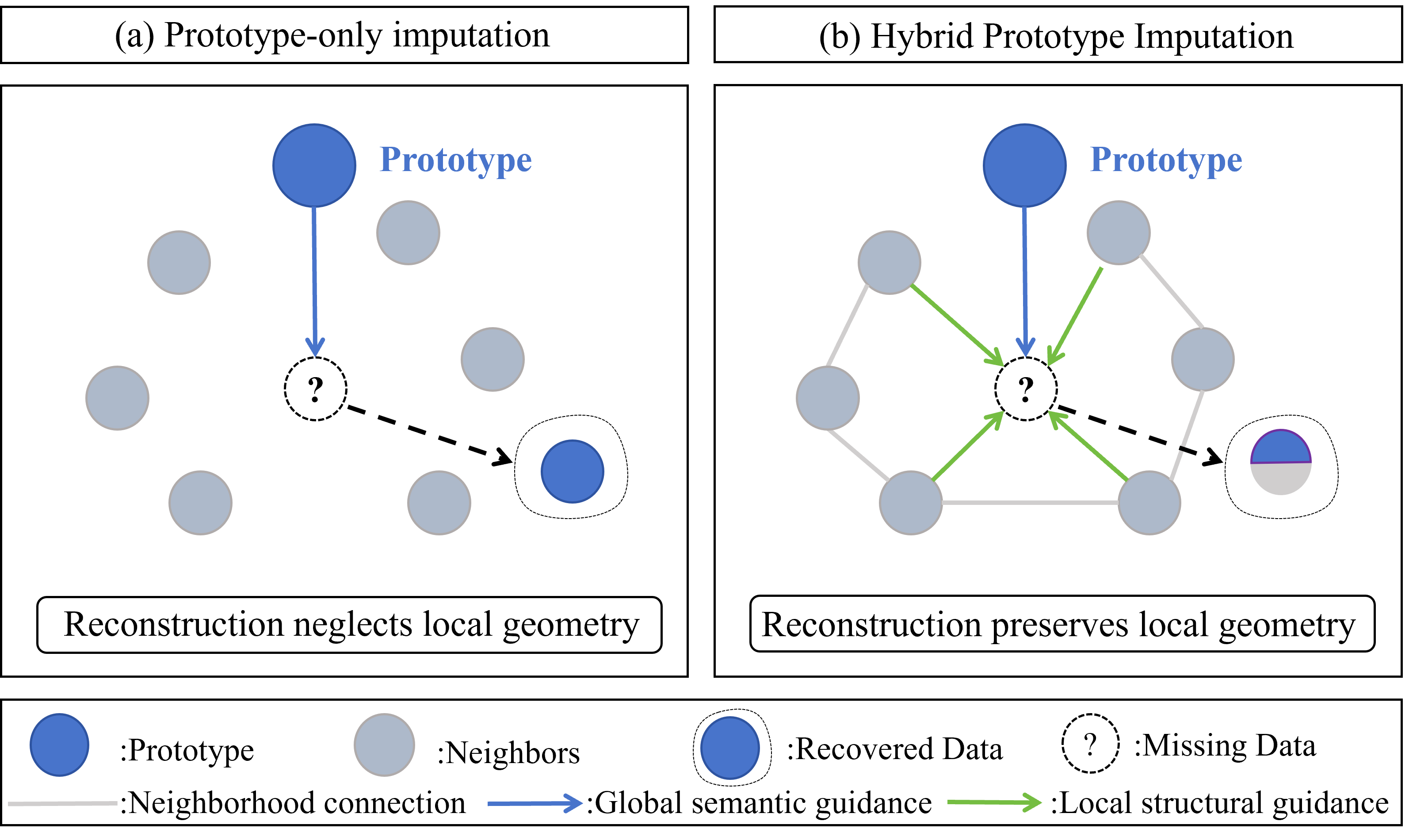}
    \caption{Illustration of the proposed hybrid prototype imputation strategy. (a) Existing methods rely solely on the most relevant prototype for missing-view reconstruction, primarily capturing global semantics while neglecting local neighborhood information. (b) Our method jointly leverages prototype-level global semantics and neighbor-level local semantics to perform more robust missing feature imputation.}
    \label{fig:hyrid}
\end{figure}

\subsection{Hybrid Prototype Imputation Strategy}
After prototype alignment, these prototypes are employed as semantic anchors for missing-view reconstruction. Existing methods \cite{yuan2025prototype,jin2023deep,li2023incomplete} typically rely solely on the most relevant prototypes to recover missing features, as illustrated in Fig. \ref{fig:hyrid}(a). Such approaches mainly exploit global shared information encoded in prototypes, while neglecting the local semantics captured by neighboring instances. Consequently, their performance often degrades under high missing rates or when unreliable prototypes are generated.

To mitigate this drawback, we propose a novel hybrid prototype imputation strategy which utilizes both prototypes and neighboring information as shown in Figure.\ref{fig:hyrid}(b). Specifically, for a missing data in view $v$ of sample $j$, we first calculate the average cosine similarity between sample $j$ and all prototypes in views where sample $j$ is complete. Then, for all other samples who have complete data in view $v$, we calculate the average similarity between them and sample $j$ in views where both samples have complete features. In addition following Eq.(\ref{eq:impute_proto}) and Eq.(\ref{eq:impute_neighbor}), we match the prototype and sample with the largest average cosine similarity with sample $j$, and denote their features in view $v$ as $\hat{\mathrm{\textbf{p}}}_{j}^v$ and $\hat{\mathrm{\textbf{n}}}_{j}^v$ respectively.
Lastly, the missing information of sample $j$ in view $v$ is imputed by a weighted sum of $\hat{\mathrm{\textbf{p}}}_{j}^v$ and $\hat{\mathrm{\textbf{n}}}_{j}^v$  as Eq.(\ref{eq:hybrid_imputation}), where the imputed feature is denoted by $\hat{\mathrm{\textbf{h}}}_j^v$, and the weights are $\gamma$ and $1 - \gamma$ with $\gamma \in[0,1]$. 
\begin{align}
    \hat{\mathrm{\textbf{p}}}_{j}^v &= \arg\max_{\mathrm{\textbf{p}}_k^v \in \mathrm{\textbf{P}}_v} 
    \frac{1}{|\mathcal{V}_j|} \sum_{v\in\mathcal{V}_j} S(\mathrm{\textbf{p}}_k^v, \mathrm{\textbf{h}}_j^v)
    \label{eq:impute_proto} \\
    \hat{\mathrm{\textbf{n}}}_{j}^v &= \arg\max_{\mathrm{\textbf{h}}_i^v \in \tilde{\mathrm{\textbf{H}}}^{v}} 
    \frac{1}{|\mathcal{V}_j \cap \mathcal{V}_i|} \sum_{u \in \mathcal{V}_j \cap \mathcal{V}_i} S(\mathrm{\textbf{h
    }}_i^u, \mathrm{\textbf{h}}_j^u) \label{eq:impute_neighbor} \\
    \hat{\mathrm{\textbf{h}}}_j^v &= \gamma \cdot \hat{\mathrm{\textbf{p}}}_{j}^v + (1 - \gamma) \cdot \hat{\mathrm{\textbf{n}}}_{j}^v, 
    \quad \gamma \in [0, 1] \label{eq:hybrid_imputation}
\end{align}

By exploiting both prototype- and neighborhood-based information for missing data imputation, our method achieves stable clustering performance even under increasing missing rates, as demonstrated in the subsequent experiments.

\section{Experiments}
In this section, we demonstrate the superior performance of our proposed SPORT model through extensive evaluations, including experimental settings, performance comparisons,  ablation study, sensitive analysis and clustering visualization.

\subsection{Experimental Settings}

\textbf{Benchmark Datasets}: We construct comprehensive experiments on 6 benchmark multi-view datasets namely 
ALOI\_100\footnote{\url{https://github.com/JethroJames/Awesome-Multi-View-Learning-Datasets}}, 
Animal\footnotemark[\value{footnote}], 
Digit4k\footnotemark[\value{footnote}], 
100Leaves\footnotemark[\value{footnote}], 
Reuters\_21578\footnotemark[\value{footnote}], 
and VGGFace2\_50\footnotemark[\value{footnote}]. The detailed statistics of these datasets are shown in TABLE \ref{table:3}. 

\begin{table}[H]  
\centering
\caption{Statistics of 6 Benchmark Datasets.}

\resizebox{\columnwidth}{!}{
    \begin{tabular}{@{} l c c c c @{}}
    \toprule
    \textbf{Dataset} & \textbf{Samples ($n$)} & \textbf{Views ($v$)} & \textbf{Clusters ($k$)} & \textbf{Dimensionality($d_v$)} \\
    \midrule
    Reuters\_21578 & 1500  & 5 & 6   & 21531/24892/34251/15506/11547 \\
    100Leaves      & 1600  & 3 & 100 & 64/64/64 \\
    Digit4k        & 4000  & 2 & 4   & 30/30 \\
    Animal         & 10158 & 2 & 50  & 4096/4096 \\
    ALOI\_100      & 10800 & 4 & 100 & 77/13/64/125 \\
    VGGFace2\_50   & 34027 & 4 & 50  & 944/576/512/640 \\
    \bottomrule
    \end{tabular}
}
\label{table:3}
\end{table}
\textbf{Evaluating Metrics}: We used three widely-used evaluation criteria to demonstrate the validity of the experiment. These metrics are accuracy(ACC), Adjusted Rand Index(ARI) and F-Score(F-Sco).

\begin{table*}[!t]
\centering
\caption{Performance Comparison on Animal, ALOI\_100 and Digit4k Under
Different Missing rates. The Optimal Results are In \textbf{Bold} and The Second-best
Results are \underline{Underlined}. O/M Represents Out of Memory.}
\label{tab:1}
\setlength{\tabcolsep}{7pt}
\renewcommand{\arraystretch}{1.05}
\resizebox{\textwidth}{!}{%
\begin{tabular}{l l  c c c  c c c  c c c  c c c}
\toprule
\multicolumn{2}{c}{Missing Rate}
  & \multicolumn{3}{c}{0.1}
  & \multicolumn{3}{c}{0.3}
  & \multicolumn{3}{c}{0.5}
  & \multicolumn{3}{c}{0.7} \\
\cmidrule{3-5}\cmidrule{6-8}\cmidrule{9-11}\cmidrule{12-14}
\multicolumn{2}{c}{Metric}
  & ACC & ARI & F-score
  & ACC & ARI & F-score
  & ACC & ARI & F-score
  & ACC & ARI & F-score \\
\midrule
 
\multirow{15}{*}{\rotatebox{90}{Animal}}
  & CDIMC-net(IJCAI'21) \cite{wen2021structural}  &  5.94 &  0.66 &  0.60 & 3.76 &  -0.01 &  0.72 & 3.90 & -0.01 & 0.77 & 3.90 &  0.00 & 0.77 \\
  & COMPLETER~(CVPR'21)  \cite{lin2021completer}   & 21.29 & 15.40 & 15.84 & 19.67 & 13.42 & 13.12 & 20.37 & 15.59 & 14.22 & 20.93 & 11.52 & 14.84 \\
  & SURE(TPAMI'22) \cite{yang2022robust}      & 35.67 & 23.00 & 32.72 & 31.42 & 20.66 & 27.69 & 27.46 & 15.90 & 24.92 & 25.18 & 13.47 & 21.98 \\
  & ProImp (IJCAI'23) \cite{li2023incomplete}    & 41.40 & 30.05 & 38.02 & 38.51 & 27.47 & 34.63 & 38.14 & 26.90 & 34.52 & 37.65 & 26.06 & 34.44 \\
  & APADC (TIP'23) \cite{xu2023adaptive}       & 22.73 &  5.39 & 20.78 & 34.61 & 21.69 & 31.91 & 32.50 & 21.23 & 27.25 & 23.60 & 12.70 & 16.41 \\
  & IMVC-IE (ICASSP'24) \cite{huang2024incomplete}   & 17.55 &  2.96 & 18.82 & 11.36 &  0.37 & 12.23 &  7.95 &  0.05 &  7.60 &  8.26 &  1.44 &  6.92 \\
  & RPCIC (ACM MM'24) \cite{yuan2024robust}     & 43.84 & 31.05 & 45.41 & 46.05 & 34.39 & 47.42 & 44.93 & 32.40 & 46.42 & 35.64 & 23.27 & 36.44 \\
  & BURG (ICCV'25) \cite{jin2025deep}        & 60.82 & 50.40 & \underline{63.81} & \underline{50.57} & 38.36 & \underline{54.34} & \underline{49.67} & \underline{36.18} & \underline{54.03} & \underline{39.46} & \underline{25.97} & \underline{42.97} \\
  & BRIDGE (ICCV'25) \cite{jiang2025unified}      & 20.12 & 14.60 & 17.46 & 18.80 & 13.16 & 15.92 & 17.54 & 11.94 & 15.75 & 27.58 & 17.76 & 25.74 \\
  & HSACC (NeurIPS'25) \cite{ding2025incomplete1}   & 20.29 & 12.98 & 16.82 & 24.03 & 17.63 & 17.55 & 24.77 & 18.89 & 20.10 & 23.02 & 16.33 & 17.60 \\
  & IMC-MCL (TKDE'25) \cite{yin2025incomplete}    & 25.98 & 17.17 & 15.39 & 24.63 & 16.82 &  0.41 & 25.35 & 16.17 &  0.49 & 26.45 & 18.16 & 16.36 \\
  & MGCCFF(AAAI'25) \cite{zhao2025incomplete}      & 11.69 &  4.45 &  9.70 & 11.53 &  4.43 &  9.40 & 10.92 &  3.86 &  9.13 & 10.81 &  3.87 &  8.46 \\
    & DGIMVCM (AAAI'26) \cite{zhang2026dynamic}  & \underline{62.45} & \underline{52.07} & 59.09 & 49.10 & \underline{40.75} & 44.72 & 45.27 & 34.44 & 41.50 & 36.10 & 23.76 & 35.53 \\
      & GMAE (TPAMI'26) \cite{11494292}  & 56.22 & 43.41 & 47.78 & 29.53 & 15.92 & 16.20 & 14.60 & 7.40 & 6.32 & 20.55 & 12.56 & 10.97 \\
\midrule
\rowcolor{gray!10}  & \textbf{SPORT (Ours)} & \textbf{64.71} & \textbf{54.80} & \textbf{64.39} & \textbf{55.98} & \textbf{46.78} & \textbf{54.90} & \textbf{58.48} & \textbf{45.88} & \textbf{57.57} & \textbf{51.66} & \textbf{39.47} & \textbf{51.20} \\
\midrule
 
\multirow{15}{*}{\rotatebox{90}{ALOI\_100}}
  & CDIMC-net(IJCAI'21) \cite{wen2021structural}  & 62.02 & 48.70 & 57.36 & 37.92 & 18.34 & 38.04 & 21.52 &  7.85 & 21.85 & 13.48 &  3.77 & 14.05 \\
  & COMPLETER~(CVPR'21)  \cite{lin2021completer}   & 20.83 & 11.77 & 20.12 & 20.28 & 12.02 & 19.29 & 16.96 &  9.83 & 14.39 & 18.94 & 10.68 & 18.28 \\
  & SURE(TPAMI'22) \cite{yang2022robust}      & 61.75 & 52.36 & 60.64 & 63.24 & 53.62 & 62.20 & 56.37 & \underline{46.45} & 54.52 & 57.09 & 46.53 & 55.45 \\
  & ProImp (IJCAI'23) \cite{li2023incomplete}    & 45.67 & 33.92 & 44.09 & 48.06 & 35.24 & 46.86 & 43.44 & 31.32 & 42.42 & 42.16 & 29.24 & 41.12 \\
  & APADC (TIP'23) \cite{xu2023adaptive}       & 26.55 & 10.12 & 20.59 & 16.41 &  8.86 & 10.50 & 16.41 &  7.32 & 11.70 & 22.35 & 10.14 & 19.31 \\
  & IMVC-IE (ICASSP'24) \cite{huang2024incomplete}   & 35.92 & 23.28 & 33.07 & 30.04 & 18.13 & 27.47 & 26.64 & 16.48 & 25.08 & 23.23 & 13.27 & 21.61 \\
  & RPCIC (ACM MM'24) \cite{yuan2024robust}     & 74.99 & 64.99 & 73.20 & \underline{71.35} & \textbf{63.07} & \underline{68.91} & 55.36 & 45.21 & \underline{56.03} & \underline{63.23} & \underline{54.50} & \underline{61.43} \\
  & BURG (ICCV'25) \cite{jin2025deep}        & 61.80 & 48.14 & 60.02 & 50.19 & 29.84 & 51.06 & 41.10 & 19.06 & 43.49 & 29.94 & 14.34 & 32.29 \\
  & BRIDGE (ICCV'25) \cite{jiang2025unified}      & \textbf{78.63} & \textbf{72.04} & \textbf{75.47} & 70.30 & 60.89 & 66.75 & \underline{56.72} & 43.21 & 53.49 & 32.61 & 18.73 & 31.89 \\
  & HSACC (NeurIPS'25) \cite{ding2025incomplete1}   & 20.19 & 12.56 & 18.42 & 20.30 & 11.74 & 19.50 & 19.94 & 11.20 & 19.21 & 19.10 & 11.63 & 17.76 \\
  & IMC-MCL (TKDE'25) \cite{yin2025incomplete}    & 17.47 & 10.36 &  0.90 & 17.76 & 11.38 &  0.91 & 15.44 &  9.46 &  0.65 & 16.20 &  9.89 &  0.44 \\
  & MGCCFF(AAAI'25) \cite{zhao2025incomplete}      & O/M  & O/M  & O/M  & O/M  & O/M  & O/M  & O/M  & O/M  & O/M  & O/M  & O/M  & O/M  \\
    & DGIMVCM (AAAI'26) \cite{zhang2026dynamic}  & O/M & O/M & O/M & O/M & O/M & O/M & O/M & O/M & O/M & O/M & O/M & O/M \\
      & GMAE (TPAMI'26) \cite{11494292}  & O/M & O/M & O/M & O/M & O/M & O/M & O/M & O/M & O/M & O/M & O/M & O/M \\
\midrule
\rowcolor{gray!10}  & \textbf{SPORT (Ours)} & \underline{77.15} & \underline{69.05} & \underline{74.18} & \textbf{72.93} & \underline{62.87} & \textbf{70.27} & \textbf{74.56} & \textbf{64.04} & \textbf{71.32} & \textbf{70.51} & \textbf{57.88} & \textbf{67.27} \\
\midrule
 
\multirow{15}{*}{\rotatebox{90}{Digit4k}}
  & CDIMC-net(IJCAI'21) \cite{wen2021structural}  & 34.10 &  2.91 & 21.57 & 46.02 & 24.47 & 27.31 & 33.25 &  1.84 & 25.46 & 57.65 & 34.42 & 51.13 \\
  & COMPLETER~(CVPR'21)  \cite{lin2021completer}   & 30.10 &  0.86 & 19.85 & 29.20 &  0.41 & 18.30 & 30.12 &  0.73 & 19.90 & 32.28 &  2.16 & 23.35 \\
  & SURE(TPAMI'22) \cite{yang2022robust}      & 64.30 & 33.29 & 64.61 & 46.78 & 24.15 & 46.36 & 48.38 & 18.59 & 47.36 & 42.55 & 13.69 & 41.37 \\
  & ProImp (IJCAI'23) \cite{li2023incomplete}    & 61.96 & 35.17 & 62.22 & 53.10 & 21.06 & 53.30 & 78.57 & 52.51 & \underline{78.51} & 62.32 & 29.83 & 62.86 \\
  & APADC (TIP'23) \cite{xu2023adaptive}       & 38.48 &  5.86 & 30.67 & 30.70 &  5.89 & 20.43 & 36.78 &  9.39 & 27.55 & 34.82 &  7.26 & 25.47 \\
  & IMVC-IE (ICASSP'24) \cite{huang2024incomplete}   & \underline{82.05} & \underline{60.85} & 60.80 & 62.72 & 31.13 & 46.10 & 56.50 & 20.68 & 44.36 & 57.45 & 22.63 & 44.70 \\
  & RPCIC (ACM MM'24) \cite{yuan2024robust}     & 80.60 & 57.70 & \textbf{80.34} & 55.60 & 26.40 & 56.39 & 39.92 & 18.77 & 39.90 & 42.58 & 14.25 & 43.89 \\
  & BURG (ICCV'25) \cite{jin2025deep}        & 69.45 & 38.68 & \underline{69.89} & 68.85 & 27.49 & \textbf{69.35} & \underline{79.35} & \underline{56.68} & \textbf{78.61} & 53.72 & 35.71 & 50.60 \\
  & BRIDGE (ICCV'25) \cite{jiang2025unified}      & 47.17 & 16.83 & 36.89 & 71.48 & 45.76 & 53.03 & 68.14 & 46.00 & 49.26 & \underline{68.04} & \underline{44.02} & 49.74 \\
  & HSACC (NeurIPS'25) \cite{ding2025incomplete1}   & 35.45 &  4.42 & 28.63 & 35.22 &  3.95 & 28.26 & 35.20 &  6.71 & 28.17 & 37.08 &  6.46 & 30.73 \\
  & IMC-MCL (TKDE'25) \cite{yin2025incomplete}    & 73.85 & 40.04 & 54.50 & \textbf{81.58} & \underline{59.56} & 59.18 & 64.62 & 32.49 & 48.02 & 61.88 & 30.17 & 48.93 \\
  & MGCCFF(AAAI'25) \cite{zhao2025incomplete}      & 64.84 & 40.80 & 64.31 & 61.49 & 39.43 & 61.01 & 65.05 & 38.94 & 64.27 & 62.37 & 35.82 & 61.28 \\
    & DGIMVCM (AAAI'26) \cite{zhang2026dynamic}  & 64.18 & 56.35 & 59.66 & 70.18 & \textbf{60.80} & 60.02 & 63.58 & 43.07 & 63.56 & 63.80 & 41.66 & \underline{63.68} \\
      & GMAE (TPAMI'26) \cite{11494292}   & 65.45 & 57.50 & 61.74 & 66.15 & 28.99 & 61.28 & 59.62 & 46.66 & 56.18 & 55.70 & 37.42 & 52.43\\
\midrule
\rowcolor{gray!10}  & \textbf{SPORT (Ours)} & \textbf{85.70} & \textbf{67.06} & 65.14 & \underline{81.22} & 58.41 & \underline{62.58} & \textbf{83.95} & \textbf{63.52} & 63.90 & \textbf{84.12} & \textbf{64.40} & \textbf{63.92} \\
\bottomrule
\end{tabular}}
\end{table*}

\begin{table*}[!t]
\centering
\caption{Performance Comparison on 100Leaves, Reuters\_21578 and VGGFace2\_50 Under
Different Missing rates. The Optimal Results are In \textbf{Bold} and The Second-best
Results are \underline{Underlined}. O/M Represents Out of Memory.}
\label{tab:2}
\setlength{\tabcolsep}{7pt}
\renewcommand{\arraystretch}{1.05}
\resizebox{\textwidth}{!}{%
\begin{tabular}{l l  c c c  c c c  c c c  c c c}
\toprule
\multicolumn{2}{c}{Missing Rate}
  & \multicolumn{3}{c}{0.1}
  & \multicolumn{3}{c}{0.3}
  & \multicolumn{3}{c}{0.5}
  & \multicolumn{3}{c}{0.7} \\
\cmidrule(lr){3-5}\cmidrule(lr){6-8}\cmidrule(lr){9-11}\cmidrule(l){12-14}
\multicolumn{2}{c}{Metric}
  & ACC & ARI & F-score
  & ACC & ARI & F-score
  & ACC & ARI & F-score
  & ACC & ARI & F-score \\
\midrule
 
\multirow{15}{*}{\rotatebox{90}{100Leaves}}
  & CDIMC-net(IJCAI'21) \cite{wen2021structural}  & 70.75 & 61.18 & 64.96 & 47.56 & \underline{38.38} & 43.82 & 31.37 & 20.27 & 29.35 & 24.88 & 11.27 & 23.95 \\
  & COMPLETER~(CVPR'21)  \cite{lin2021completer}   & 29.38 & 16.97 & 27.13 & 26.44 & 14.10 & 24.21 & 37.44 & 21.27 & 37.81 & 29.06 & 15.54 & 28.07 \\
  & SURE(TPAMI'22) \cite{yang2022robust}      & 56.62 & 41.57 & 53.66 & 51.38 & 37.40 & 48.84 & 48.00 & 35.09 & 45.19 & 37.62 & 24.15 & 36.05 \\
  & ProImp (IJCAI'23) \cite{li2023incomplete}    & 63.44 & 56.51 & 58.70 & 48.62 & 37.08 & 45.33 & 45.94 & 34.29 & 43.37 & 41.94 & 27.43 & 41.07 \\
  & APADC (TIP'23) \cite{xu2023adaptive}       & 24.06 & 12.29 & 19.01 & 18.75 &  8.12 & 15.32 & 20.81 &  8.00 & 18.06 & 19.31 &  8.18 & 16.37 \\
  & IMVC-IE (ICASSP'24) \cite{huang2024incomplete}   & 39.38 & 24.22 & 38.86 & 29.19 & 13.76 & 28.67 & 22.75 &  7.92 & 23.10 & 15.44 &  3.11 & 15.28 \\
  & RPCIC (ACM MM'24) \cite{yuan2024robust}     & 74.50 & 62.36 & 74.70 & 59.25 & 38.22 & 61.77 & 47.06 & 30.51 & 48.42 & 45.56 & 31.63 & 44.40 \\
  & BURG (ICCV'25) \cite{jin2025deep}        & 71.19 & 58.87 & 69.12 & 57.06 & 37.48 & 56.85 & 45.38 & 21.17 & 47.49 & 34.62 & 14.42 & 36.58 \\
  & BRIDGE (ICCV'25) \cite{jiang2025unified}      & 49.30 & 35.14 & 47.68 & 29.07 & 13.63 & 28.67 & 26.29 &  9.57 & 26.53 & 28.66 & 11.95 & 28.66 \\
  & HSACC (NeurIPS'25) \cite{ding2025incomplete1}   & 27.31 & 15.61 & 24.49 & 29.38 & 18.24 & 25.82 & 29.38 & 18.14 & 26.49 & 31.00 & 16.15 & 28.90 \\
  & IMC-MCL (TKDE'25) \cite{yin2025incomplete}    &  9.00 &  7.87 &  0.07 &  7.12 &  5.54 &  2.07 &  6.62 &  4.37 &  0.06 &  6.75 &  3.40 &  0.18 \\
  & MGCCFF(AAAI'25) \cite{zhao2025incomplete}      &  6.56 &  3.88 &  0.21 &  9.80 &  5.88 &  0.41 & 10.66 &  5.09 &  0.24 & 10.84 &  6.72 &  4.72 \\
    & DGIMVCM (AAAI'26) \cite{zhang2026dynamic} & \textbf{85.21} & \textbf{81.28} & \textbf{83.76} & \textbf{72.56} & \textbf{64.68} & \underline{70.14} & \underline{63.50} & \underline{53.01} & \underline{61.11} & \underline{48.88} & \underline{36.25} & \underline{46.71} \\
  & GMAE (TPAMI'26) \cite{11494292}  & 67.81 & 56.83 & 64.96 & 42.44 & 23.94 & 42.06 & 37.56 & 26.55 & 35.71& 35.19 & 20.11 & 34.82 \\
\midrule
\rowcolor{gray!10}   & \textbf{SPORT (Ours)} & \underline{80.62} & \underline{72.95} & \underline{78.71} & \underline{74.06} & \underline{62.31} & \textbf{71.18} & \textbf{69.56} & \textbf{54.35} & \textbf{67.47} & \textbf{56.06} & \textbf{36.68} & \textbf{54.24} \\
\midrule
 
\multirow{15}{*}{\rotatebox{90}{Reuters\_21578}}
  & CDIMC-net(IJCAI'21) \cite{wen2021structural} & O/M  & O/M  & O/M  & O/M  & O/M  & O/M  & O/M  & O/M  & O/M  & O/M  & O/M  & O/M  \\
  & COMPLETER~(CVPR'21)  \cite{lin2021completer}   & 33.87 &  1.03 & 18.64 & 30.93 &  1.99 & 17.31 & 34.40 &  1.27 & 18.92 & 32.40 &  1.91 & 19.71 \\
  & SURE(TPAMI'22) \cite{yang2022robust}      &  32.27   & 9.36     & 28.75    &  31.40   & 7.68    &  29.01   & 29.80 &  4.00 & 23.90 & 29.33 &  4.64 & 25.29 \\
  & ProImp (IJCAI'23) \cite{li2023incomplete}    & 40.87 & 16.06 & 38.46 & 41.13 & 17.09 & 28.34 & 31.00 &  9.49 & 29.28 & 35.33 & 10.58 & 31.42 \\
  & APADC (TIP'23) \cite{xu2023adaptive}       & 28.07 & 18.30 & 41.09 & 27.40 &  7.29 & 29.05 & 28.53 & 18.83 & \underline{47.69} & 29.93 &  9.84 & 38.89 \\
  & IMVC-IE (ICASSP'24) \cite{huang2024incomplete}   &  3.93 & 18.95 & 47.43 &  7.33 & 12.10 & 39.40 &  6.80 & 11.21 & 39.11 &  3.93 &  8.29 & 32.06 \\
  & RPCIC (ACM MM'24) \cite{yuan2024robust}     & \underline{46.00} & 18.38 & 46.40 & \underline{46.93} & 17.54 & \underline{46.54} & \underline{46.13} & 17.77 & 44.78 & \textbf{48.07} & \underline{20.32} & \textbf{46.16} \\
  & BURG (ICCV'25) \cite{jin2025deep}        & 45.93 & 11.81 & 35.53 & 44.20 & 10.08 & 35.28 & 27.67 & $-$0.05 & 12.28 & \underline{47.07} & 14.33 & \underline{42.47} \\
  & BRIDGE (ICCV'25) \cite{jiang2025unified}      &  5.40 & \textbf{25.94} & 44.69 &  6.39 & \textbf{23.57} & 46.43 &  7.58 & \underline{23.39} & 44.71 &  2.77 &  8.08 & 35.69 \\
  & HSACC (NeurIPS'25) \cite{ding2025incomplete1}   & 38.20 &  5.14 & 31.71 & 34.93 &  1.59 & 30.04 & 37.20 &  4.69 & 31.83 & 32.93 &  2.94 & 24.51 \\
  & IMC-MCL (TKDE'25) \cite{yin2025incomplete}    & 33.67 & 11.72 & 37.56 & 38.27 & 11.85 & 41.65 & 29.67 &  7.27 & 31.63 & 35.33 &  7.07 & 32.39 \\
  & MGCCFF(AAAI'25) \cite{zhao2025incomplete}      & 44.73 & 16.87 & 39.15 & 46.79 & 22.50 & 46.09 & 45.81 & 17.21 & 40.26 & 46.53 & \textbf{21.86} & \textbf{46.16} \\
    & DGIMVCM (AAAI'26) \cite{zhang2026dynamic}  & 40.60 & 9.03 & 20.62 & 38.20 & 6.14 & 21.39 & 32.87 & 5.37 & 22.39 & 29.00 & 3.29 & 20.31 \\
      & GMAE (TPAMI'26) \cite{11494292}  & 34.00 & 2.06 & 20.60 & 31.80 & 0.07 & 14.43 & 25.87 & 2.40 & 19.89 & 27.13 & 1.57 & 23.98 \\
\midrule
\rowcolor{gray!10}  & \textbf{SPORT (Ours)} & \textbf{46.33} & \underline{21.73} & \underline{48.27} & \textbf{51.00} & \underline{23.45} & \textbf{51.73} & \textbf{53.47} & \textbf{26.44} & \textbf{51.39} & 38.13 & 15.87 & 39.20 \\
\midrule
 
\multirow{15}{*}{\rotatebox{90}{VGGFace2\_50}}
  & CDIMC-net(IJCAI'21) \cite{wen2021structural}  & O/M  & O/M  & O/M  & O/M  & O/M  & O/M  & O/M  & O/M  & O/M  & O/M  & O/M  & O/M  \\
  & COMPLETER~(CVPR'21)  \cite{lin2021completer}   &  4.72 &  0.73 &  4.10 &  5.27 &  0.94 &  3.74 &  4.70 &  0.74 &  3.91 &  4.76 &  0.82 &  4.05 \\
  & SURE(TPAMI'22) \cite{yang2022robust}      & O/M  & O/M  & O/M  & O/M  & O/M  & O/M  & O/M  & O/M  & O/M  & O/M  & O/M  & O/M  \\
  & ProImp (IJCAI'23) \cite{li2023incomplete}    &  7.62 &  1.48 &  7.57 &  7.15 &  1.37 &  7.05 &  8.30 &  1.66 &  8.19 &  7.18 &  1.41 &  7.13 \\
  & APADC (TIP'23) \cite{xu2023adaptive}       &  5.42 &  0.94 &  4.02 &  5.58 &  0.92 &  3.84 &  5.24 &  0.84 &  3.21 &  5.62 &  1.12 &  2.87 \\
  & IMVC-IE (ICASSP'24) \cite{huang2024incomplete}   &  6.82 &  1.23 &  6.43 &  5.44 &  0.63 &  5.32 &  4.79 &  0.38 &  4.76 &  4.62 &  0.38 &  4.20 \\
  & RPCIC (ACM MM'24) \cite{yuan2024robust}     &  8.21 &  1.92 &  8.19 & \underline{9.62} & \underline{2.53} & \underline{9.62} & \underline{9.96} & \underline{2.58} & \underline{10.03} & \textbf{10.31} & \underline{2.82} & \textbf{10.46} \\
  & BURG (ICCV'25) \cite{jin2025deep}        & \underline{10.04} & \underline{2.90} & \underline{9.96} &  9.40 &  1.82 &  9.52 &  8.11 &  1.44 &  8.26 &  5.79 &  0.78 &  5.81 \\
  & BRIDGE (ICCV'25) \cite{jiang2025unified}      &  5.03 &  0.78 &  4.82 &  5.49 &  1.25 &  4.78 &  5.60 &  0.79 &  5.61 &  5.53 &  0.88 &  5.43 \\
  & HSACC (NeurIPS'25) \cite{ding2025incomplete1}   &  5.39 &  1.06 &  4.21 &  5.47 &  0.97 &  4.04 &  5.33 &  0.94 &  4.08 &  5.51 &  0.92 &  4.10 \\
  & IMC-MCL (TKDE'25) \cite{yin2025incomplete}    &  5.05 &  0.73 &  3.56 &  4.93 &  0.63 &  3.89 &  4.64 &  0.59 &  2.66 &  5.00 &  0.75 &  3.54 \\
  & MGCCFF(AAAI'25) \cite{zhao2025incomplete}      & O/M  & O/M  & O/M  & O/M  & O/M  & O/M  & O/M  & O/M  & O/M  & O/M  & O/M  & O/M  \\
    & DGIMVCM (AAAI'26) \cite{zhang2026dynamic}  & O/M & O/M & O/M & O/M & O/M & O/M & O/M & O/M & O/M & O/M & O/M & O/M \\
      & GMAE (TPAMI'26) \cite{11494292}  & O/M & O/M & O/M & O/M & O/M & O/M & O/M & O/M & O/M & O/M & O/M & O/M \\
\midrule
\rowcolor{gray!10}  & \textbf{SPORT (Ours)} & \textbf{10.42} & \textbf{3.05} & \textbf{10.23} & \textbf{10.41} & \textbf{3.01} & \textbf{10.33} & \textbf{11.29} & \textbf{3.17} & \textbf{11.23} & \underline{10.24} & \textbf{2.92} & \underline{9.82} \\
\bottomrule
\end{tabular}}
\end{table*}

\begin{figure*}[!t]
    \centering
    \begin{minipage}{0.32\textwidth}
        \centering
        \includegraphics[width=\linewidth]{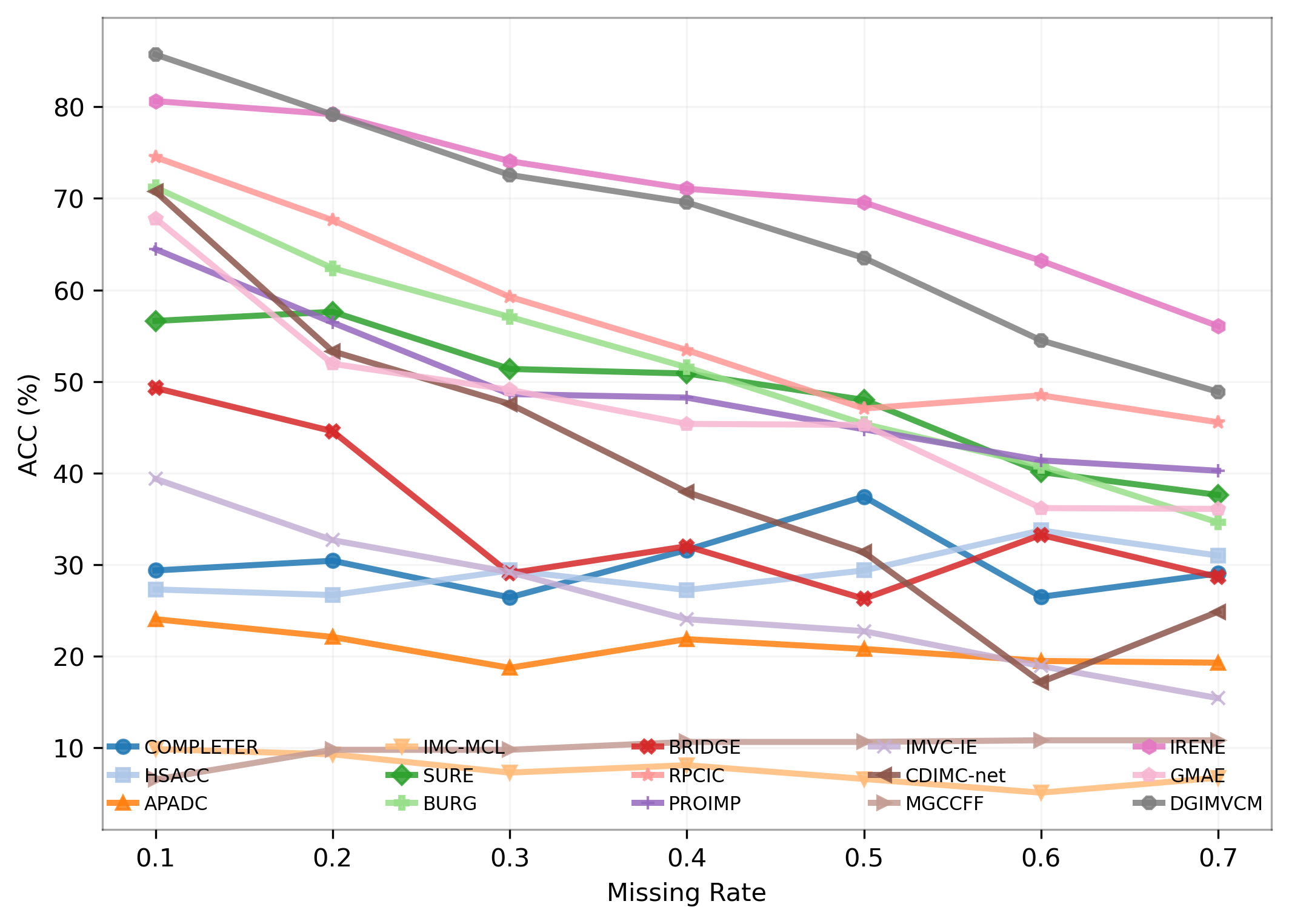}
        \small (a) 100Leaves
    \end{minipage}\hfill
    \begin{minipage}{0.32\textwidth}
        \centering
        \includegraphics[width=\linewidth]{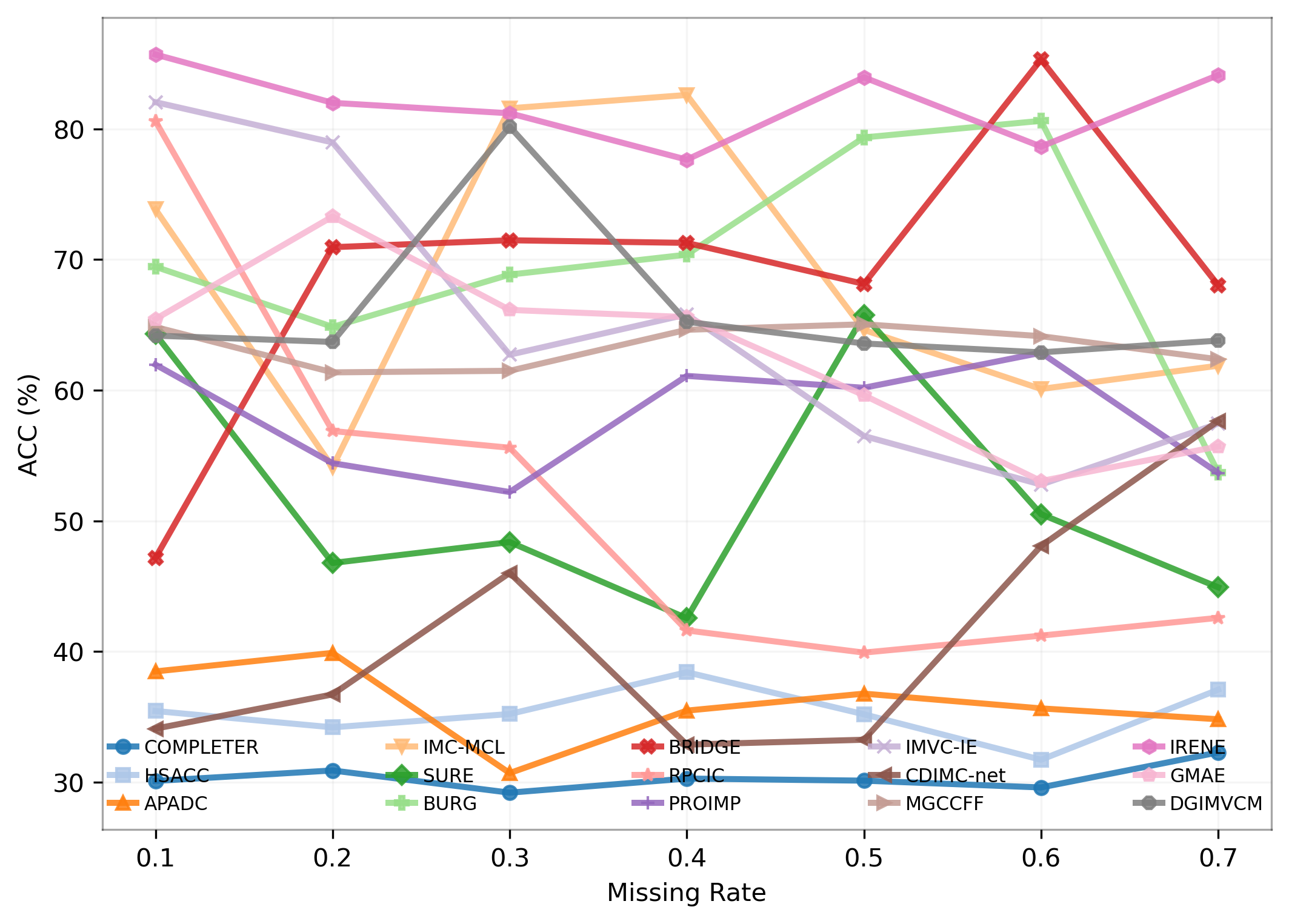}
        \small (b) Digit4k
    \end{minipage}\hfill
    \begin{minipage}{0.32\textwidth}
        \centering
        \includegraphics[width=\linewidth]{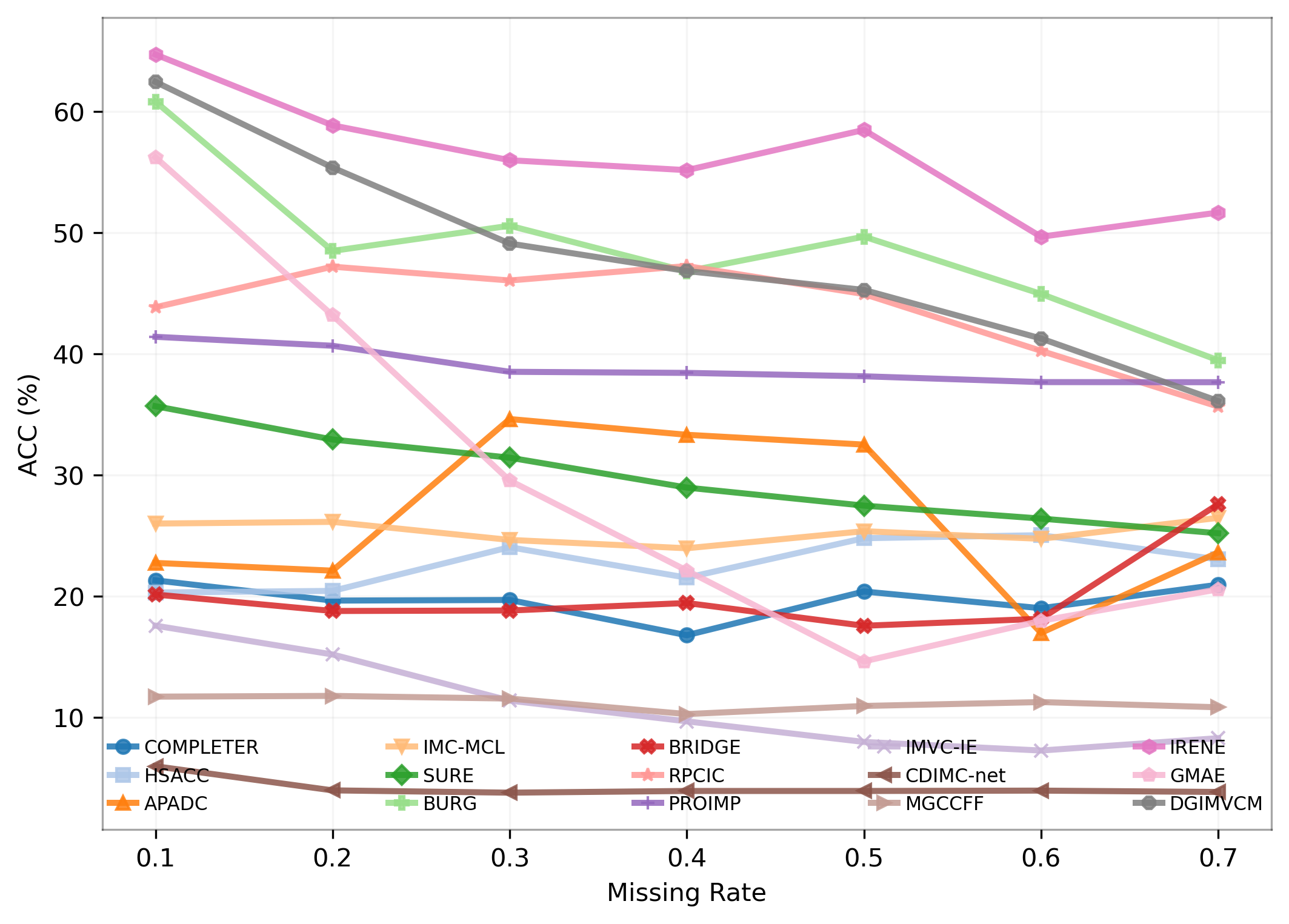}
        \small (c) Animal
    \end{minipage}
    \caption{The clustering performance results with different missing rates on 100Leaves, Digit4k and Animal.}
    \label{fig:main_three_bars}
\end{figure*}

\begin{table*}[!t]
\caption{Ablation Study Results on Six Benchmark Datasets With 0.5 Missing Rate. The Best Results are In \textbf{Bold}, and the Second-best Result are \underline{Underlined}.}
\label{tab:ablation}

\resizebox{\textwidth}{!}{ 
\begin{tabular}{@{} ccccc ccc ccc ccc @{}}
\toprule

\multicolumn{5}{c}{Components} & \multicolumn{3}{c}{Reuters\_21578} & \multicolumn{3}{c}{100Leaves} & \multicolumn{3}{c}{Digit4k} \\
\cmidrule(lr){1-5} \cmidrule(lr){6-8} \cmidrule(lr){9-11} \cmidrule(lr){12-14}
$\mathcal{L}_{REC}$ & $\mathcal{L}_{CON}$ & $\mathcal{L}_{PRO}$ & $\mathcal{L}_{SHARED}$ & $\mathcal{L}_{DE}$ 
& ACC & ARI & F-Sco & ACC & ARI & F-Sco & ACC & ARI & F-Sco \\
\midrule

$\checkmark$ & & & & 
& 44.08 & 18.50 & 44.57 
& 66.50 & 50.96 & 63.84 
& 83.90 & 63.44 & 63.86 \\

& $\checkmark$ & & & 
& 43.40 & 17.45 & 43.21 
& 69.06 & 53.83 & 67.25 
& 81.52 & 58.85 & 62.47 \\

& & $\checkmark$ & $\checkmark$ & $\checkmark$ 
& \underline{51.87} & \underline{25.88} & \textbf{52.84} 
& 67.62 & 51.59 & 64.71 
& 83.42 & 62.44 & 63.63 \\

$\checkmark$ & $\checkmark$ & & & 
& 45.27 & 20.37 & 47.12 
& 67.38 & 51.78 & 65.63 
& \underline{83.92} & \underline{63.48} & \underline{63.87} \\

$\checkmark$ & & $\checkmark$ & $\checkmark$ & $\checkmark$ 
& 44.80 & 18.53 & 44.58 
& 66.44 & 50.26 & 63.10 
& 83.38 & 62.46 & 63.44 \\

& $\checkmark$ & $\checkmark$ & $\checkmark$ & $\checkmark$ 
& 46.67 & 19.57 & 46.85 
& 68.25 & 53.01 & 66.12 
& 81.50 & 58.80 & 62.46 \\

$\checkmark$ & $\checkmark$ & $\checkmark$ & $\checkmark$ & 
& 45.07 & 20.29 & 45.92 
& \underline{69.38} & 53.46 & \underline{66.70} 
& 81.50 & 58.80 & 62.46 \\

$\checkmark$ & $\checkmark$ & $\checkmark$ & & $\checkmark$ 
& 45.13 & 20.28 & 47.00 
& \underline{69.38} & \underline{53.73} & \textbf{67.47} 
& 81.52 & 58.85 & 62.47 \\

$\checkmark$ & $\checkmark$ & $\checkmark$ & $\checkmark$ & $\checkmark$ 
& \textbf{53.47} & \textbf{26.44} & \underline{51.39} 
& \textbf{69.56} & \textbf{54.35} & \textbf{67.47} 
& \textbf{83.95} & \textbf{63.52} & \textbf{63.90} \\
\bottomrule

\addlinespace[3ex]
\toprule

\multicolumn{5}{c}{Components} & \multicolumn{3}{c}{Animal} & \multicolumn{3}{c}{ALoi100} & \multicolumn{3}{c}{VGGFace2\_50} \\
\cmidrule(lr){1-5} \cmidrule(lr){6-8} \cmidrule(lr){9-11} \cmidrule(lr){12-14}
$\mathcal{L}_{REC}$ & $\mathcal{L}_{CON}$ & $\mathcal{L}_{PRO}$ & $\mathcal{L}_{SHARED}$ & $\mathcal{L}_{DE}$ 
& ACC & ARI & F-Sco & ACC & ARI & F-Sco & ACC & ARI & F-Sco \\
\midrule

$\checkmark$ & & & & 
& 53.72 & 42.08 & 53.80 
& 70.59 & 58.71 & 67.73 
& 10.50 & 3.01  & 10.20 \\

& $\checkmark$ & & & 
& 39.88 & 27.11 & 40.72 
& 66.31 & 50.02 & 63.47 
& 4.75  & 0.71  & 4.66 \\

& & $\checkmark$ & $\checkmark$ & $\checkmark$ 
& 52.64 & 40.35 & 52.93 
& 71.30 & 60.35 & 67.58 
& 10.02 & 2.88  & 9.74 \\

$\checkmark$ & $\checkmark$ & & & 
& 52.81 & 42.04 & 52.69 
& 73.06 & 61.20 & 69.90 
& 10.44 & 3.02  & 10.03 \\

$\checkmark$ & & $\checkmark$ & $\checkmark$ & $\checkmark$ 
& 56.65 & 44.34 & 56.24 
& 70.80 & 59.98 & 67.31 
& 9.90  & 2.75  & 9.95 \\

& $\checkmark$ & $\checkmark$ & $\checkmark$ & $\checkmark$ 
& 41.77 & 28.46 & 41.86 
& 67.15 & 55.43 & 64.53 
& 4.87  & 0.73  & 4.73 \\

$\checkmark$ & $\checkmark$ & $\checkmark$ & $\checkmark$ & 
& \underline{56.91} & \underline{45.21} & \underline{56.89} 
& \underline{73.36} & \underline{61.52} & \underline{70.33} 
& 11.04 & 3.11  & 10.91 \\

$\checkmark$ & $\checkmark$ & $\checkmark$ & & $\checkmark$ 
& 52.96 & 42.24 & 52.77 
& 72.22 & 60.97 & 69.12 
& \underline{11.13} & \textbf{3.18}  & \underline{10.99} \\

$\checkmark$ & $\checkmark$ & $\checkmark$ & $\checkmark$ & $\checkmark$ 
& \textbf{58.48} & \textbf{45.88} & \textbf{57.57} 
& \textbf{74.56} & \textbf{64.04} & \textbf{71.32} 
& \textbf{11.29} & \underline{3.17}  & \textbf{11.23} \\

\bottomrule
\end{tabular}
}
\end{table*}

\textbf{Comparing Methods}: We choose 14 state-of-the-art IMVC methods as baselines and compare them with our SPORT framework to investigate SPORT's effectiveness. These baselines are CDIMC-net (IJCAI'21) \cite{wen2021structural}, COMPLETER (CVPR'21) \cite{lin2021completer}, SURE (TPAMI'22) \cite{yang2022robust}, ProImp (IJCAI'23) \cite{li2023incomplete}, APADC (TIP'23) \cite{xu2023adaptive}, IMVC-IE (ICASSP'24) \cite{huang2024incomplete}, RPCIC (ACM MM'24) \cite{yuan2024robust}, BURG (ICCV'25) \cite{jin2025deep}, BRIDGE (ICCV'25) \cite{jiang2025unified}, HSACC (NeurIPS'25) \cite{ding2025incomplete1}, IMC-MCL (TKDE'25) \cite{yin2025incomplete}, MGCCFF (AAAI'25) \cite{zhao2025incomplete}, DGIMVCM (AAAI'26) \cite{zhang2026dynamic} and GMAE (TPAMI'26) \cite{11494292}.

\textbf{Experiment Configurations}: All experiments are conducted on an NVIDIA RTX L20 GPU with 48 GB memory using PyTorch 2.11.0+CUDA 13.0. For each dataset we set the missing rate as $[0.1,0.3,0.5,0.7]$. The implementation of SPORT consists of two phases, namely the pretraining phase and fine-tuning phase with the learning rate set to 0.0005 and 0.00001 respectively. We set the pretraining epoch $E_{PRE}$ to 200 and fine-tuning epoch $E_{FIN}$ to 100. Our method adopts an autoencoder with the encoder dimensionality structured as Input dimension-512-512-1024-$d$, and the decoder dimensionality structured as $d$-1024-512-512-Input dimension where $d$ is set to 256. Additionally, we set the the batch size to 512, and $d_c$ to 128.
For the loss function, $\alpha$ and $\beta$, the trade-off parameters of the overall loss are set to $1.6 \times 10^{-5}$ and 5 respectively, while $\lambda$ and $\eta$ which balance the shared alignment loss and view-specific de-correlation loss are set to 10 and 5. Moreover, the temperature parameters $[\tau,\tau_w]$ are set to $[0.75,0.5]$ and $t$ the exponent of the shared-alignment loss is set to 3. Furthermore, $\gamma$ is set to 0.3 for our hybrid imputation strategy.
During our experiment, we selectively fine-tune a subset 
of parameters certain datasets to accommodate varying dataset characteristics. The specific configurations are provided in the code repository at \url{https://github.com/EricGuo2004/SPORT_IMVC}.

\subsection{Performance Comparisons}
The clustering performances of SPORT and 14 competing methods on six multi-view datasets under different missing rates are summarized in TABLE \ref{tab:1} and TABLE \ref{tab:2}, where the optimal results are highlighted in \textbf{bold} and the second-best results are \underline{underlined}. From these tables, we obtain the following observations:

(1) SPORT performs better than other baseline methodss on most datasets. For example, on the Animal dataset, SPORT obtains the best result in terms of ACC, ARI and F-score under all missing rates. Specifically, when the missing rate is 0.5, comparing to the second-best model (BURG) SPORT improves the ACC ARI and F-Score by $17.7\%$, $26.8\%$ and $6.5\%$ respectively. Such clustering performance benefits from the structure-aware relational contrastive learning module and prototype partial alignment module where SPORT  is able to jointly enforce structural and cross-view consistency while leveraging view-specific information. 

(2) The proposed method demonstrates remarkable robustness against increasing missing rates. For instance, on the 100Leaves dataset, SPORT only encounters a $8.1\%$ and $6.7\%$ fluctuation on ACC when the missing rate increases from 0.1 to 0.3 and 0.3 to 0.5 respectively, while the acc of GMAE, which exhibits strong clustering results under small missing rates, drops $14.8\%$ and $12.5\%$. In addition, on the Digit4k dataset, the ACC remains above $80\%$ for all missing rates, and only encounters a performance decline of $1.8\%$ when the missing rate increases from $0.1\%$ to $0.7\%$. These results justify that our hybrid imputation strategy simultaneously captures global and local semantics, which allows SPORT  to demonstrate strong robustness under large missing rates when most features are deprecated and low-quality prototypes are generated.

To further validate the superiority of SPORT, Figure.\ref{fig:main_three_bars} provides an intuitive visualization of the performance degradation trends for all models with increasing missing rates. As shown in Figure.\ref{fig:main_three_bars} (a),~(b) and (c), when the missing rate increases, almost all methods suffer a performance decline, indicating that missing information impairs model performance. However, comparing to other methods, in most scenarios the ACC curves of SPORT are positioned at the top while remaining relative smooth and flat with increasing missing rates. This observation further demonstrates the effectiveness and robustness of the proposed method under different levels of missingness.

\subsection{Ablation Study}
As defined in Eq.{(\ref{eq:3})}, our SPORT model consists of three major modules namely the feature representation learning module, the structure-aware relational contrastive learning module and the prototype partial alignment module. To further validate the contributions of each individual module, in this subsection we conduct ablation study on all six datasets with a missing rate of 0.5. The components are systematically removed while the model is retrained under the same configuration. The experimental results are shown in TABLE \ref{tab:ablation}. From this table we can infer that all parts of SPORT contributes to the overall clustering performance and omitting any key component leads to varying degrees of performance degradation.
Specifically, relying solely on representation learning and contrastive learning without prototype partial alignment results in significant accuracy drops. Taking the Reuters\_21578 example, the ACC decreases from 53.47\% to 45.27\% while the F-sco drops from 51.39\% to 47.12\% when the generated prototypes are not partially aligned. Furthermore, neglecting the representation learning module or the contrastive learning module also triggers sharp performance plunges. Notably, on the Animal dataset, masking $\mathcal{L}_{REC}$ dramatically drags the ARI down by 38.97\%, whereas on the Reuters\_21578 dataset, dropping $\mathcal{L}_{CON}$ causes an 8.67\% decrease in ACC. Furthermore, components within the prototype partial alignment module, such as shared alignment, also prove indispensable; dropping it on the Animal dataset leads to a clear 5.52\% drop in ACC.
Overall, the results confirm that the three modules are highly complementary. Their joint optimization enables more discriminative feature representations, improves cross-view semantic consistency while preserving view-specific information, and ultimately leads to superior clustering performance.

\begin{figure}[!t]
    \centering

    \begin{minipage}{0.48\linewidth}
        \centering
        \includegraphics[width=\linewidth]{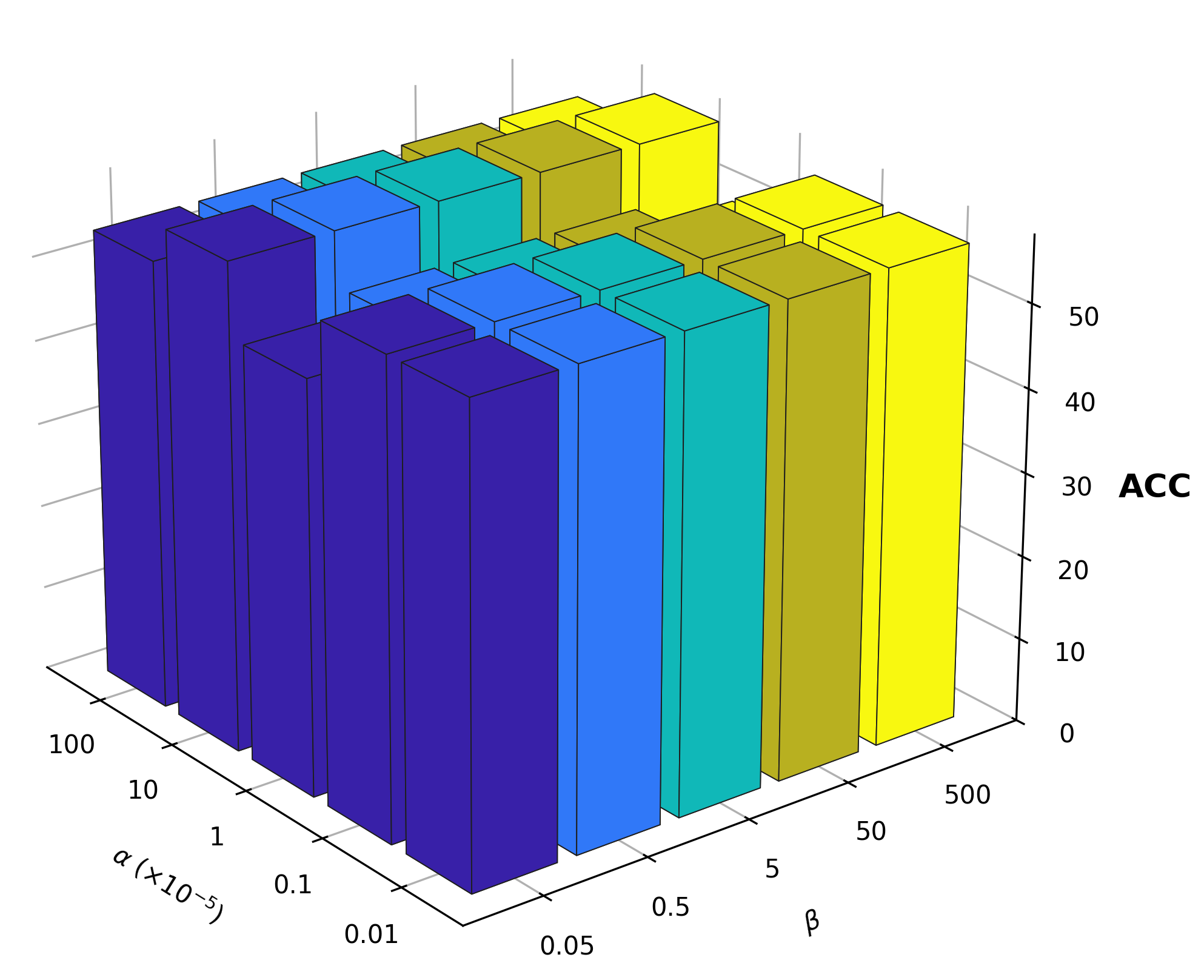}
        \small (a) Animal
    \end{minipage}\hfill
    \begin{minipage}{0.48\linewidth}
        \centering
        \includegraphics[width=\linewidth]{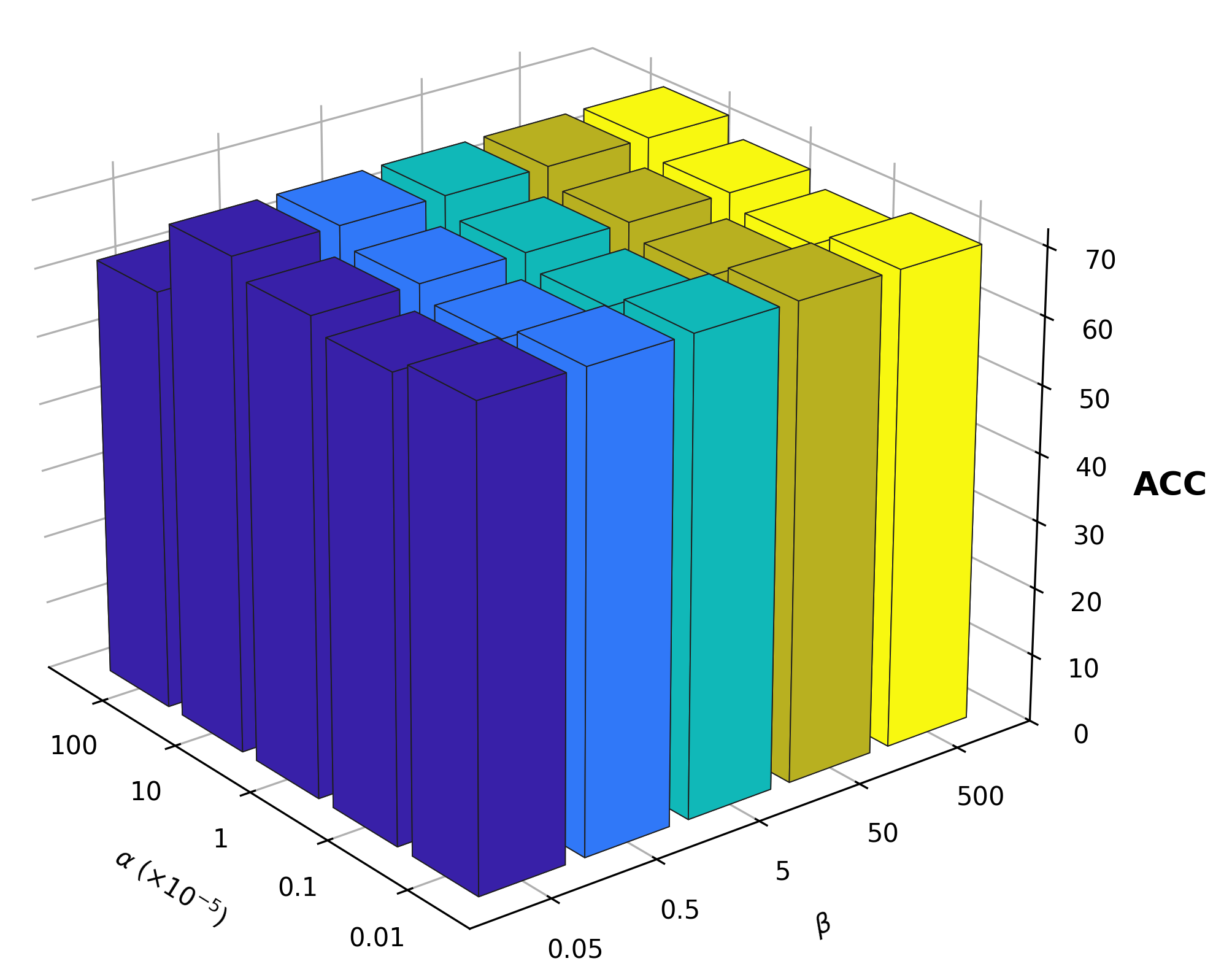}
        \small (b) ALOI\_100
    \end{minipage}

    \begin{minipage}{0.48\linewidth}
        \centering
        \includegraphics[width=\linewidth]{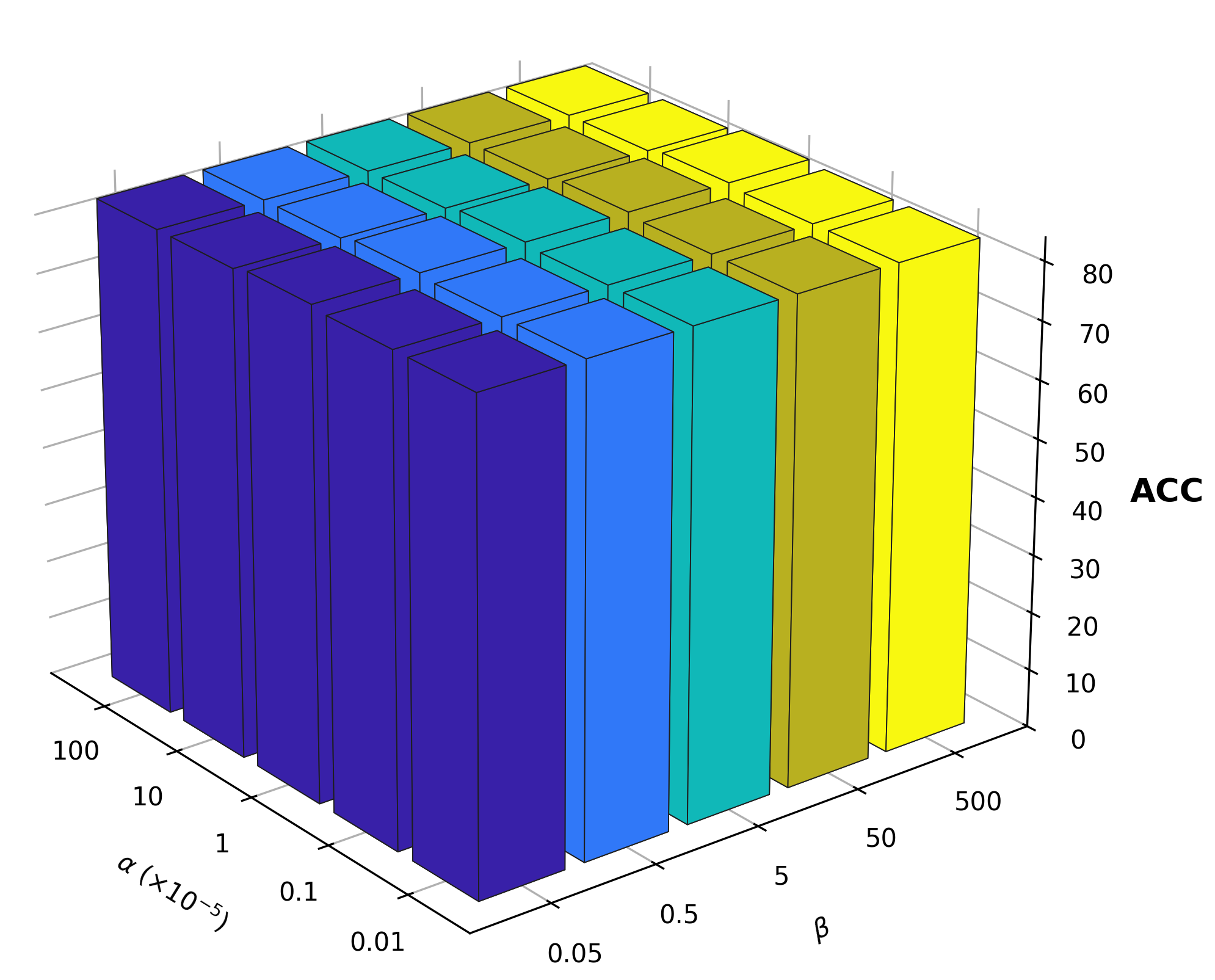}
        \small (c) Digit4k
    \end{minipage}\hfill
    \begin{minipage}{0.48\linewidth}
        \centering
        \includegraphics[width=\linewidth]{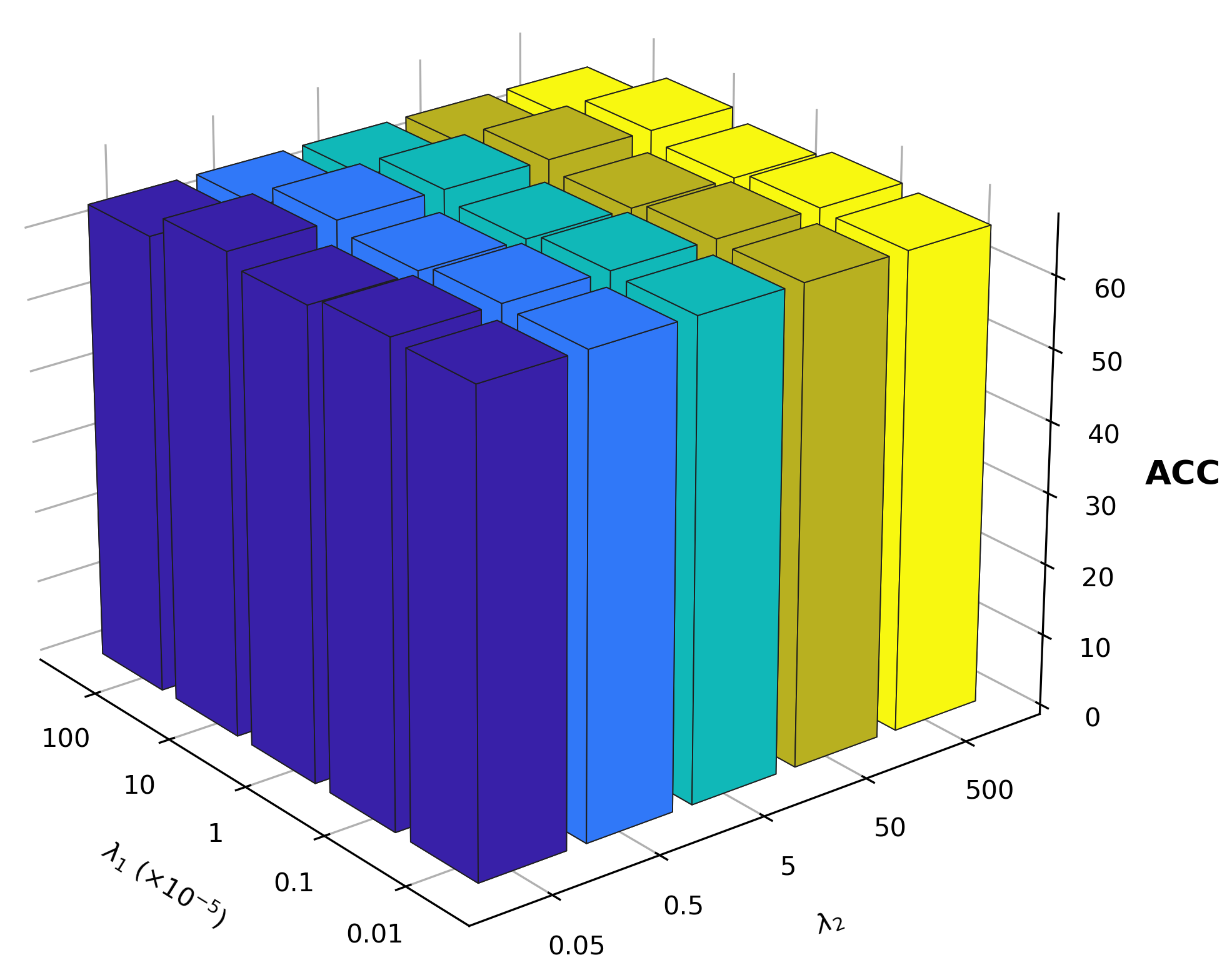}
        \small (d) 100Leaves
    \end{minipage}

    \begin{minipage}{0.48\linewidth}
        \centering
        \includegraphics[width=\linewidth]{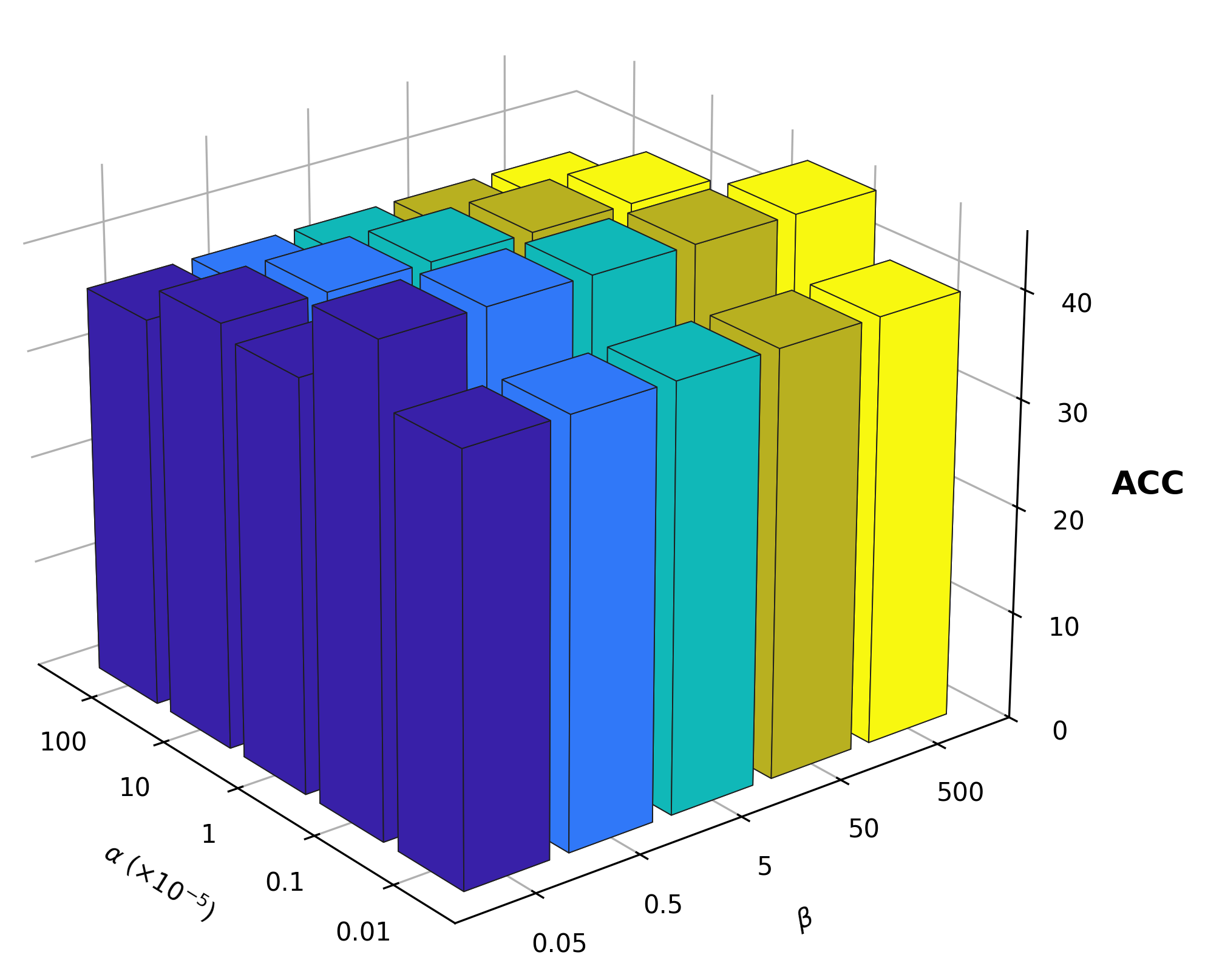}
        \small (e) Reuters\_21578
    \end{minipage}\hfill
    \begin{minipage}{0.48\linewidth}
        \centering
        \includegraphics[width=\linewidth]{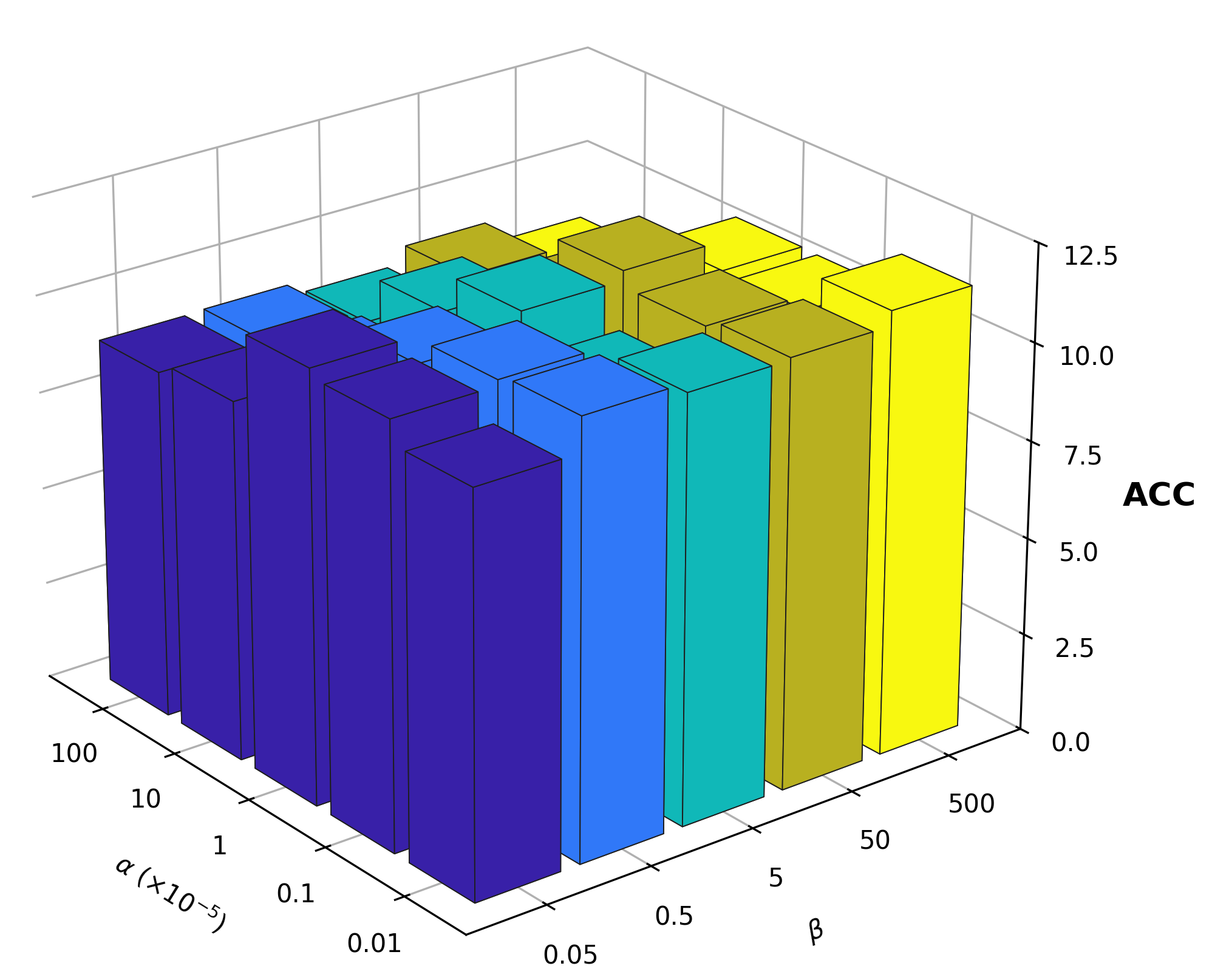}
        \small (f) VGGFace2\_50
    \end{minipage}

    \caption{Sensitive of $\alpha$ and $\beta$ with a missing rate of 0.5.}
    \label{fig:combined_lambda_plots}
\end{figure}
\begin{figure}[!t]
    \centering

    \begin{minipage}{0.48\linewidth}
        \centering
        \includegraphics[width=\linewidth]{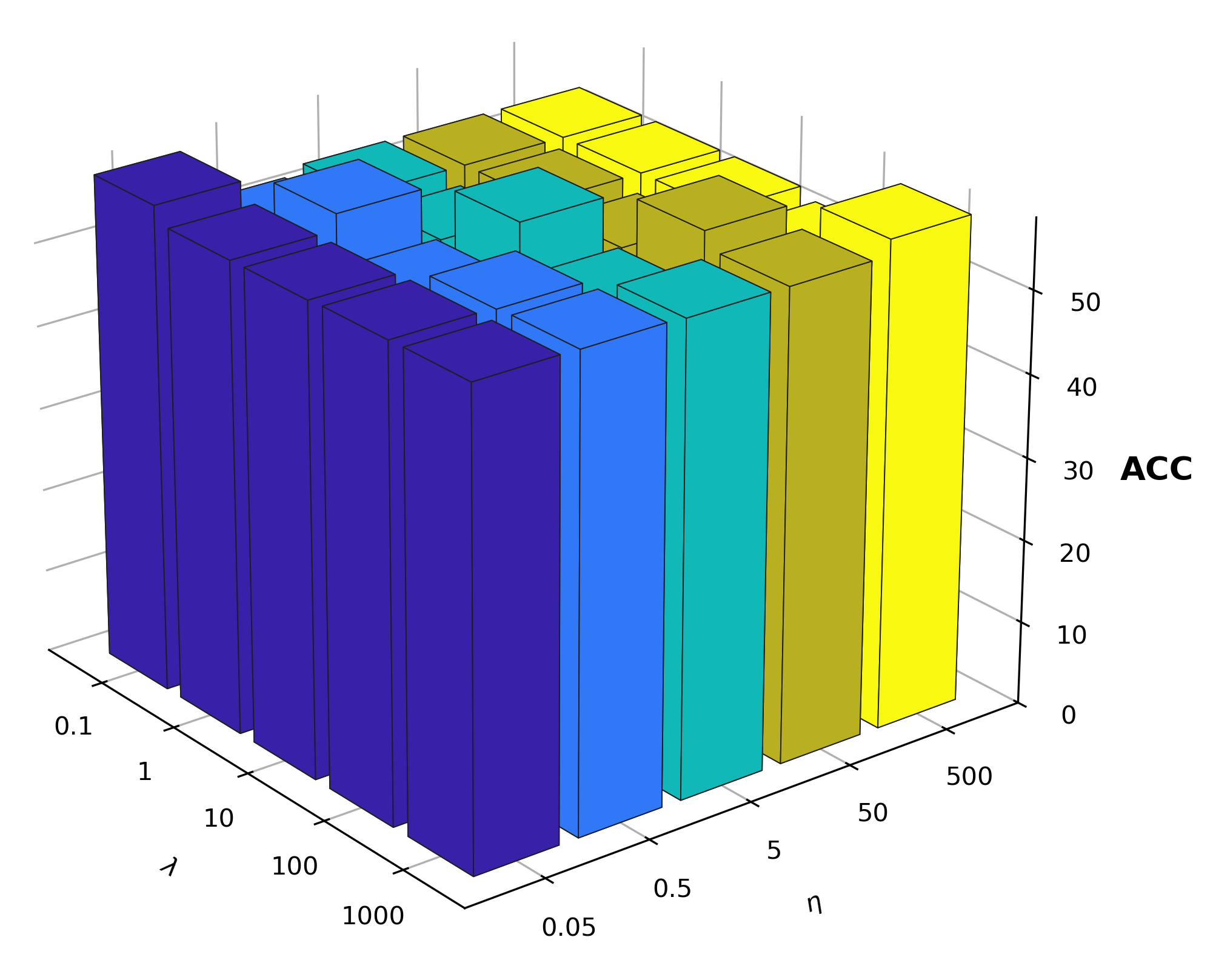}
        \small (a) Animal
    \end{minipage}\hfill
    \begin{minipage}{0.48\linewidth}
        \centering
        \includegraphics[width=\linewidth]{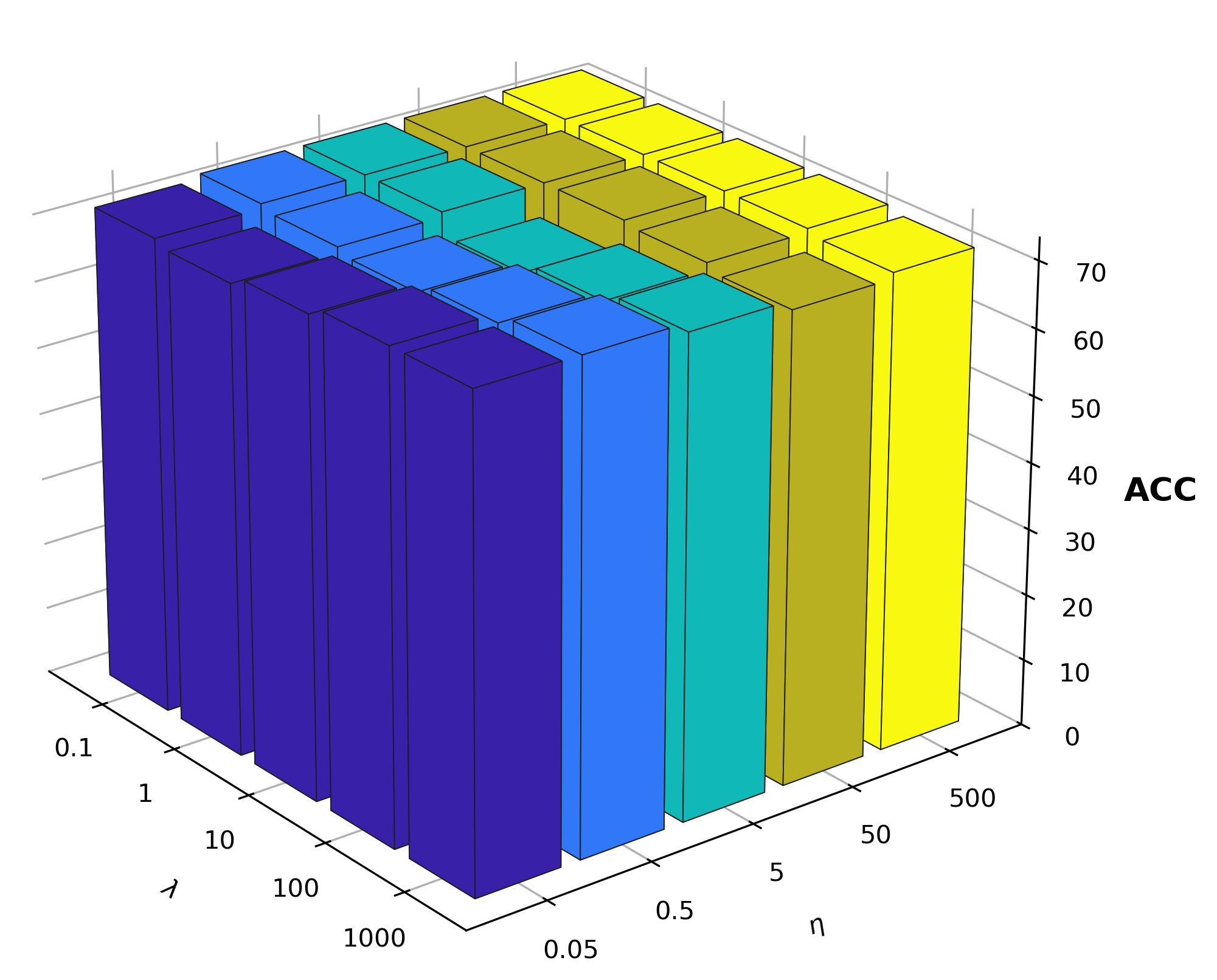}
        \small (b) ALOI\_100
    \end{minipage}

    \begin{minipage}{0.48\linewidth}
        \centering
        \includegraphics[width=\linewidth]{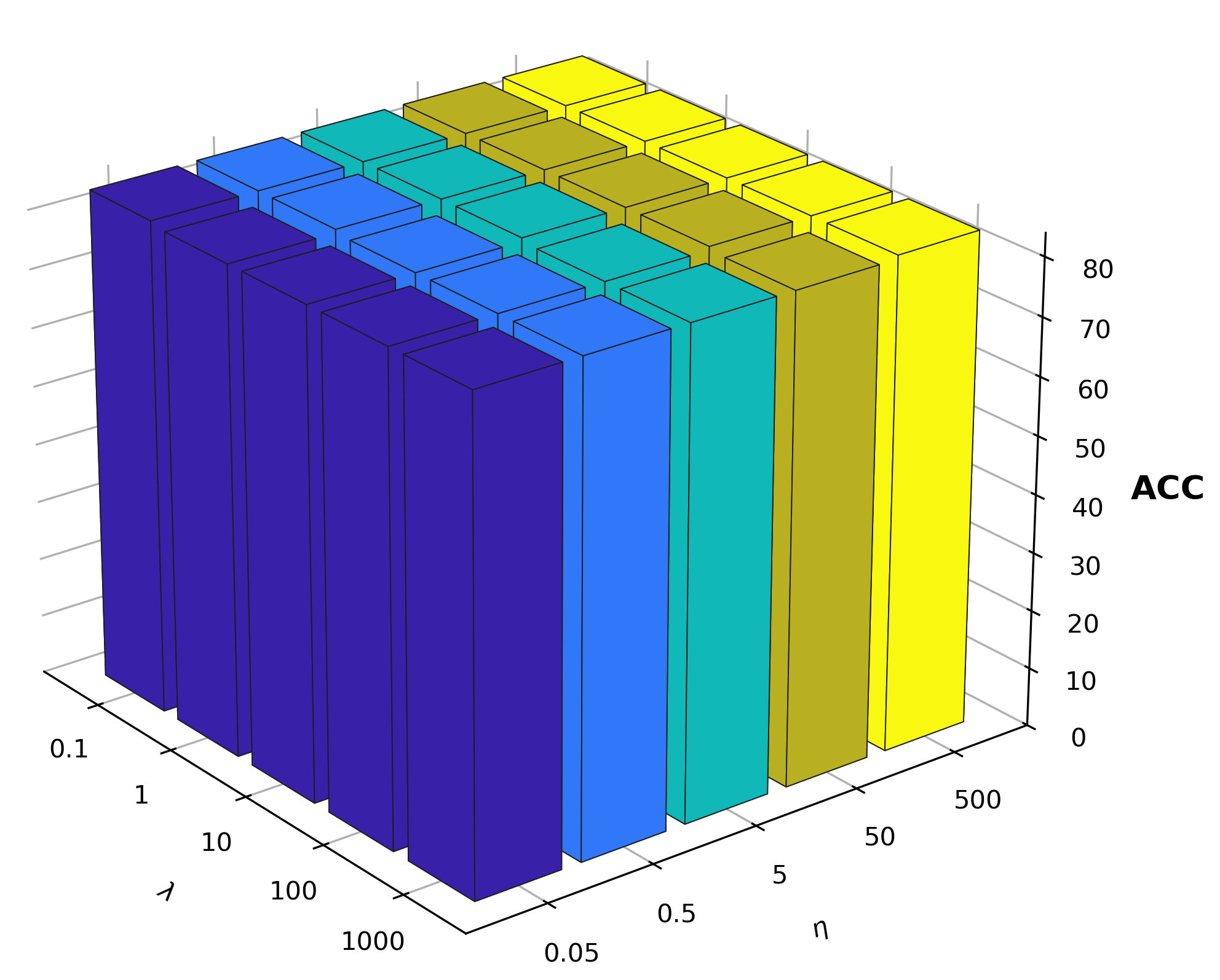}
        \small (c) Digit4k
    \end{minipage}\hfill
    \begin{minipage}{0.48\linewidth}
        \centering
        \includegraphics[width=\linewidth]{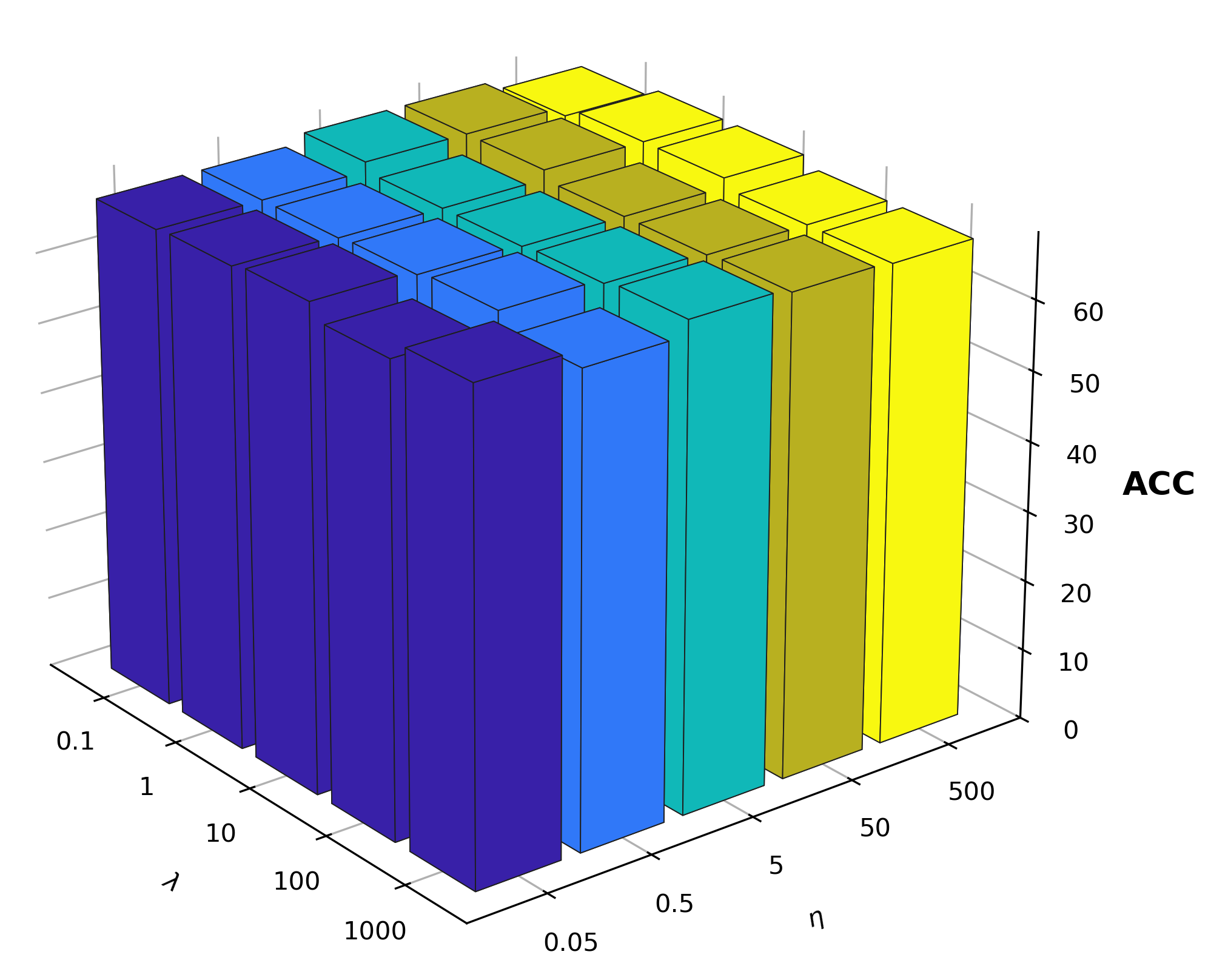}
        \small (d) 100Leaves
    \end{minipage}

    \begin{minipage}{0.48\linewidth}
        \centering
        \includegraphics[width=\linewidth]{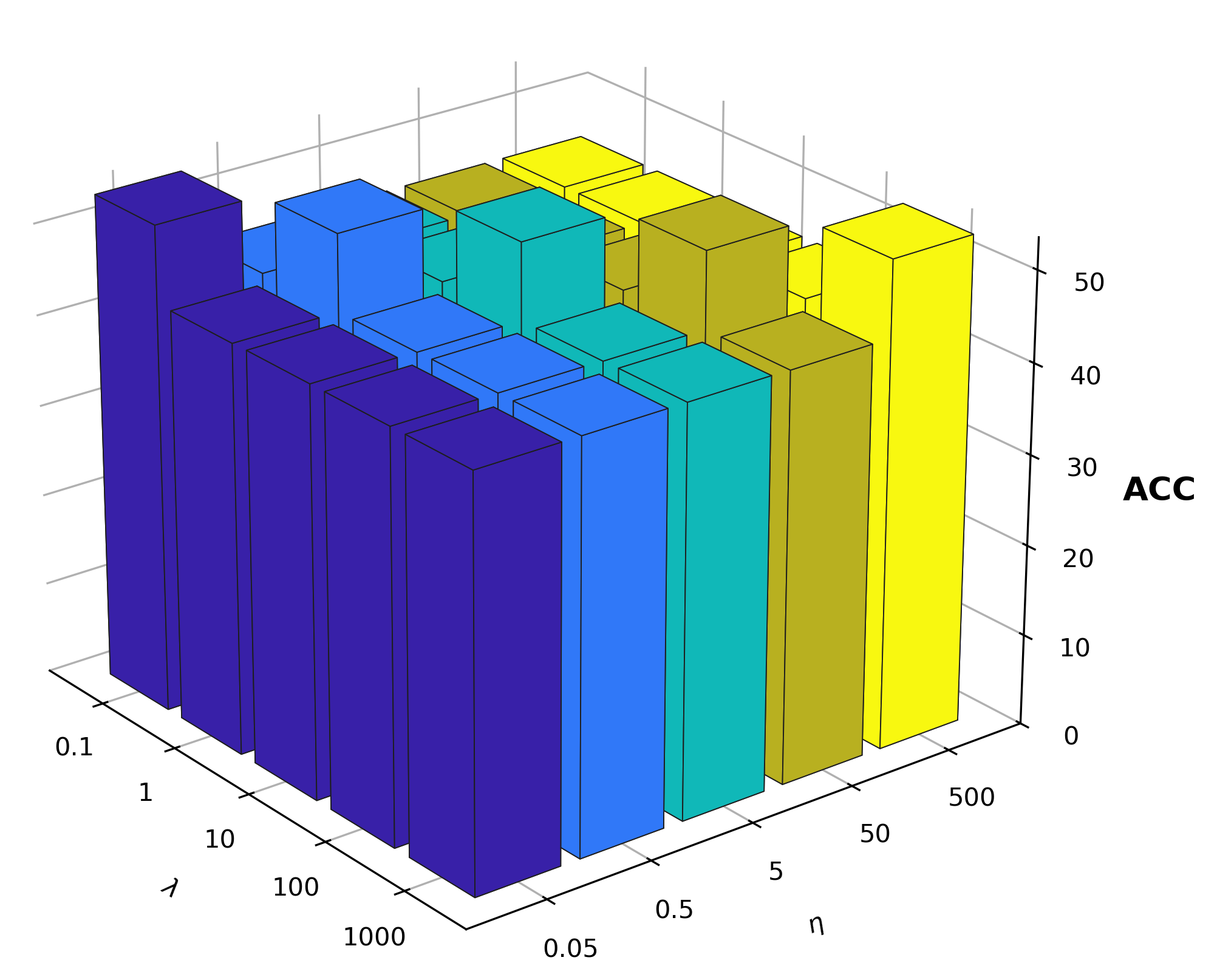}
        \small (e) Reuters\_21578
    \end{minipage}\hfill
    \begin{minipage}{0.48\linewidth}
        \centering
        \includegraphics[width=\linewidth]{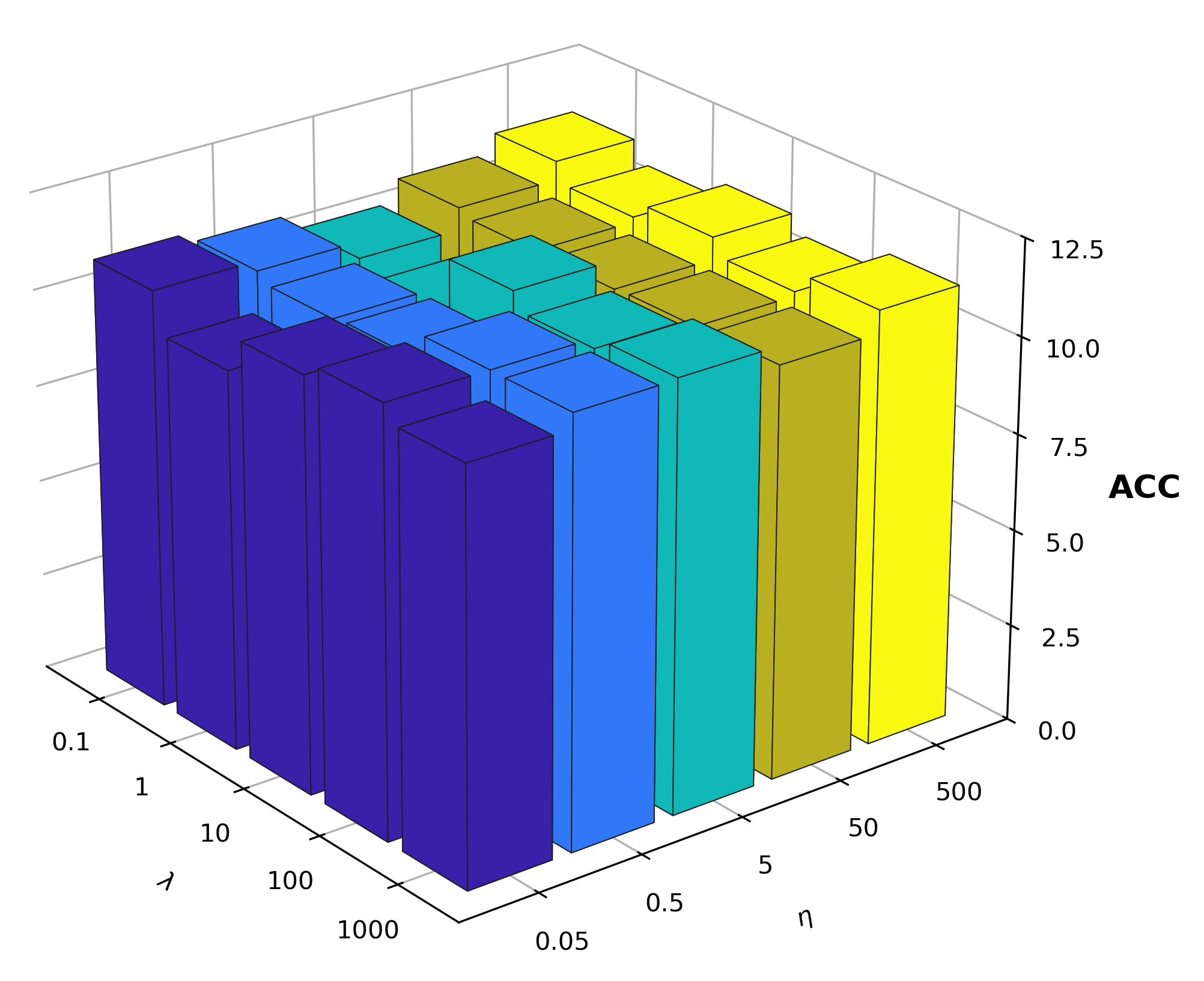}
        \small (f) VGGFace2\_50
    \end{minipage}

    \caption{Sensitive of $\lambda$ and $\eta$ with a missing rate of 0.5.}
    \label{fig:combined_lambda34_plots}
\end{figure}

\begin{figure}[!t]
    \centering

    \begin{minipage}{0.48\linewidth}
        \centering
        \includegraphics[width=\linewidth]{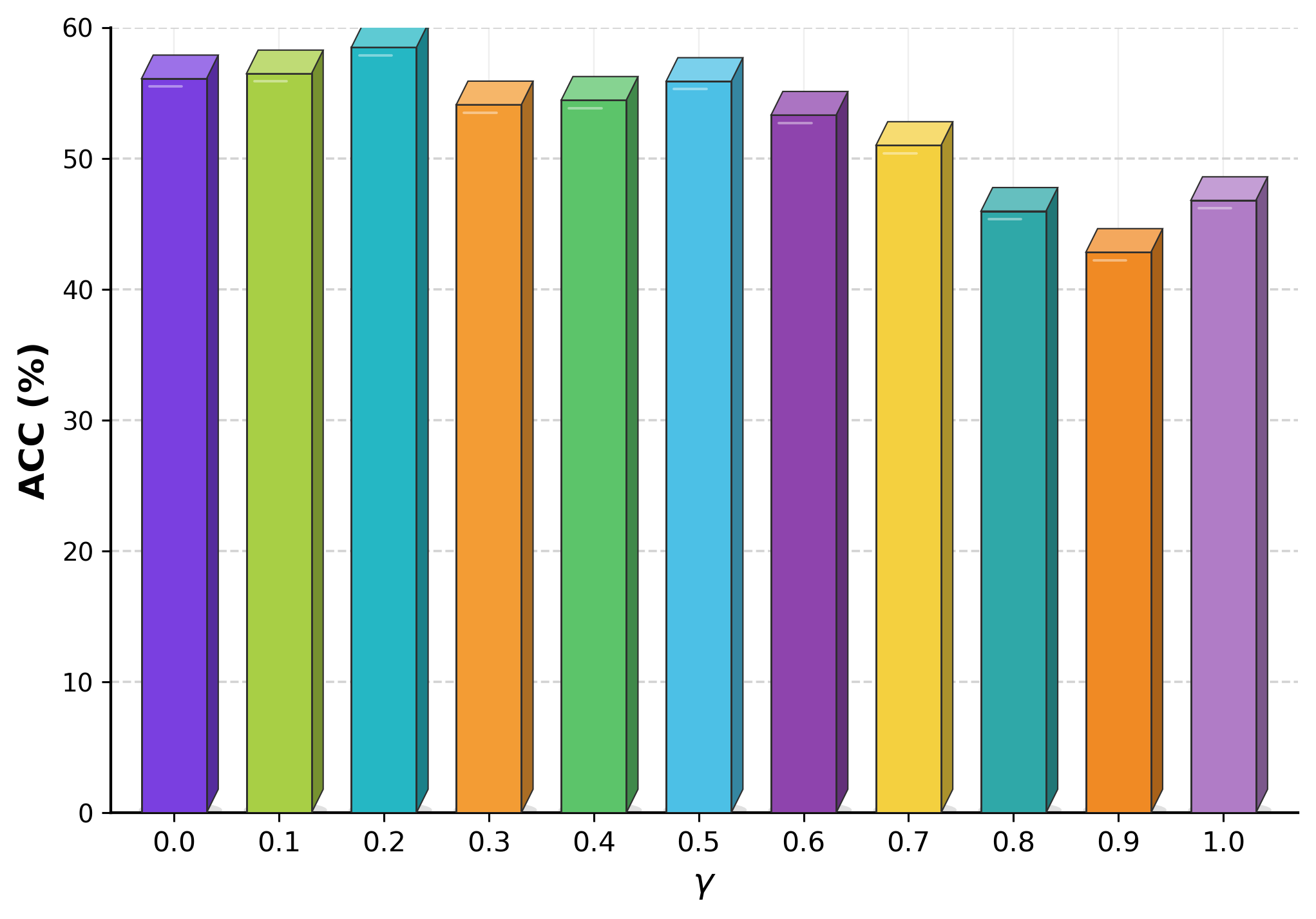}
        \footnotesize (a) Animal
    \end{minipage}\hfill
    \begin{minipage}{0.48\linewidth}
        \centering
        \includegraphics[width=\linewidth]{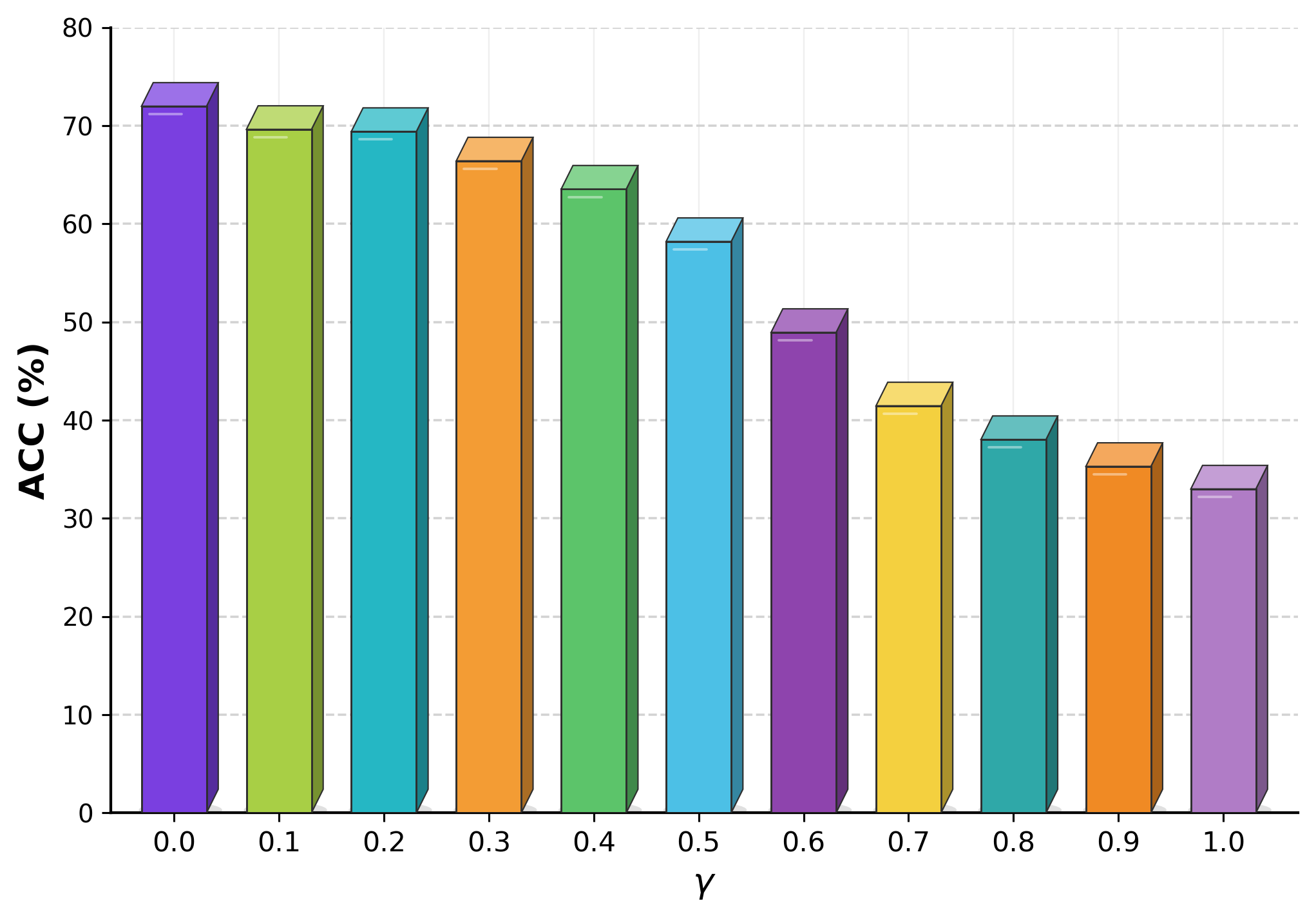}
        \footnotesize (b) ALOI\_100
    \end{minipage}

    \vspace{0.0em}

    \begin{minipage}{0.48\linewidth}
        \centering
        \includegraphics[width=\linewidth]{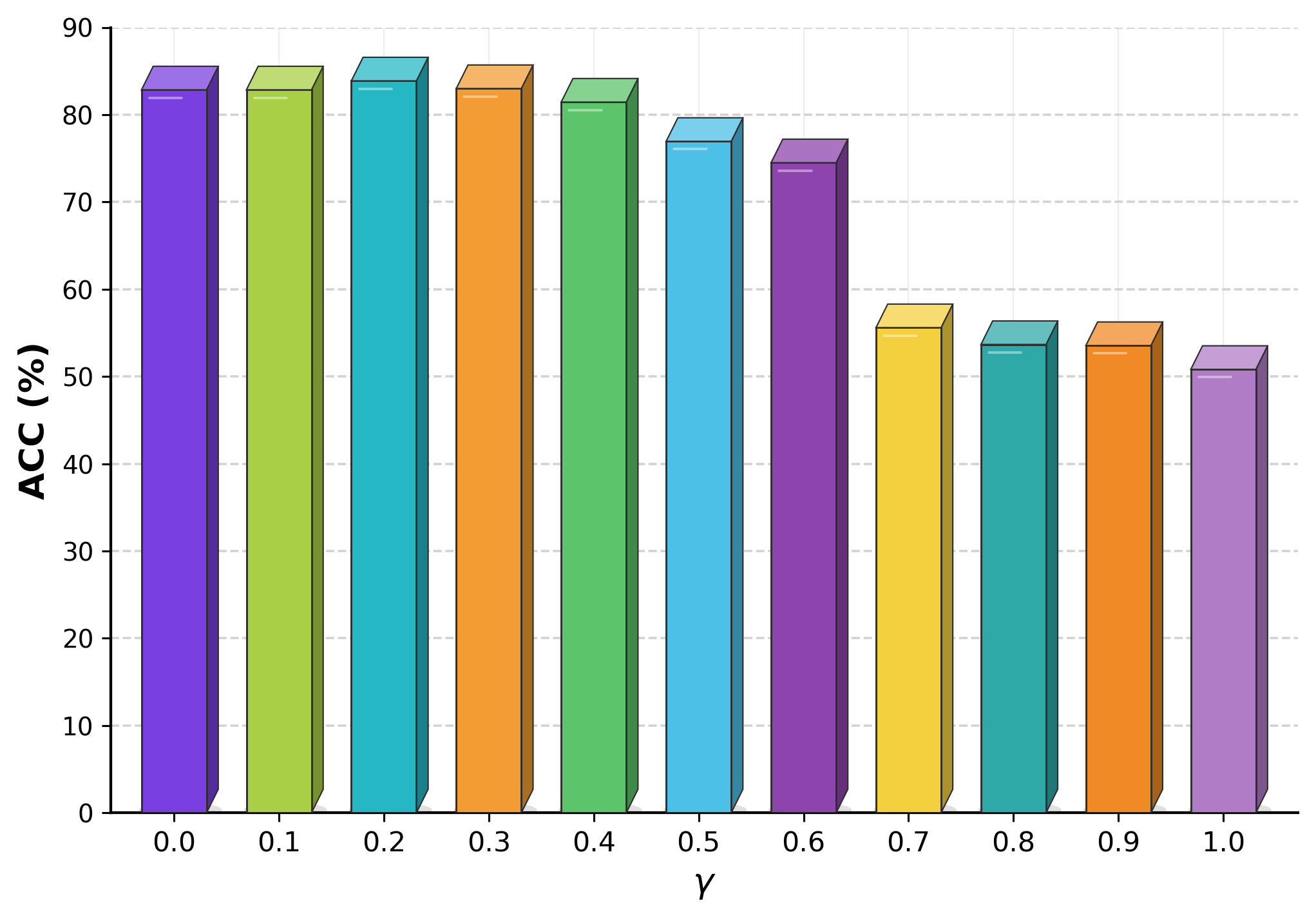}
        \footnotesize (c) Digit4k
    \end{minipage}\hfill
    \begin{minipage}{0.48\linewidth}
        \centering
        \includegraphics[width=\linewidth]{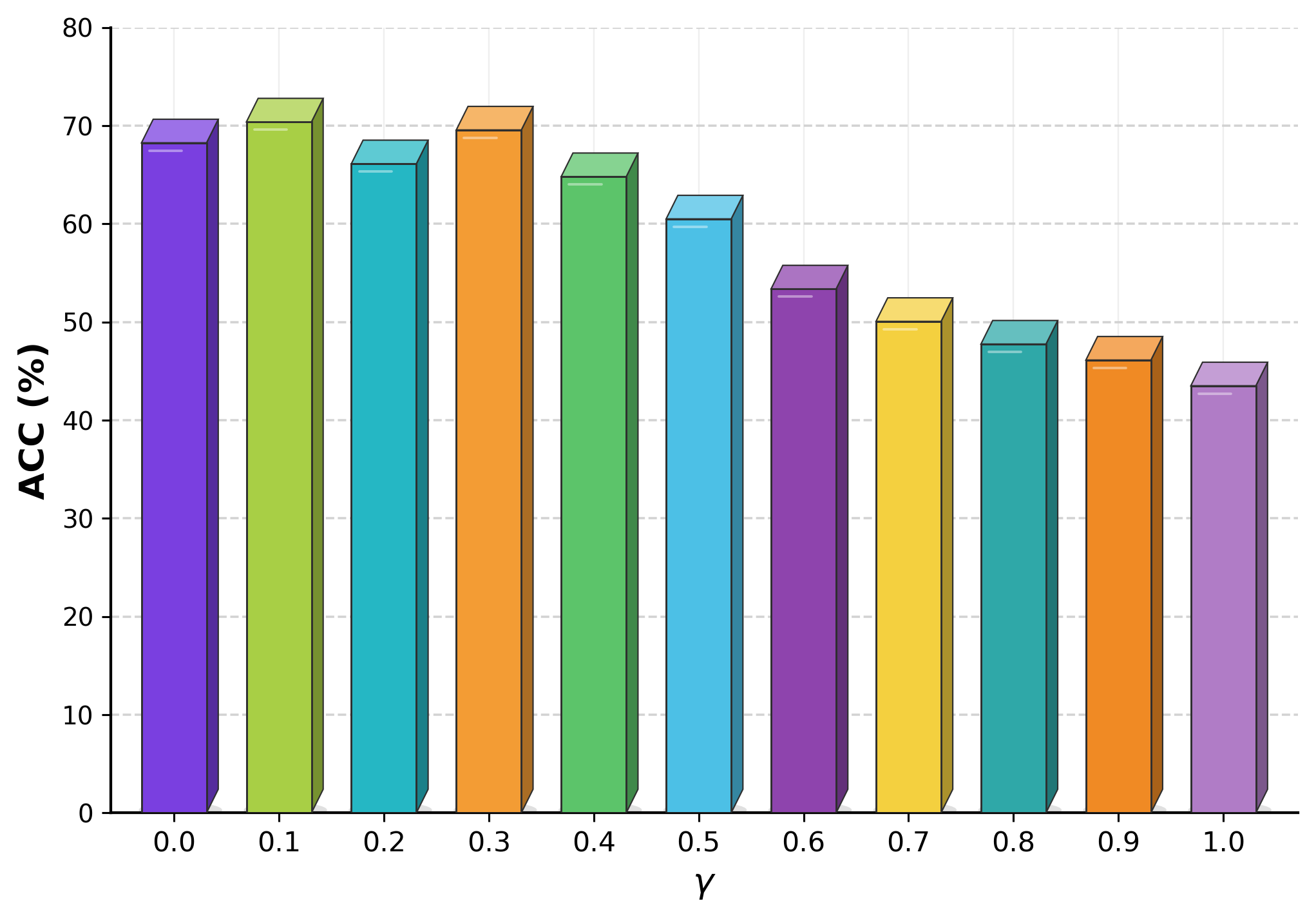}
        \footnotesize (d) 100Leaves
    \end{minipage}

    \vspace{0.0em}

    \begin{minipage}{0.48\linewidth}
        \centering
        \includegraphics[width=\linewidth]{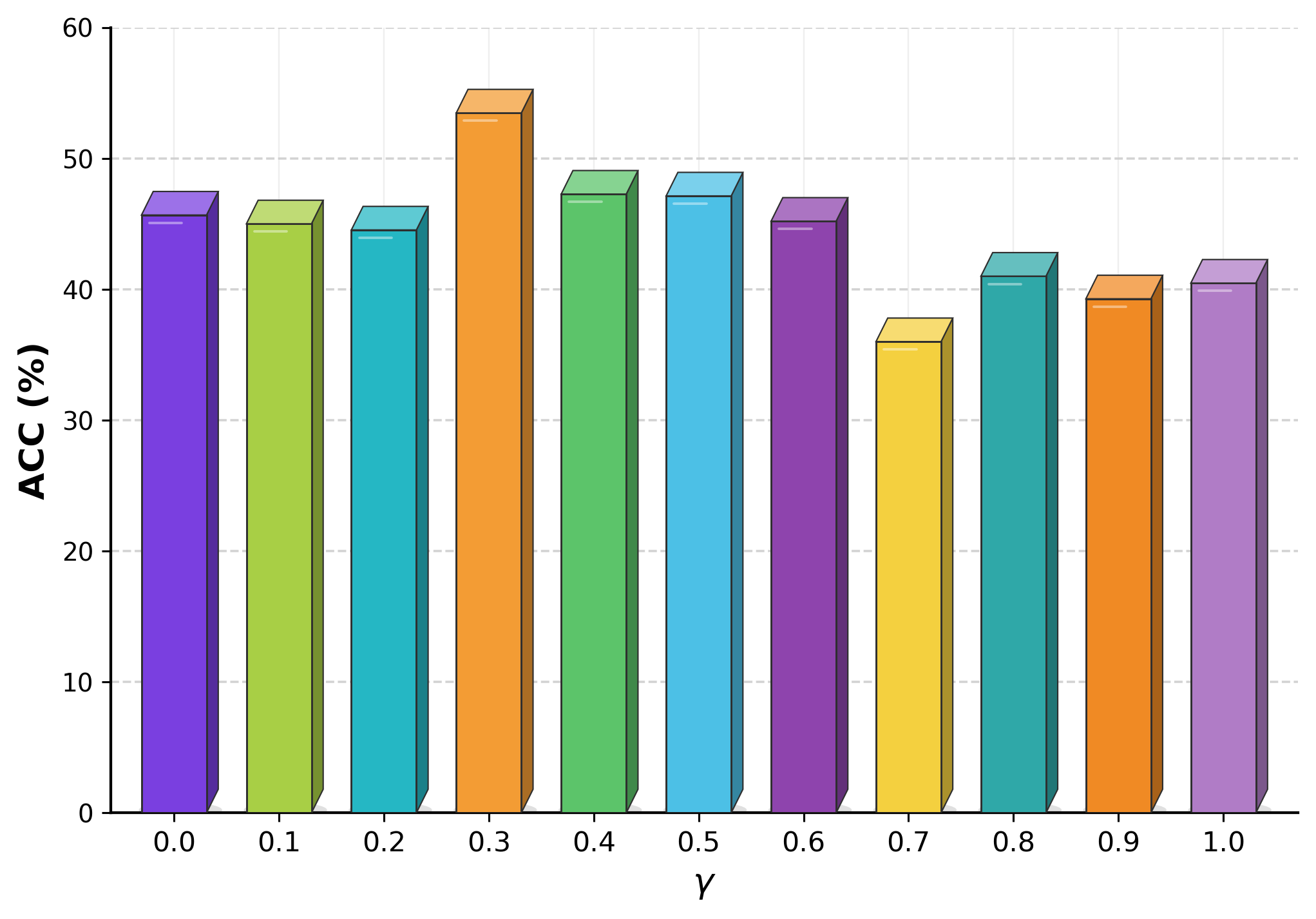}
        \footnotesize (e) Reuters\_21578
    \end{minipage}\hfill
    \begin{minipage}{0.48\linewidth}
        \centering
        \includegraphics[width=\linewidth]{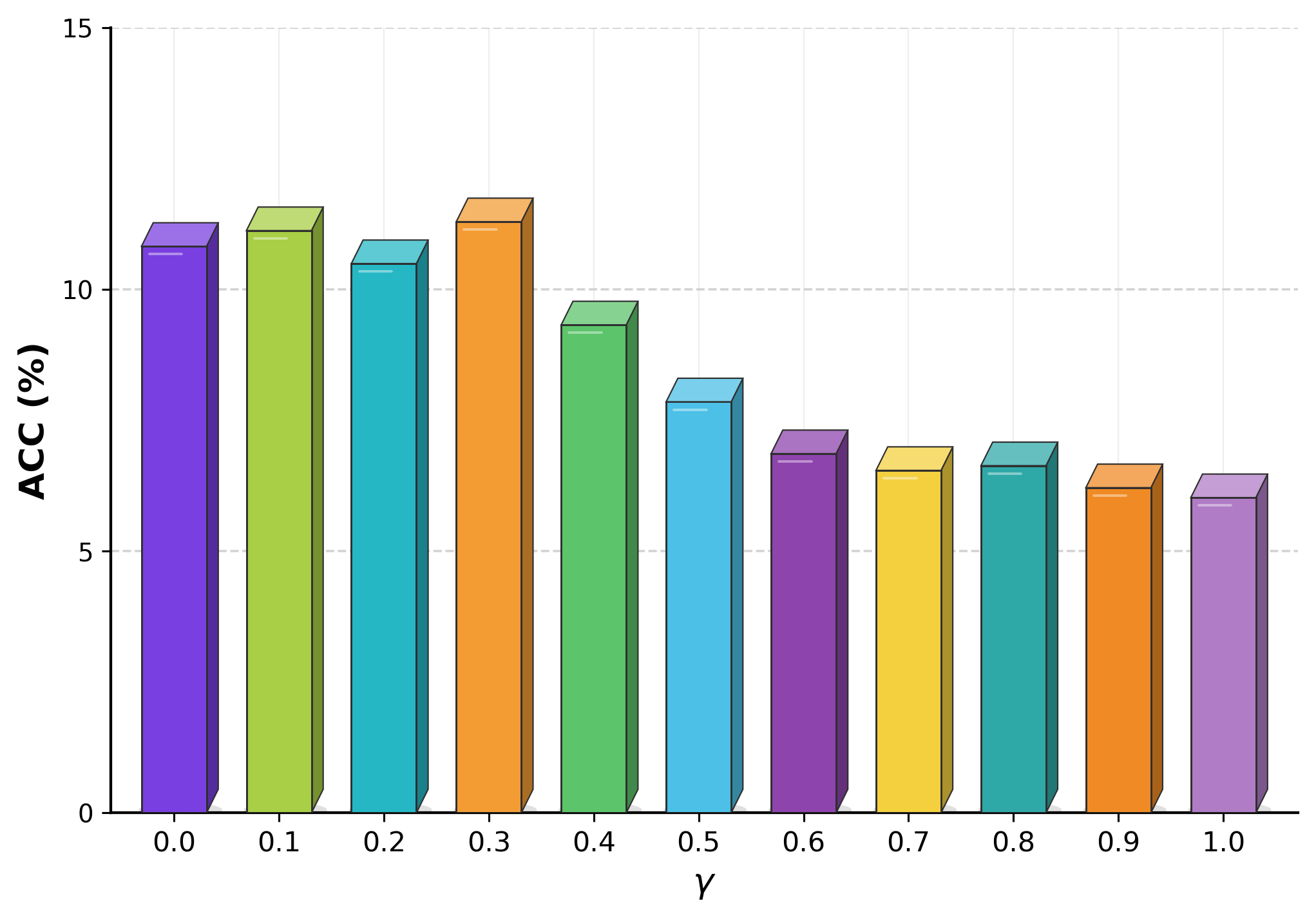}
        \footnotesize (f) VGGFace2\_50
    \end{minipage}

    \vspace{0em}

    \caption{\small Analysis of hybrid imputation under 0.5 missing rate.}
    \label{fig:combined_lam5_plots}
\end{figure}


\subsection{Sensitive Analysis}
In this paper, there are four major hyperparameters, $[\alpha,\beta,\lambda,$ $\eta]$, with $\alpha$ and $\beta$ balancing the structure-aware relational contrastive learning loss and prototype partial alignment loss. $\lambda$ and $\eta$ are used to balance the shared and view-specific loss. All six datasets are used to test the stability of the model with respect to $\alpha,\beta,\lambda,\eta$,with $\alpha \in \{10^{-7}, 10^{-6}, 10^{-5}, 10^{-4}, 10^{-3}\}$, 
$\beta \in \{5 \times 10^{-2}, 5 \times 10^{-1}, 5 \times 10^0, 5 \times 10^1, 5 \times 10^2\}$, $\lambda \in \{0.1,1,10,100,1000\}$, $\eta \in \{0.05,0.5, 5, 50, 500\}$. The missing rate for all of the dataset is set to 0.5, the detailed results are visualized in Figure.\ref{fig:combined_lambda_plots} and Figure.\ref{fig:combined_lambda34_plots}. Overall, despite some minor fluctuations under varying hyperparameters, SPORT exhibits strong robustness to hyperparameter variations. Take the Digit4k dataset for example, the ACC merely fluctuates 0.86\% under different $\alpha$ and $\beta$ settings. Additionally, under varying $\lambda$ and $\gamma$, the ACC on the 100Leaves dataset remains above 68.00\% with a maximum gap of only 1.32\% between the highest and lowest ACC values. These results demonstrate that the proposed model maintains stable performance under hyperparameter fluctuations.

\begin{figure*}[t]
    \centering
    \begin{minipage}{0.32\textwidth}
        \centering
        \includegraphics[width=\linewidth]{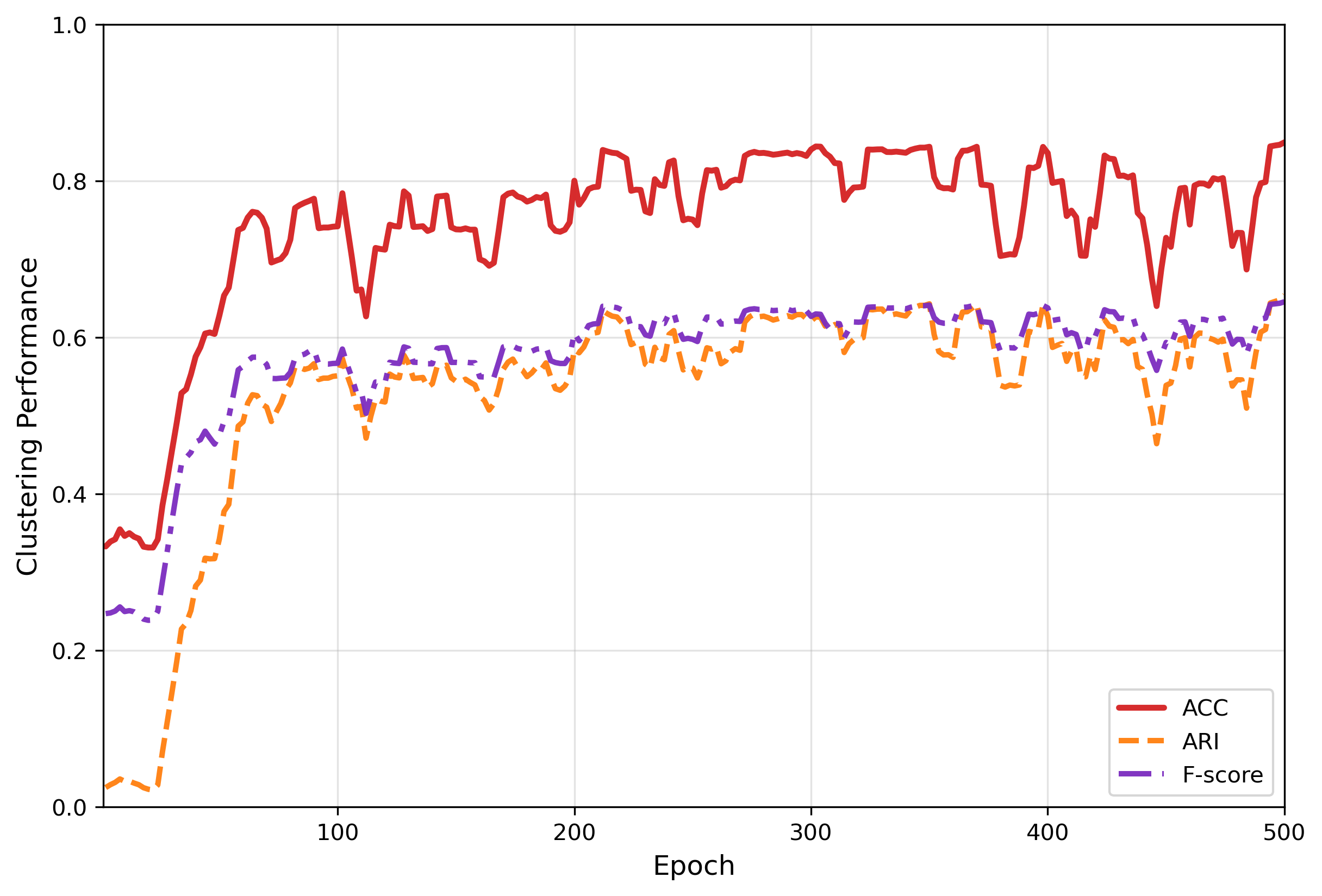}
        \small (a) Digit4k
    \end{minipage}\hfill
    \begin{minipage}{0.32\textwidth}
        \centering
        \includegraphics[width=\linewidth]{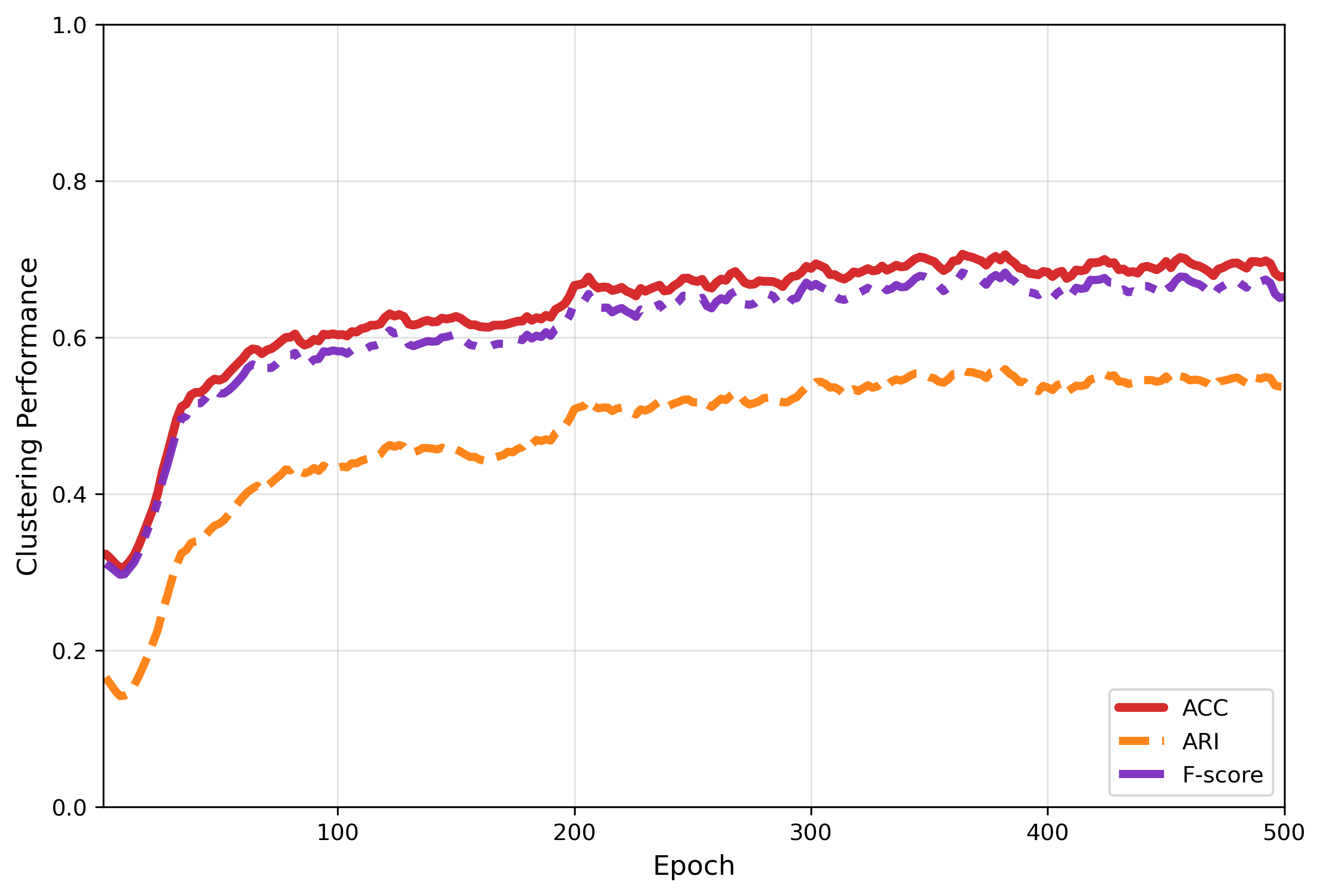}
        \small (b) Leaves\_100
    \end{minipage}\hfill
    \begin{minipage}{0.32\textwidth}
        \centering
        \includegraphics[width=\linewidth]{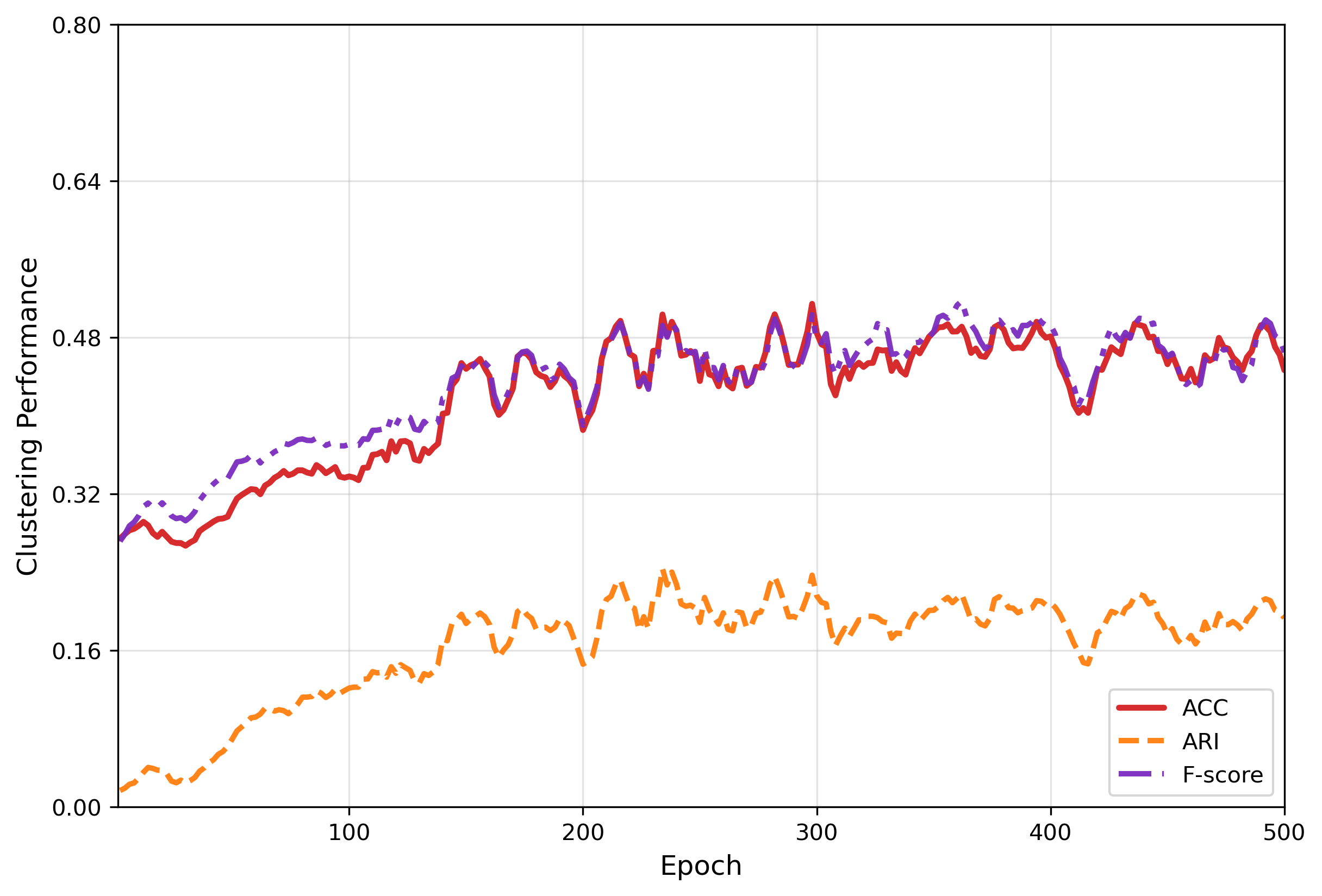}
        \small (c) Reuters\_21578
    \end{minipage}
   
    \caption{Convergence analysis of Digit4k, Leaves\_100 and Reuters\_21578.}
    \label{fig:curve}
\end{figure*}

\begin{figure*}[t] 
    \centering

    \begin{minipage}{0.23\textwidth}
        \centering
        \includegraphics[width=\linewidth]{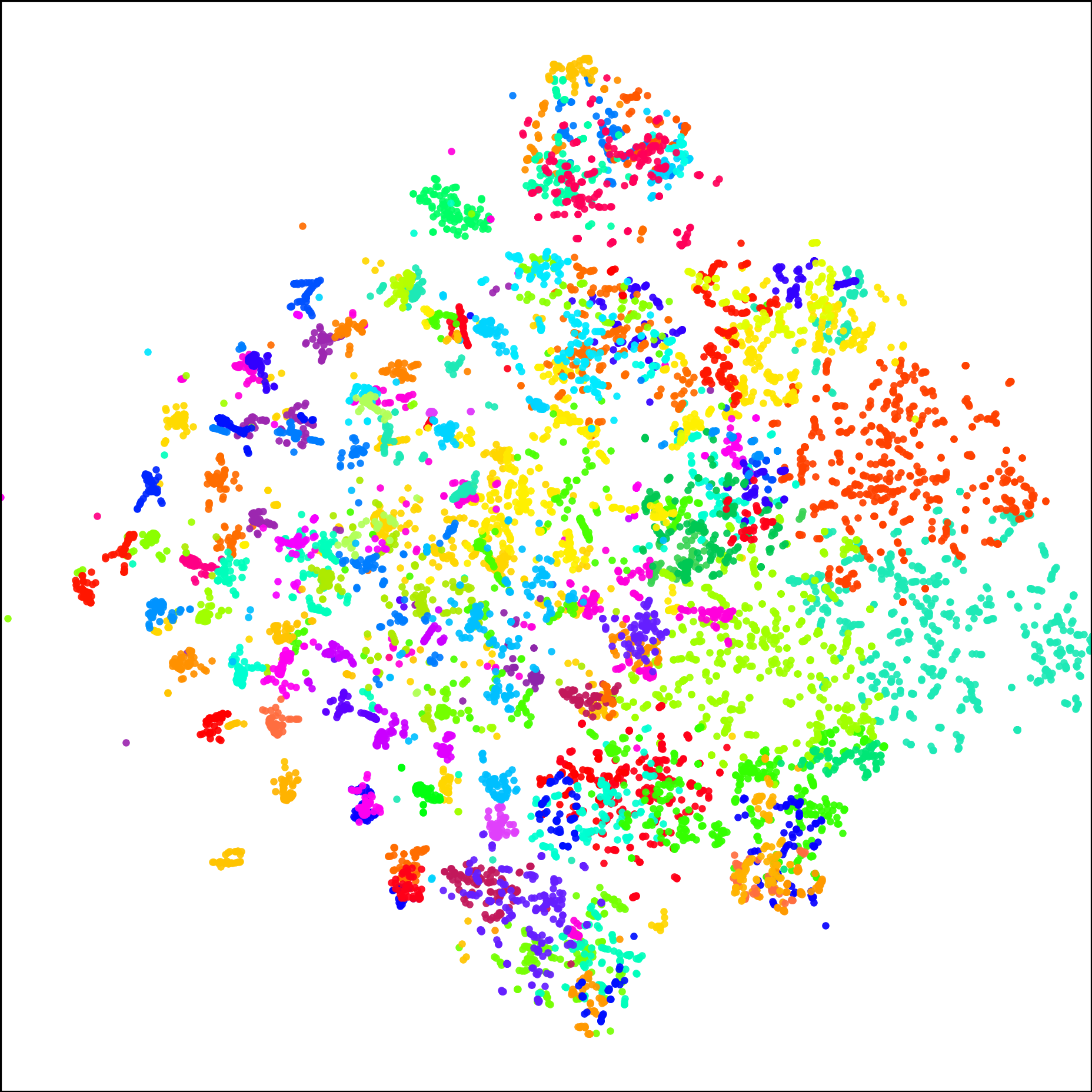}
        \small (a) Before pretraining\par
        \scriptsize (ACC:0.4142 ARI:0.2718 F-Sco:0.4253)
    \end{minipage}\hfill
    \begin{minipage}{0.23\textwidth}
        \centering
        \includegraphics[width=\linewidth]{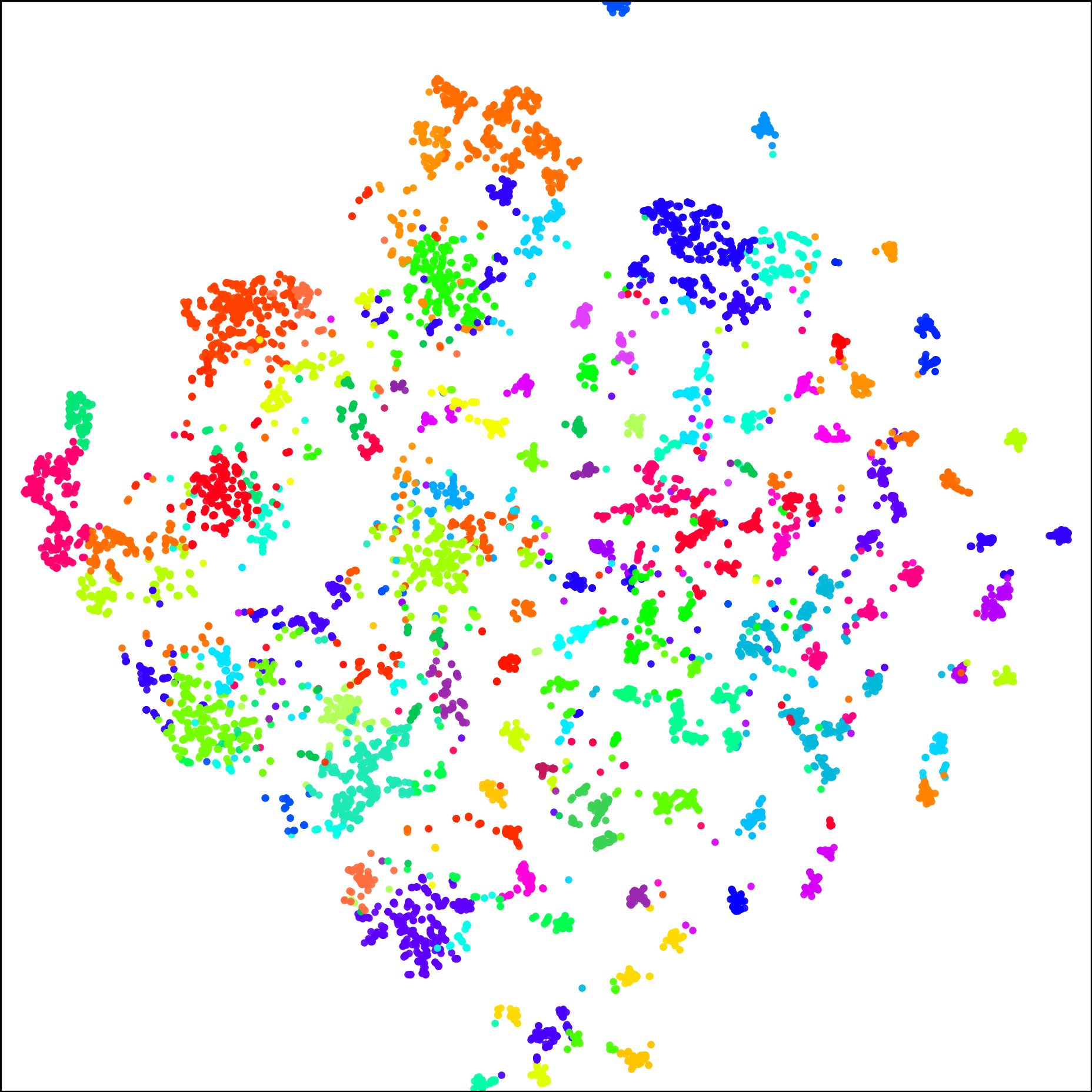}
        \small (b) After pretraining\par
        \scriptsize (ACC:0.4650 ARI:0.3115 F-Sco:0.4866)
    \end{minipage}\hfill
    \begin{minipage}{0.23\textwidth}
        \centering
        \includegraphics[width=\linewidth]{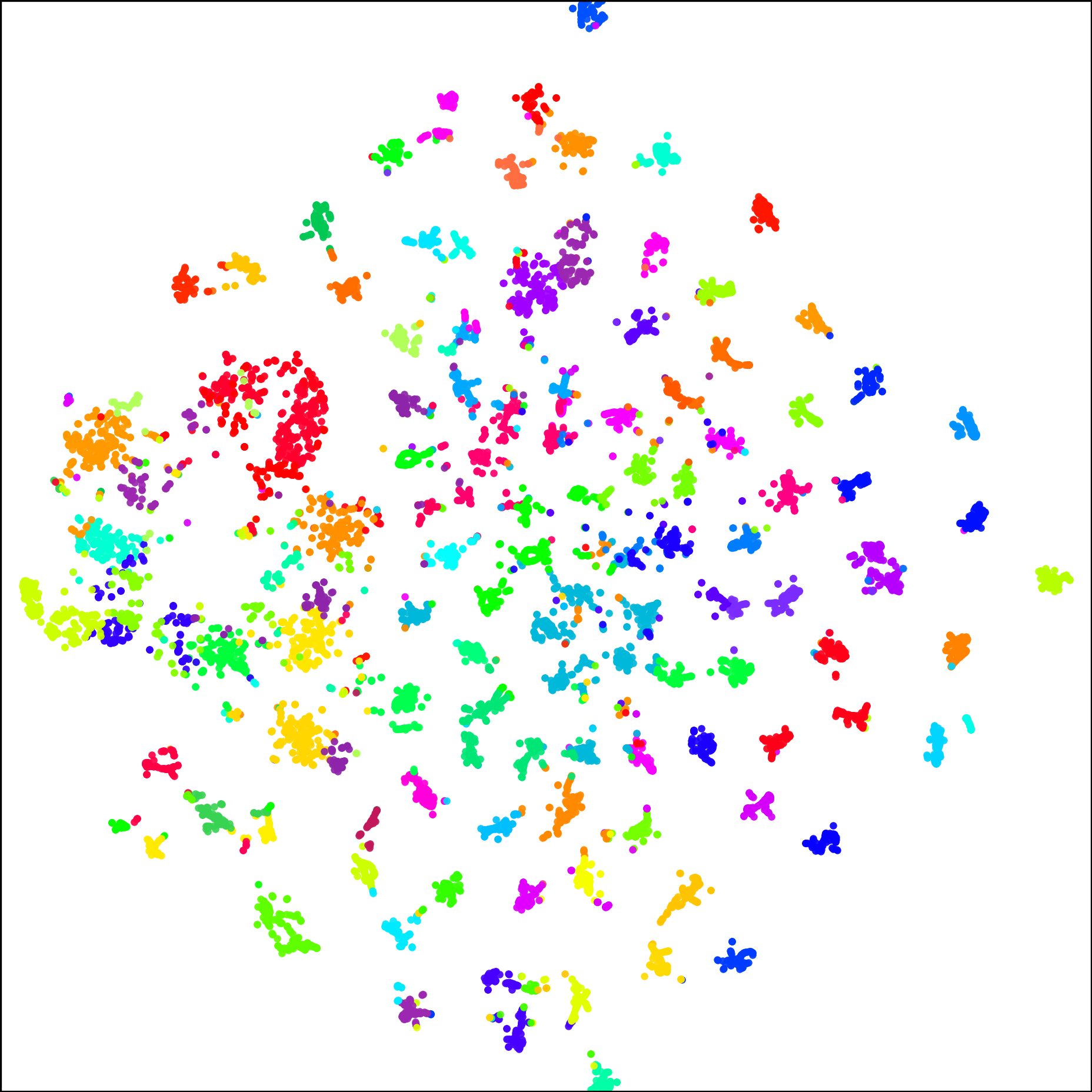}
        \small (c) After fine-tuning\par
        \scriptsize (ACC:0.5045 ARI:0.3303 F-Sco:0.5381)
    \end{minipage}\hfill
    \begin{minipage}{0.23\textwidth}
        \centering
        \includegraphics[width=\linewidth]{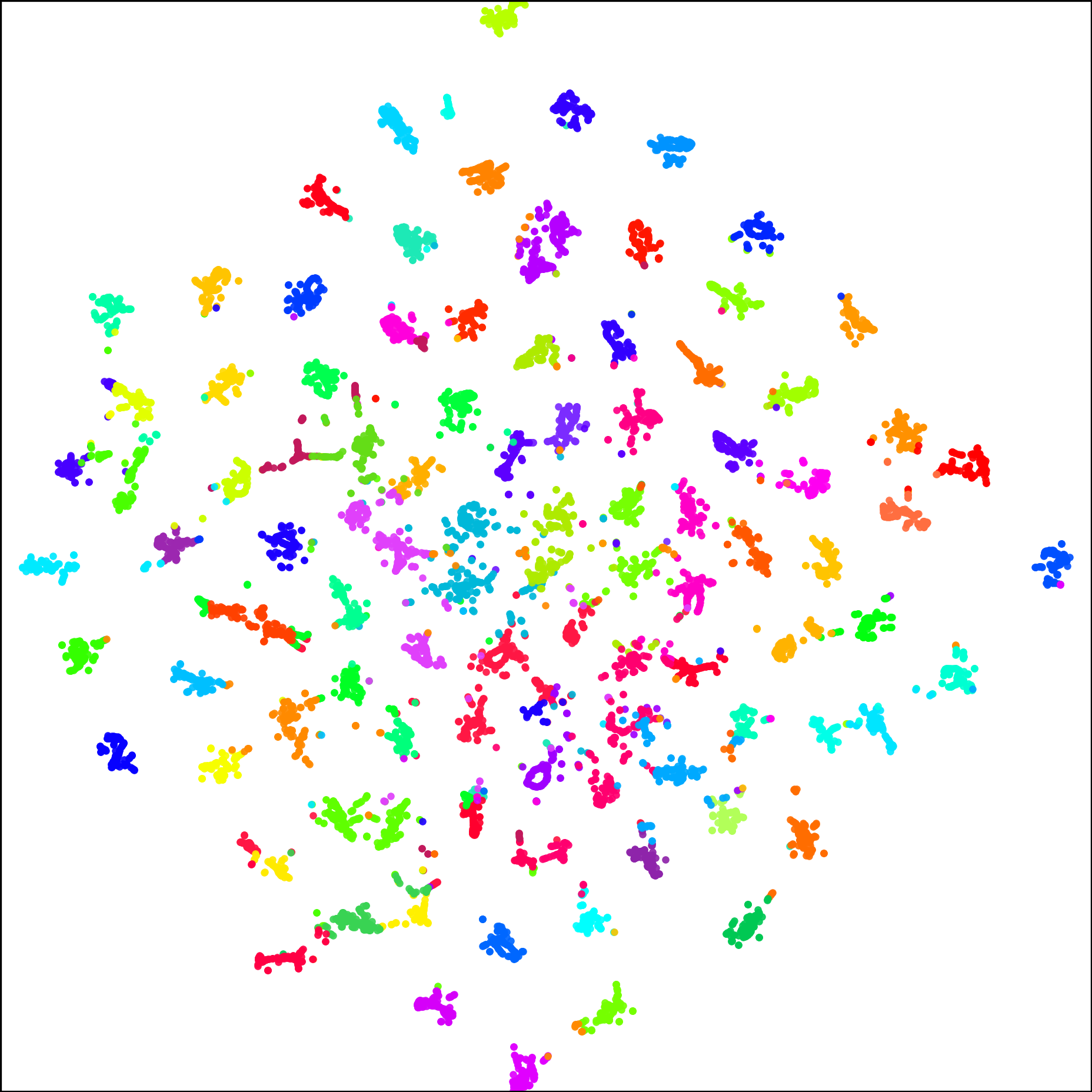}
        \small (d) After prototype imputation\par
        \scriptsize (ACC:0.7293 ARI:0.6287 F-Sco:0.7027)
    \end{minipage}

    \caption{Visualization of feature representations of ALOI\_100 at four stages.}
    \label{aloi_vis}
\end{figure*}

\subsection{Effectiveness of Hybrid Imputation}
A key component of the proposed framework lies in its hybrid imputation strategy, which reconstructs missing data by jointly leveraging informative prototypes and neighboring samples, where $\gamma$ controls their relative contributions. To better understand the behavior of this design, Fig.~\ref{fig:combined_lam5_plots} reports the ACC of SPORT on six datasets with a missing rate of 0.5, where $\gamma$ varies from 0 to 1 with a step size of 0.1. Here, $\gamma=0$ and $\gamma=1$ correspond to the neighbor-only and prototype-only imputation strategies, respectively. As shown, the hybrid strategy outperforms both degenerate variants across all datasets. For example, in Fig.~\ref{fig:combined_lam5_plots}(e), on the Reuters\_21578 dataset, SPORT achieves an ACC of 53.47\% when $\gamma=0.3$, while the neighbor-only and prototype-only variants obtain 45.67\% and 40.47\%, respectively. Similarly, in Fig.~\ref{fig:combined_lam5_plots}(f), setting $\gamma=0.3$ yields relative ACC gains of 4.33\% over the neighbor-only strategy and 87.27\% over the prototype-only strategy. It can therefore be concluded that the hybrid imputation strategy effectively balances prototype-level global structure and neighborhood-level local consistency. By integrating these two complementary sources of information, the model achieves more reliable reconstruction of missing views, which in turn leads to improved feature quality and more robust clustering performance across different datasets.

\subsection{Convergence Analysis}
To further validate the convergence behavior of SPORT, we extend 
the training to 500 epochs on (a) Digit4k, (b) 100Leaves, and 
(c) Reuters\_21578 under a missing rate of 0.5, with detailed 
results presented in Fig.~\ref{fig:curve}. Taking the 100Leaves dataset as an example, during the first 200 epochs, SPORT is pretrained to learn feature representations with high reconstruction fidelity. In this stage, the ACC smoothly increases from 33.06\% to 65.31\%, the ARI rises from 6.23\% to 52.45\%, and the F-Score improves from 28.45\% to 64.37\%. In the subsequent 100 epochs, where structure-aware relational contrastive learning and prototype partial alignment are further applied, SPORT continues to refine the learned representations and prototypes, leading to a improvement by 6.51\%, 3.7\%, and 4.75\% in ACC, ARI, and F-Score, respectively. However, extending training to 300--500 epochs results in only minor fluctuations across all three metrics without consistent improvement. This behavior suggests that the proposed framework reaches stable convergence within a moderate number of epochs, and further training yields negligible improvements.

\subsection{Representation Visualization}

Fig.\ref{aloi_vis} presents the t-SNE visualizations of the learned multi-view feature representations at different training stages, including: (a) before pretraining with raw features, (b) after pretraining, (c) after fine-tuning, and (d) after prototype imputation. The experiments are conducted on the ALOI\_100 dataset under a missing rate of 0.3. Across these four stages, the learned representations become progressively more discriminative, evolving from highly mixed and scattered distributions to compact and well-separated cluster structures. For example as shown in Fig.\ref{aloi_vis}(a), directly clustering the raw features leads to poor clustering results, with ACC, ARI, and F-Sco values of only 41.42\%, 27.18\%, and 42.53\%, respectively. Most samples are squeezed into a large mixed region on the right side of the figure, indicating that the raw heterogeneous features fail to reveal clear cluster structures. After pretraining, and fine-tuning, as illustrated in Fig.\ref{aloi_vis}(b), (c) and (d), after pretraining, fine-tuning and prototype imputation, the mixed region breaks into multiple compact clusters with distinguishable boundaries, where the corresponding ACC, ARI, and F-Sco are enhanced by 76.07\%, 131.31\% and 65.08\% respectively. Overall, the visualization results clearly illustrate a gradual transition from highly entangled feature distributions to well-separated cluster structures. This progression aligns with the staged training strategy, confirming that pretraining, fine-tuning, and prototype imputation each contribute distinctly to improving feature separability under incomplete multi-view conditions.


\section{Conclusion}
In this paper, we propose a novel incomplete multi-view clustering framework, termed SPORT, which aims to jointly preserve cross-view consistency, structural relationships, and view-specific semantics. To this end, we design a feature representation learning module to extract shared semantic representations across views, a structure-aware relational contrastive learning module to align representations by incorporating neighborhood-level structural information, and a prototype partial alignment module to effectively model complementary prototype components across views. In addition, a hybrid prototype imputation strategy is introduced to recover missing information by jointly exploiting global prototype semantics and local neighborhood context, thereby improving the reliability of cross-view feature reconstruction. Extensive experiments on multiple benchmark datasets, together with in-depth analyses and visualizations, demonstrate the superiority of the proposed framework and its effectiveness in learning discriminative representations under incomplete multi-view settings.
\section*{Acknowledgment}
This work was supported in part by the National Natural Science Foundation of China (No. U24A20331, 62536002, 62506116), the Beijing Natural Science Foundation (No. L251005), the Science Research Project of the Hebei Education Department (No. BJ2026004), and the Open Project of the Key Laboratory of Tibetan Information Processing, Ministry of Education, Qinghai Normal University (No. QHSFCS-2603).

\bibliographystyle{IEEEtranN}   
\bibliography{references}

@article{yuan2025prototype,
  author  = {Yuan, H. and Sun, Y. and Zhou, F. and Wen, J. and Yuan, S. and You, X. and Ren, Z.},
  title   = {Prototype matching learning for incomplete multi-view clustering},
  journal = {IEEE Transactions on Image Processing},
  volume  = {34},
  pages   = {828--841},
  year    = {2025}
}

@inproceedings{ding2025incomplete1,
  author    = {Ding, X. and Zhao, L. and Li, X. and Zhu, X.},
  title     = {Incomplete multi-view clustering via hierarchical semantic alignment and cooperative completion},
  booktitle = {Advances in Neural Information Processing Systems},
  volume    = {39},
  year      = {2025}
}

@article{chao2021survey,
  author  = {Chao, G. and Sun, S. and Bi, J.},
  title   = {A survey on multiview clustering},
  journal = {IEEE Transactions on Artificial Intelligence},
  volume  = {2},
  number  = {2},
  pages   = {146--168},
  year    = {2021}
}

@inproceedings{zhao2017multiview,
  author    = {Zhao, H. and Ding, Z. and Fu, Y.},
  title     = {Multi-view clustering via deep matrix factorization},
  booktitle = {Proceedings of the AAAI Conference on Artificial Intelligence},
  volume    = {31},
  number    = {1},
  year      = {2017}
}

@inproceedings{hu2018doubly,
  author    = {Hu, M. and Chen, S.},
  title     = {Doubly aligned incomplete multi-view clustering},
  booktitle = {International Joint Conference on Artificial Intelligence},
  pages     = {2262--2268},
  year      = {2018}
}

@article{wen2024matrix,
  author  = {Wen, J. and Xu, G. and Tang, Z. and Wang, W. and Fei, L. and Xu, Y.},
  title   = {Graph regularized and feature aware matrix factorization for robust incomplete multi-view clustering},
  journal = {IEEE Transactions on Circuits and Systems for Video Technology},
  volume  = {34},
  number  = {5},
  pages   = {3728--3741},
  year    = {2024}
}

@inproceedings{zhou2019consensus,
  author    = {Zhou, W. and Wang, H. and Yang, Y.},
  title     = {Consensus graph learning for incomplete multi-view clustering},
  booktitle = {Lecture Notes in Computer Science},
  pages     = {529--540},
  year      = {2019}
}

@article{wen2020incomplete,
  author  = {Wen, J. and Xu, Y. and Liu, H.},
  title   = {Incomplete multiview spectral clustering with adaptive graph learning},
  journal = {IEEE Transactions on Cybernetics},
  volume  = {50},
  number  = {4},
  pages   = {1418--1429},
  year    = {2020}
}

@article{deng2024projective,
  author  = {Deng, S. and Wen, J. and Liu, C. and Yan, K. and Xu, G. and Xu, Y.},
  title   = {Projective incomplete multi-view clustering},
  journal = {IEEE Transactions on Neural Networks and Learning Systems},
  volume  = {35},
  number  = {8},
  pages   = {10539--10551},
  year    = {2024}
}

@inproceedings{xu2018partial,
  author    = {Xu, N. and Guo, Y. and Zheng, X. and Wang, Q. and Luo, X.},
  title     = {Partial multi-view subspace clustering},
  booktitle = {Proceedings of the 26th ACM International Conference on Multimedia},
  pages     = {1794--1801},
  year      = {2018}
}

@article{ye2017consensus,
  author  = {Ye, Y. and Liu, X. and Liu, Q. and Yin, J.},
  title   = {Consensus kernel k-means clustering for incomplete multiview data},
  journal = {Computational Intelligence and Neuroscience},
  volume  = {2017},
  pages   = {1--11},
  year    = {2017}
}

@article{liu2019multiple,
  author  = {Liu, X. and Zhu, X. and Li, M. and Wang, L. and Zhu, E. and Liu, T. and Kloft, M. and Shen, D. and Yin, J. and Gao, W.},
  title   = {Multiple kernel k-means with incomplete kernels},
  journal = {IEEE Transactions on Pattern Analysis and Machine Intelligence},
  volume  = {42},
  number  = {5},
  pages   = {1191--1204},
  year    = {2019}
}

@inproceedings{guo2019anchors,
  author    = {Guo, J. and Ye, J.},
  title     = {Anchors bring ease: An embarrassingly simple approach to partial multi-view clustering},
  booktitle = {Proceedings of the AAAI Conference on Artificial Intelligence},
  volume    = {33},
  number    = {01},
  pages     = {118--125},
  year      = {2019}
}

@inproceedings{jin2023deep,
  title={Deep incomplete multi-view clustering with cross-view partial sample and prototype alignment},
  author={Jin, Jiaqi and Wang, Siwei and Dong, Zhibin and Liu, Xinwang and Zhu, En},
  booktitle={Proceedings of the IEEE/CVF conference on computer vision and pattern recognition},
  pages={11600--11609},
  year={2023}
}

@article{li2023incomplete,
  title={Incomplete multi-view clustering via prototype-based imputation},
  author={Li, Haobin and Li, Yunfan and Yang, Mouxing and Hu, Peng and Peng, Dezhong and Peng, Xi},
  journal={arXiv preprint arXiv:2301.11045},
  year={2023}
}

@article{lin2022dual,
  title={Dual contrastive prediction for incomplete multi-view representation learning},
  author={Lin, Yijie and Gou, Yuanbiao and Liu, Xiaotian and Bai, Jinfeng and Lv, Jiancheng and Peng, Xi},
  journal={IEEE Transactions on Pattern Analysis and Machine Intelligence},
  volume={45},
  number={4},
  pages={4447--4461},
  year={2022},
  publisher={IEEE}
}

@article{wang2023self,
  title={Self-supervised image clustering from multiple incomplete views via constrastive complementary generation},
  author={Wang, Jiatai and Xu, Zhiwei and Yang, Xuewen and Guo, Dongjin and Liu, Limin},
  journal={IET Computer Vision},
  volume={17},
  number={2},
  pages={189--202},
  year={2023},
  publisher={Wiley Online Library}
}

@inproceedings{lin2021completer,
  title={Completer: Incomplete multi-view clustering via contrastive prediction},
  author={Lin, Yijie and Gou, Yuanbiao and Liu, Zitao and Li, Boyun and Lv, Jiancheng and Peng, Xi},
  booktitle={Proceedings of the IEEE/CVF conference on computer vision and pattern recognition},
  pages={11174--11183},
  year={2021}
}

@article{xu2023adaptive,
  title={Adaptive feature projection with distribution alignment for deep incomplete multi-view clustering},
  author={Xu, Jie and Li, Chao and Peng, Liang and Ren, Yazhou and Shi, Xiaoshuang and Shen, Heng Tao and Zhu, Xiaofeng},
  journal={IEEE Transactions on Image Processing},
  volume={32},
  pages={1354--1366},
  year={2023},
  publisher={IEEE}
}

@inproceedings{huang2024incomplete,
  title={Incomplete multi-view clustering via inference and evaluation},
  author={Huang, Binqiang and Huang, Zhijie and Lan, Shoujie and Zheng, Qinghai and Yu, Yuanlong},
  booktitle={ICASSP 2024-2024 IEEE International Conference on Acoustics, Speech and Signal Processing (ICASSP)},
  pages={8180--8184},
  year={2024},
}

@inproceedings{yuan2024robust,
  title={Robust prototype completion for incomplete multi-view clustering},
  author={Yuan, Honglin and Lai, Shiyun and Li, Xingfeng and Dai, Jian and Sun, Yuan and Ren, Zhenwen},
  booktitle={Proceedings of the 32nd ACM international conference on multimedia},
  pages={10402--10411},
  year={2024}
}

@inproceedings{jin2025deep,
  title={Deep incomplete multi-view clustering with distribution dual-consistency recovery guidance},
  author={Jin, Jiaqi and Wang, Siwei and Dong, Zhibin and Yang, Xihong and Liu, Xinwang and Zhu, En and He, Kunlun},
  booktitle={Proceedings of the IEEE/CVF International Conference on Computer Vision},
  pages={1016--1026},
  year={2025}
}

@inproceedings{jiang2025unified,
  title={A Unified Framework to BRIDGE Complete and Incomplete Deep Multi-View Clustering under Non-IID Missing Patterns},
  author={Jiang, Xiaorui and He, Buyun and Zhou, Peng Yuan and Chen, Xinyue and Guo, Jingcai and Xu, Jie and Liao, Yong},
  booktitle={Proceedings of the IEEE/CVF International Conference on Computer Vision},
  pages={594--603},
  year={2025}
}

@article{yin2025incomplete,
  title={Incomplete multi-view clustering via multi-level contrastive learning},
  author={Yin, Jun and Wang, Pei and Sun, Shiliang and Zheng, Zhonglong},
  journal={IEEE Transactions on Knowledge and Data Engineering},
  year={2025},
  publisher={IEEE}
}

@article{yang2022robust,
  title={Robust multi-view clustering with incomplete information},
  author={Yang, Mouxing and Li, Yunfan and Hu, Peng and Bai, Jinfeng and Lv, Jiancheng and Peng, Xi},
  journal={IEEE Transactions on Pattern Analysis and Machine Intelligence},
  volume={45},
  number={1},
  pages={1055--1069},
  year={2022},
  publisher={IEEE}
}

@inproceedings{zhao2025incomplete,
  title={Incomplete and unpaired multi-view graph clustering with cross-view feature fusion},
  author={Zhao, Liang and Wang, Ziyue and Wang, Xiao and Chen, Zhikui and Xu, Bo},
  booktitle={Proceedings of the AAAI Conference on Artificial Intelligence},
  volume={39},
  number={21},
  pages={22786--22794},
  year={2025}
}

@article{wang2021generative,
  title={Generative partial multi-view clustering with adaptive fusion and cycle consistency},
  author={Wang, Qianqian and Ding, Zhengming and Tao, Zhiqiang and Gao, Quanxue and Fu, Yun},
  journal={IEEE Transactions on Image Processing},
  volume={30},
  pages={1771--1783},
  year={2021},
  publisher={IEEE}
}

@inproceedings{tang2022deep,
  title={Deep safe incomplete multi-view clustering: Theorem and algorithm},
  author={Tang, Huayi and Liu, Yong},
  booktitle={International conference on machine learning},
  pages={21090--21110},
  year={2022},
}

@inproceedings{wen2020dimc,
  title={Dimc-net: Deep incomplete multi-view clustering network},
  author={Wen, Jie and Zhang, Zheng and Zhang, Zhao and Wu, Zhihao and Fei, Lunke and Xu, Yong and Zhang, Bob},
  booktitle={Proceedings of the 28th ACM international conference on multimedia},
  pages={3753--3761},
  year={2020}
}

@article{lin2023ccr,
  title={CCR-Net: Consistent contrastive representation network for multi-view clustering},
  author={Lin, Renjie and Lin, Yongkun and Lin, Zhenghong and Du, Shide and Wang, Shiping},
  journal={Information Sciences},
  volume={637},
  pages={118937},
  year={2023},
  publisher={Elsevier}
}

@article{zhang2024incomplete,
  title={Incomplete multi-view clustering via self-attention networks and feature reconstruction: Y. Zhang et al.},
  author={Zhang, Yong and Jiang, Li and Liu, Da and Liu, Wenzhe},
  journal={Applied Intelligence},
  volume={54},
  number={4},
  pages={2998--3016},
  year={2024},
  publisher={Springer}
}

@article{xu2023untie,
  title={UNTIE: Clustering analysis with disentanglement in multi-view information fusion},
  author={Xu, Jie and Ren, Yazhou and Shi, Xiaoshuang and Shen, Heng Tao and Zhu, Xiaofeng},
  journal={Information Fusion},
  volume={100},
  pages={101937},
  year={2023},
  publisher={Elsevier}
}

@article{goodfellow2020generative,
  title={Generative adversarial networks},
  author={Goodfellow, Ian and Pouget-Abadie, Jean and Mirza, Mehdi and Xu, Bing and Warde-Farley, David and Ozair, Sherjil and Courville, Aaron and Bengio, Yoshua},
  journal={Communications of the ACM},
  volume={63},
  number={11},
  pages={139--144},
  year={2020},
  publisher={ACM New York, NY, USA}
}

@article{li2023comprehensive,
  title={A comprehensive survey on design and application of autoencoder in deep learning},
  author={Li, Pengzhi and Pei, Yan and Li, Jianqiang},
  journal={Applied Soft Computing},
  volume={138},
  pages={110176},
  year={2023},
  publisher={Elsevier}
}

@inproceedings{chao2024incomplete,
  title={Incomplete contrastive multi-view clustering with high-confidence guiding},
  author={Chao, Guoqing and Jiang, Yi and Chu, Dianhui},
  booktitle={Proceedings of the AAAI conference on artificial intelligence},
  volume={38},
  number={10},
  pages={11221--11229},
  year={2024}
}

@article{lin2023consistent,
  title={Consistent graph embedding network with optimal transport for incomplete multi-view clustering},
  author={Lin, Renjie and Du, Shide and Wang, Shiping and Guo, Wenzhong},
  journal={Information Sciences},
  volume={647},
  pages={119418},
  year={2023},
  publisher={Elsevier}
}

@article{bai2024graph,
  title={Graph-guided imputation-free incomplete multi-view clustering},
  author={Bai, Shunshun and Zheng, Qinghai and Ren, Xiaojin and Zhu, Jihua},
  journal={Expert Systems with Applications},
  volume={258},
  pages={125165},
  year={2024},
  publisher={Elsevier}
}

@inproceedings{pu2024adaptive,
  title={Adaptive feature imputation with latent graph for deep incomplete multi-view clustering},
  author={Pu, Jingyu and Cui, Chenhang and Chen, Xinyue and Ren, Yazhou and Pu, Xiaorong and Hao, Zhifeng and Yu, Philip S and He, Lifang},
  booktitle={Proceedings of the AAAI conference on artificial intelligence},
  volume={38},
  number={13},
  pages={14633--14641},
  year={2024}
}

@inproceedings{wu2025imputation,
  title={Imputation-free incomplete multi-view clustering via knowledge distillation},
  author={Wu, Benyu and Du, Wei and Wang, Jun and Yu, Guoxian},
  booktitle={Proceedings of the Computer Vision and Pattern Recognition Conference},
  pages={5071--5081},
  year={2025}
}

@inproceedings{zhang2026dynamic,
  title={Dynamic deep graph learning for incomplete multi-view clustering with masked graph reconstruction loss},
  author={Zhang, Zhenghao and Xie, Jun and Chen, Xingchen and Yu, Tao and Yi, Hongzhu and Xu, Kaixin and Wang, Yuanxiang and Zong, Tianyu and Wang, Xinming and Chen, Jiahuan and others},
  booktitle={Proceedings of the AAAI Conference on Artificial Intelligence},
  volume={40},
  number={34},
  pages={28600--28608},
  year={2026}
}

@inproceedings{chen2025deep,
  title={Deep Variational Incomplete Multi-View Clustering with Information-Theoretic Guidance},
  author={Chen, Wenlan and Gao, Lu and Liang, Cheng and Guo, Fei},
  booktitle={Proceedings of the 33rd ACM International Conference on Multimedia},
  pages={2457--2466},
  year={2025}
}

@article{dong2025selective,
  title={Selective cross-view topology for deep incomplete multi-view clustering},
  author={Dong, Zhibin and Hu, Dayu and Jin, Jiaqi and Wang, Siwei and Liu, Xinwang and Zhu, En},
  journal={IEEE Transactions on Image Processing},
  year={2025},
  publisher={IEEE}
}

@inproceedings{feng2024partial,
  title={Partial multi-view clustering via self-supervised network},
  author={Feng, Wei and Sheng, Guoshuai and Wang, Qianqian and Gao, Quanxue and Tao, Zhiqiang and Dong, Bo},
  booktitle={Proceedings of the AAAI conference on artificial intelligence},
  volume={38},
  number={11},
  pages={11988--11995},
  year={2024}
}

@inproceedings{shang2017vigan,
  title={VIGAN: Missing view imputation with generative adversarial networks},
  author={Shang, Chao and Palmer, Aaron and Sun, Jiangwen and Chen, Ko-Shin and Lu, Jin and Bi, Jinbo},
  booktitle={2017 IEEE International conference on big data (Big Data)},
  pages={766--775},
  year={2017},
}

@article{wang2026llm,
  title={LLM-DAMVC: A Large Language Model Assisted Dynamic Agent for Multi-View Clustering},
  author={Wang, Qianqian},
  journal={Advances in Neural Information Processing Systems},
  volume={38},
  pages={119515--119534},
  year={2026}
}

@article{xin2025multilevel,
  title={Multilevel reliable guidance for unpaired multiview clustering},
  author={Xin, Like and Yang, Wanqi and Wang, Lei and Yang, Ming},
  journal={IEEE Transactions on Neural Networks and Learning Systems},
  year={2025},
    volume={36},
  number={10},
  pages={18968-18982},
}

@article{gu2026hypergraph,
  title={Hypergraph-Enhanced Contrastive Learning for Multi-View Clustering with Hyper-Laplacian Regularization},
  author={Gu, Zhibin and Weili Wang},
  journal={Advances in Neural Information Processing Systems},
  volume={38},
  pages={54600--54620},
  year={2026}
}

@ARTICLE{gu2026twin,
  author={Gu, Zhibin and Feng, Songhe},
  journal={IEEE Transactions on Multimedia}, 
  title={Twin Tensor Learning for Consistency and Inconsistency: A Unified Affinity Learning Framework for Multi-View Clustering}, 
  year={2026},
  volume={28},
  number={},
  pages={4233-4244},
}

@inproceedings{wang2026adversarial,
  title={Adversarial Fair Incomplete Multi-View Clustering},
  author={Wang, Qianqian and Xu, Haiming and Feng, Wei and Gao, Quanxue},
  booktitle={Proceedings of the AAAI Conference on Artificial Intelligence},
  volume={40},
  number={12},
  pages={10011--10019},
  year={2026}
}

@inproceedings{wu2025koala,
  title={Koala: Kernel coupling and element imputation induced multi-view clustering},
  author={Wu, Tingting and Li, Zhendong and Gu, Zhibin and Yuan, Jiazheng and Feng, Songhe},
  booktitle={Proceedings of the AAAI Conference on Artificial Intelligence},
  volume={39},
  number={20},
  pages={21581--21589},
  year={2025}
}

@article{zhong2026gaussian,
  title={Gaussian regression-driven tensorized incomplete multi-view clustering with dual manifold regularization},
  author={Zhong, Zhenhao and Gu, Zhibin and Yang, Pengpeng and Guo, Ruiqiang and others},
  journal={Advances in Neural Information Processing Systems},
  volume={38},
  pages={72102--72131},
  year={2026}
}

@inproceedings{xue2021clustering,
  title={Clustering-Induced Adaptive Structure Enhancing Network for Incomplete Multi-View Data.},
  author={Xue, Zhe and Du, Junping and Zheng, Changwei and Song, Jie and Ren, Wenqi and Liang, Meiyu},
  booktitle={Ijcai},
  pages={3235--3241},
  year={2021}
}

@inproceedings{wang2018partial,
  title={Partial multi-view clustering via consistent GAN},
  author={Wang, Qianqian and Ding, Zhengming and Tao, Zhiqiang and Gao, Quanxue and Fu, Yun},
  booktitle={2018 IEEE International Conference on Data Mining (ICDM)},
  pages={1290--1295},
  year={2018},
}

@inproceedings{xu2019adversarial,
  title={Adversarial incomplete multi-view clustering.},
  author={Xu, Cai and Guan, Ziyu and Zhao, Wei and Wu, Hongchang and Niu, Yunfei and Ling, Beilei},
  booktitle={Ijcai},
  volume={7},
  pages={3933--3939},
  year={2019}
}

@inproceedings{wen2021structural,
  title={Structural deep incomplete multi-view clustering network},
  author={Wen, Jie and Wu, Zhihao and Zhang, Zheng and Fei, Lunke and Zhang, Bob and Xu, Yong},
  booktitle={Proceedings of the 30th ACM international conference on information \& knowledge management},
  pages={3538--3542},
  year={2021}
}

@inproceedings{teng2024urrl,
  title={URRL-IMVC: Unified and robust representation learning for incomplete multi-view clustering},
  author={Teng, Ge and Mao, Ting and Shen, Chen and Tian, Xiang and Liu, Xuesong and Chen, Yaowu and Ye, Jieping},
  booktitle={Proceedings of the 30th ACM SIGKDD Conference on Knowledge Discovery and Data Mining},
  pages={2888--2899},
  year={2024}
}

@article{zhang2020deep2,
  title={Deep partial multi-view learning},
  author={Zhang, Changqing and Cui, Yajie and Han, Zongbo and Zhou, Joey Tianyi and Fu, Huazhu and Hu, Qinghua},
  journal={IEEE transactions on pattern analysis and machine intelligence},
  volume={44},
  number={5},
  pages={2402--2415},
  year={2020},
  publisher={IEEE}
}

@article{xu2015multi3,
  title={Multi-view intact space learning},
  author={Xu, Chang and Tao, Dacheng and Xu, Chao},
  journal={IEEE transactions on pattern analysis and machine intelligence},
  volume={37},
  number={12},
  pages={2531--2544},
  year={2015},
  publisher={IEEE}
}

@article{white2012convex,
  title={Convex multi-view subspace learning},
  author={White, Martha and Zhang, Xinhua and Schuurmans, Dale and Yu, Yao-liang},
  journal={Advances in neural information processing systems},
  volume={25},
  pages = {1--14},
  year={2012},
}

@inproceedings{xu2022multi2,
  title={Multi-level feature learning for contrastive multi-view clustering},
  author={Xu, Jie and Tang, Huayi and Ren, Yazhou and Peng, Liang and Zhu, Xiaofeng and He, Lifang},
  booktitle={Proceedings of the IEEE/CVF conference on computer vision and pattern recognition},
  pages={16051--16060},
  year={2022}
}

@inproceedings{xu2016discriminatively,
  title={Discriminatively embedded k-means for multi-view clustering},
  author={Xu, Jinglin and Han, Junwei and Nie, Feiping},
  booktitle={Proceedings of the IEEE conference on computer vision and pattern recognition},
  pages={5356--5364},
  year={2016}
}

@article{cui2023novel,
  title={A novel approach for effective multi-view clustering with information-theoretic perspective},
  author={Cui, Chenhang and Ren, Yazhou and Pu, Jingyu and Li, Jiawei and Pu, Xiaorong and Wu, Tianyi and Shi, Yutao and He, Lifang},
  journal={Advances in neural information processing systems},
  volume={36},
  pages={44847--44859},
  year={2023}
}

@article{zhang2016multi2,
  title={Multi-task multi-view clustering},
  author={Zhang, Xiaotong and Zhang, Xianchao and Liu, Han and Liu, Xinyue},
  journal={IEEE Transactions on Knowledge and Data Engineering},
  volume={28},
  number={12},
  pages={3324--3338},
  year={2016},
  publisher={IEEE}
}

@inproceedings{trosten2023effects,
  title={On the effects of self-supervision and contrastive alignment in deep multi-view clustering},
  author={Trosten, Daniel J and L{\o}kse, Sigurd and Jenssen, Robert and Kampffmeyer, Michael C},
  booktitle={Proceedings of the IEEE/CVF conference on computer vision and pattern recognition},
  pages={23976--23985},
  year={2023}
}

@inproceedings{xiao2025easemvc,
  title={EASEMVC: Efficient Dual Selection Mechanism for Deep Multi-View Clustering},
  author={Xiao, Baili and Dong, Zhibin and Liang, Ke and Liu, Suyuan and Wang, Siwei and Liu, Tianrui and Hu, Xingchen and Zhu, En and Liu, Xinwang},
  booktitle={Proceedings of the Computer Vision and Pattern Recognition Conference},
  pages={20716--20726},
  year={2025}
}

@inproceedings{wen2024diffusion,
  title={Diffusion-based missing-view generation with the application on incomplete multi-view clustering},
  author={Wen, Jie and Deng, Shijie and Wong, Waikeung and Chao, Guoqing and Huang, Chao and Fei, Lunke and Xu, Yong},
  booktitle={Forty-first international conference on machine learning},
  pages = 	 {52762--52778},
  volume = 	 {235},
  year={2024}
}

@article{fang2024incomplete,
  title={Incomplete multi-view clustering via diffusion completion},
  author={Fang, Sifan and Yang, Zuyuan and Chen, Junhang},
  journal={Multimedia Tools and Applications},
  volume={83},
  number={18},
  pages={55889--55902},
  year={2024},
  publisher={Springer}
}

@inproceedings{zhang2025incomplete,
  title={Incomplete multi-view clustering via diffusion contrastive generation},
  author={Zhang, Yuanyang and Lin, Yijie and Yan, Weiqing and Yao, Li and Wan, Xinhang and Li, Guangyuan and Zhang, Chao and Ke, Guanzhou and Xu, Jie},
  booktitle={Proceedings of the AAAI Conference on Artificial Intelligence},
  volume={39},
  number={21},
  pages={22650--22658},
  year={2025}
}

@article{ho2020denoising,
  title={Denoising diffusion probabilistic models},
  author={Ho, Jonathan and Jain, Ajay and Abbeel, Pieter},
  journal={Advances in neural information processing systems},
  volume={33},
  pages={6840--6851},
  year={2020}
}

@article{du2025pgformer,
  title={PGFormer: A Prototype-Graph Transformer for Incomplete Multiview Clustering},
  author={Du, Yiming and Wang, Yao and Wang, Ziyu and Ning, Rui and Li, Lusi},
  journal={IEEE Transactions on Neural Networks and Learning Systems},
  year={2026},
  volume={37},
  number={3},
  pages={1163-1175},
}

@inproceedings{zhu2026prototype,
  title={Prototype and Sample Level Semantic Alignment for Incomplete Multi-View Clustering},
  author={Zhu, Zhengzhong and Zhou, Pei and Bai, Lanxi and Nie, Jia and Cheng, Li and Min, Shiquan and Zhu, Jiangping},
  booktitle={Proceedings of the IEEE/CVF Conference on Computer Vision and Pattern Recognition},
  pages={5818--5827},
  year={2026}
}

@article{zhang2019cpm,
  title={CPM-Nets: Cross partial multi-view networks},
  author={Zhang, Changqing and Han, Zongbo and Fu, Huazhu and Zhou, Joey Tianyi and Hu, Qinghua and others},
  journal={Advances in Neural Information Processing Systems},
  volume={32},
  year={2019}
}

@article{yan2025neighbor,
  title={Neighbor-Based Completion for Addressing Incomplete Multiview Clustering},
  author={Yan, Wenbiao and Zhu, Jihua and Zhou, Yiyang and Chen, Jinqian and Cheng, Haozhe and Yue, Kun and Zheng, Qinghai},
  journal={IEEE Transactions on Neural Networks and Learning Systems},
  year={2025},
  publisher={IEEE}
}

@inproceedings{lu2024decoupled,
  title={Decoupled contrastive multi-view clustering with high-order random walks},
  author={Lu, Yiding and Lin, Yijie and Yang, Mouxing and Peng, Dezhong and Hu, Peng and Peng, Xi},
  booktitle={Proceedings of the AAAI conference on artificial intelligence},
  volume={38},
  number={13},
  pages={14193--14201},
  year={2024}
}

@article{corso2024graph,
  title={Graph neural networks},
  author={Corso, Gabriele and Stark, Hannes and Jegelka, Stefanie and Jaakkola, Tommi and Barzilay, Regina},
  journal={Nature Reviews Methods Primers},
  volume={4},
  number={1},
  pages={17},
  year={2024},
  publisher={Nature Publishing Group UK London}
}

@article{peterson2009k,
  title={K-nearest neighbor},
  author={Peterson, Leif E},
  journal={Scholarpedia},
  volume={4},
  number={2},
  pages={1883},
  year={2009}
}

@article{jauch2020random,
  title={Random orthogonal matrices and the Cayley transform},
  author={JAUCH, MICHAEL and HOFF, PETER D and DUNSON, DAVID B},
  journal={Bernoulli},
  volume={26},
  number={2},
  pages={1560--1586},
  year={2020}
}

@article{wang2026learning,
  title={Learning from Disjoint Views: A Contrastive Prototype Matching Network for Fully Incomplete Multi-View Clustering},
  author={Wang, Yiming and Li, Qun and Chang, Dongxia and Wen, Jie and Dai, Hua and Xiao, Fu and Zhao, Yao},
  journal={Advances in Neural Information Processing Systems},
  volume={38},
  pages={18265--18284},
  year={2026}
}

@article{li2025attention,
  title={Attention-based deep incomplete multi-view clustering via bi-alignment guidance},
  author={Li, Ao and Mei, Sanlin and Gu, Fengwei and Miao, Dehua and Gao, Tianyu},
  journal={Complex \& Intelligent Systems},
  volume={11},
  number={8},
  pages={349},
  year={2025},
  publisher={Springer}
}

@ARTICLE{11494292,
  author={Zou, Xin and Liu, Ruimeng and Tang, Chang and Li, Zhenglai and Liu, Xinwang and He, Kunlun and Li, Wanqing},
  journal={IEEE Transactions on Pattern Analysis and Machine Intelligence}, 
  title={Learning Disentangled Representations for Generalized Multi-view Clustering}, 
  year={2026},
  volume={},
  number={},
  pages={1-16},
}

@ARTICLE{liu2025reliable,
  author={Liu, Chengliang and Wen, Jie and Xu, Yong and Zhang, Bob and Nie, Liqiang and Zhang, Min},
  journal={IEEE Transactions on Pattern Analysis and Machine Intelligence}, 
  title={Reliable Representation Learning for Incomplete Multi-View Missing Multi-Label Classification}, 
  year={2025},
  volume={47},
  number={6},
  pages={4940-4956},
}

@ARTICLE{11475667,
  author={Qin, Yang and Feng, Yanglin and Sun, Yuan and Peng, Dezhong and Peng, Xi and Hu, Peng},
  journal={IEEE Transactions on Pattern Analysis and Machine Intelligence}, 
  title={Deep Information-Balanced Multimodal Learning}, 
  year={2026},
  volume={},
  number={},
  pages={1-13},
}

@ARTICLE{11458690,
  author={Qiang, Qianyao and Zhang, Bin and Hua, Yunjia and Nie, Feiping},
  journal={IEEE Transactions on Pattern Analysis and Machine Intelligence}, 
  title={Multi-View Clustering Via Bilaterally Constrained Anchor Graph}, 
  year={2026},
  volume={},
  number={},
  pages={1-14},
}

@ARTICLE{11277376,
  author={Liu, Xinwang and Liang, Ke and Wang, Jun and Liu, Suyuan and Wang, Xiangke and Wang, Huaimin},
  journal={IEEE Transactions on Pattern Analysis and Machine Intelligence}, 
  title={Two Decades of Multi-View Clustering: Taxonomy, Application, and Challenge}, 
  year={2026},
  volume={48},
  number={3},
  pages={3744-3764},
}




\vfill
\end{document}